\documentclass[11pt]{article}
\usepackage[margin=1in]{geometry}
\usepackage{amsmath,amssymb,amsthm,mathtools,bm}
\usepackage{booktabs,array,multirow}
\usepackage{enumitem}
\usepackage[numbers,square]{natbib}
\usepackage{microtype}
\usepackage{hyperref}
\usepackage{cleveref}
\usepackage{xcolor}
\usepackage{mathrsfs}
\usepackage{graphicx}
\usepackage{longtable}
\usepackage{tikz}
\usetikzlibrary{arrows.meta,positioning,calc,fit,decorations.pathreplacing,backgrounds}

\hypersetup{colorlinks=true,linkcolor=blue,citecolor=blue,urlcolor=blue}
\usepackage{changepage}
\usepackage{tocloft}
\addtocontents{toc}{\protect\setcounter{tocdepth}{-1}}
\allowdisplaybreaks
\numberwithin{equation}{section}

\newtheorem{theorem}{Theorem}[section]
\newtheorem{proposition}[theorem]{Proposition}
\newtheorem{lemma}[theorem]{Lemma}
\newtheorem{corollary}[theorem]{Corollary}
\newtheorem{assumption}[theorem]{Assumption}
\newtheorem{remark}[theorem]{Remark}

\newtheorem{definition}[theorem]{Definition}

\newcommand{\R}{\mathbb R}
\newcommand{\E}{\mathbb E}
\newcommand{\Pp}{\mathbb P}
\newcommand{\1}{\mathbf{1}}
\newcommand{\cX}{\mathcal X}
\newcommand{\cG}{\mathcal G}

\newcommand{\cC}{\mathcal C}
\newcommand{\cQ}{\mathcal Q}

\newcommand{\cP}{\mathcal P}

\newcommand{\diam}{\operatorname{diam}}
\newcommand{\dist}{\operatorname{dist}}

\newcommand{\DeltaSep}{\Delta_{\mathrm{sep}}}
\newcommand{\DeltaMacro}{D_0}
\newcommand{\Hdist}{\mathsf H}
\newcommand{\Dfunc}{D_{\mathrm{func}}}
\newcommand{\Dtop}{D_K}

\newcommand{\DOT}{D_{\mathrm{OT}}}
\newcommand{\Dmc}{D_{\mathrm{cluster}}}
\newcommand{\smax}{s_{\max}}

\newcommand{\phiSig}{\phi_{\sigma_0}}
\newcommand{\norm}[1]{\left\lVert #1\right\rVert}

\title{When Does Sparse Routing Help?\\ The Geometry of Expert Separation in Mixture of Experts}
\author{%
\begin{minipage}{\textwidth}
\centering
\normalsize
{\large
\begin{tabular}{cccc}
Huy Nguyen & Dung Le & Alessandro Rinaldo & Nhat Ho
\end{tabular}
}
\endgraf
\vspace*{.2in}
\begin{tabular}{c}
Department of Statistics and Data Science\\
The University of Texas at Austin
\end{tabular}
\endgraf
\vspace*{.2in}
\today
\end{minipage}%
}
\date{}

\begin{document}
\maketitle
\begin{abstract}
Sparse mixtures of experts combine multiple prediction models while evaluating only a small subset for each input. Although this architecture reduces computation, how expert separation and routing sparsity jointly govern expert recovery remains an open problem. In this paper, we address this question for location Gaussian mixtures of experts with a fixed number of experts by establishing how expert separation and spatial arrangement determine recovery rates. For dense models, our results reveal that when experts form well-separated clusters, the separation geometry is governed by the size of the largest cluster; without a prescribed cluster structure, this geometry is instead driven by the total number of experts. For sparse models, the separation geometry on retained routing regions is determined by the number of active experts, while the contribution of omitted regions is controlled by their total probability. Consequently, activating fewer experts reduces the separation complexity in these bounds. Furthermore, Top-1 routing yields a particularly striking result: experts can be recovered at a rate independent of the separation quantity. Our technical approach combines polynomial interpolation and Gaussian reproducing-kernel Hilbert space (RKHS) test functions with quantitative restriction inequalities that extend pointwise comparisons to expert functions.
\end{abstract}
\section{Introduction}\label{sec:introduction}
Mixtures of experts (MoE) model heterogeneous input--output relationships by combining several prediction rules through an input-dependent gate \citep{Jacobs1991,JordanJacobs1994}. The prediction rules, called \emph{experts}, may be regression functions, classifiers, or neural networks. The gate assigns weights to their outputs, allowing the contribution of each expert to vary across the input space. This structure accommodates differences between subpopulations while retaining a common predictive model.

Sparse MoE limits the number of experts evaluated at each input. A Top-$K$ router selects the $K$ highest-scoring experts and combines their outputs using normalized weights. The architecture can therefore increase its total parameter count while controlling the computation per input \citep{Shazeer2017,Fedus2021}. Applications include multilingual machine translation \citep{Lepikhin2020}, language modeling \citep{Du2022,Jiang2024}, image recognition \citep{Riquelme2021}, and multimodal learning \citep{Han2024FuseMoE}. More recent architectures, including DeepSeekMoE, explicitly seek to encourage specialization and reduce redundancy among experts \citep{Dai2024DeepSeekMoE}.

\paragraph{From predictive accuracy to expert recovery.}
Specialization motivates a statistical problem distinct from prediction: estimating the individual expert functions from observations of the aggregate conditional law. Existing MoE theory studies the identifiability and estimation of this latent decomposition, including interactions between expert and gate parameters \citep{HoYangJordan2022,NguyenDemystifying2023}. The accuracy of a fitted conditional density need not translate into comparable accuracy for its constituent experts. When experts give similar predictions where they are jointly active, changes in their functions and weights can partially cancel in the mixture. Related distinguishability results show that overlap between a fixed expert and a trainable expert can substantially slow parameter estimation \citep{YanContaminated2025}. We study this loss of information directly through functional recovery losses, which quantify the estimability of expert functions up to an appropriate matching of labels.

\paragraph{Problem setup.}\label{sec:intro-setup}
Fix a known number of experts $k\ge2$ and an active count $1\le K\le k$. Assume that $(X_1,Y_1),\ldots,(X_n,Y_n)$ are independent and identically distributed observations generated from a Gaussian mixture of experts with hard Top-$K$ routing and conditional density
\begin{equation}\label{eq:intro-model}
p_{G_*}^{(K)}(y\mid x)
:=\sum_{j\in S_{G_*}(x)}\pi_{j,G_*}^{(K)}(x)\phiSig\bigl(y-h(x,\eta_j^*)\bigr).
\end{equation}
Here $X\sim P_X=\operatorname{Unif}(\cX)$ with $\cX=[-b,b]^{d_x}\subset\R^{d_x}$, $b>0$, and $Y\in\R$. All Gaussian components have a known common variance $\sigma_0>0$, with density
\begin{equation}\label{eq:gaussian-density}
\phiSig(u):=\frac{1}{\sqrt{2\pi\sigma_0}}e^{-u^2/(2\sigma_0)}.
\end{equation}
Let $h:\cX\times\Omega_\eta\to\R$ be the expert family, where $\Omega_\eta\subset\R^{d_\eta}$ and $d_x,d_\eta\ge1$. The model is indexed by the unknown true tuple
\[
G_*=((\beta_{1j}^*,\beta_{0j}^*,\eta_j^*))_{j=1}^k,
\qquad
(\beta_{1j}^*,\beta_{0j}^*)\in\Theta_\beta\subset\R^{d_x}\times\R,
\quad \eta_j^*\in\Omega_\eta.
\]
Write $h_j^*(x):=h(x,\eta_j^*)$ for the $j$th expert function and $\ell_j^*(x):=(\beta_{1j}^*)^\top x+\beta_{0j}^*$ for its logit. The hard Top-$K$ softmax gate assigns weight
\[
\pi_{j,G_*}^{(K)}(x):=
\1_{\{j\in S_{G_*}(x)\}}
\frac{e^{\ell_j^*(x)}}{\sum_{a\in S_{G_*}(x)}e^{\ell_a^*(x)}},
\qquad j\in[k].
\]
Here $S_{G_*}(x)\subset[k]$ is the active set consisting of the indices of the $K$ largest routing scores $(\beta_{1j}^*)^\top x$, with ties resolved by a fixed priority ordering. Thus selection depends on the slope scores, whereas the weights of the selected experts depend on the full logits, including intercepts. Exactly $K$ of the $k$ labelled experts are active at each covariate.

The dense MoE is the special case $K=k$, when every expert is active. We omit the active-count superscript in this case and write
\[
\pi_j^*(x):=\pi_{j,G_*}^{(k)}(x)
=\frac{e^{\ell_j^*(x)}}{\sum_{a=1}^k e^{\ell_a^*(x)}},
\qquad
p_{G_*}(y\mid x):=p_{G_*}^{(k)}(y\mid x)
=\sum_{j=1}^k\pi_j^*(x)\phiSig\bigl(y-h_j^*(x)\bigr).
\]
We denote the sample law under $p_{G_*}^{(K)}$ by $\Pp_*^{(K)}$ and its dense specialization by $\Pp_*:=\Pp_*^{(k)}$. Statistical statements for different active counts concern their respective sampling experiments, all with covariate law $P_X$. The minimum functional separation of the true experts is
\[
\DeltaSep:=\min_{j\ne\ell}\|h_j^*-h_\ell^*\|_{L^1(P_X)},
\qquad
\|f\|_{L^1(P_X)}:=\int_{\cX}|f(x)|\,dP_X(x).
\]

Our primary goal is to estimate the individual expert functions using maximum likelihood and quantify their errors in $L^1(P_X)$, up to a matching of fitted and true labels. Define the fitted class, which contains $G_*$, by
\[
\cG_k(\Theta_\beta,\Omega_\eta)
:=\left\{G=((\beta_{1i},\beta_{0i},\eta_i))_{i=1}^k:
(\beta_{1i},\beta_{0i})\in\Theta_\beta,\ \eta_i\in\Omega_\eta\right\}.
\]
For a candidate tuple $G$ in this class, write $h_i(x)=h(x,\eta_i)$ and $\ell_i(x)=\beta_{1i}^\top x+\beta_{0i}$. The density $p_G^{(K)}$ is obtained from the same model formula by replacing $G_*$ with $G$. The Top-$K$ MLE is
\[
\widehat G_n^{(K)}\in\operatorname*{arg\,max}_{G\in\cG_k(\Theta_\beta,\Omega_\eta)}
\sum_{a=1}^n\log p_G^{(K)}(Y_a\mid X_a).
\]
When $K=k$, we write $\pi_i(x)=\pi_i(x;G):=\pi_{i,G}^{(k)}(x)$, $p_G:=p_G^{(k)}$, and $\widehat G_n:=\widehat G_n^{(k)}$ for the dense weights, density, and MLE. The likelihood results concern measurable global maximizers; Section~\ref{sec:mle} establishes existence in the dense setting and states the existence requirement for hard routing.

Writing $\widehat\eta_i$ for the expert parameters in the dense estimator, the expert-estimation error of interest is
\[
\|h(\cdot,\widehat\eta_i)-h(\cdot,\eta_j^*)\|_{L^1(P_X)}
=\int_{\cX}|h(x,\widehat\eta_i)-h(x,\eta_j^*)|\,dP_X(x),
\]
with $i$ matched to $j$. We seek rates for these individual errors with explicit dependence on $\DeltaSep$, cluster structure, and the active count $K$. Dense clustered recovery provides one matching of whole expert functions and hence rates in this full-domain norm. The unstructured result also yields one global matching, without a cluster diameter or fitted-localization condition; class-wide transport restriction supplies the required expert restriction. Sparse recovery provides errors restricted to joint routing regions, where the corresponding experts are active, with matchings that may vary between regions. Full-domain recovery of each sparse expert requires additional exposure and matching conditions. The combined expert-and-gate losses introduced below are tools for deriving these conclusions, and also quantify gate-probability errors. Section~\ref{sec:mle} gives the likelihood rates and their transfer; the deterministic inverse inequalities apply to arbitrary admissible model pairs.

\paragraph{Why separation geometry matters.}
Separation determines how stably the conditional density identifies its latent experts. At a fixed covariate, nearby Gaussian means can produce similar component densities, allowing appreciable changes in the mixing distribution to have a small effect on the observable law. Quantitative mixture-identification bounds and cluster-adaptive estimation results make this dependence explicit \citep{Nguyen2013,WuYang2020}. A minimum distance alone does not describe the configuration: a collection of small, well-separated clusters presents different local inverse problems from a single cluster containing all experts. In MoE, these configurations also vary with the covariate. Functional separation must therefore be related to pointwise separation on sufficiently informative subsets of $\cX$. Sparse routing adds the requirement that experts be distinguishable where they are active, with the probabilities of those regions determining their contribution to the recovery loss. Our bounds track these features through cluster size, the active count, and conditional geometry. This leads to two questions.

\begin{center}
\setlength{\fboxsep}{9pt}
\fbox{\begin{minipage}{0.91\linewidth}
\textbf{Question:}
 \emph{How do separation, sparsity, and cluster structure affect expert recovery?}
% \begin{enumerate}[label=(\roman*),leftmargin=1.7em,itemsep=0.4em,topsep=0.5em]
% \item How do separation and cluster structure affect expert recovery?
% \item How does the sparsity affect expert exposure?
% \end{enumerate}
\end{minipage}}
\end{center}

\subsection{Related literature}\label{sec:intro-literature}
\paragraph{Parameter and expert estimation in MoE.}
Much of the statistical theory of MoE concerns identifiability and parameter estimation. Ho et al.~\citep{HoYangJordan2022} study Gaussian MoE with covariate-free gates, and Nguyen et al.~\citep{NguyenDemystifying2023} analyze parameter interactions in softmax-gated Gaussian MoE. For hard Top-$K$ gates, Nguyen et al.~\citep{NguyenTopK2024} establish density and parameter rates using losses that account for the routing partition. Related results concern sigmoid gates \citep{NguyenSigmoid2024}, the perturbed cosine router \citep{NguyenCosine2025}, and Bayesian softmax MoE \citep{Bariletto2026}. These works identify the expert family, gating mechanism, and specification of the expert count as determinants of estimation difficulty.

Expert estimation is often studied by transferring parameter rates through regularity of the expert map. For instance, Nguyen et al.~\citep{NguyenDeepSeekTheory2025} use inequalities of the form
\[
\|h(\cdot,\widehat\eta)-h(\cdot,\eta^*)\|_{L^p(P_X)}
\le L\|\widehat\eta-\eta^*\|_2.
\]
This inequality transfers an upper bound, but does not show that the resulting rate is optimal for the functional target. Parameter perturbations can produce much smaller changes in the expert function. Moreover, constants that depend implicitly on the true model do not quantify deterioration as experts approach one another. We address these issues by proving inverse inequalities directly for functional losses and tracking their dependence on separation. The results concern different targets and assumptions from much of the parameter-based theory, and should be compared on that basis.

\paragraph{MoE theory in deep learning.}
Another line of work studies how routing and expert specialization affect learning in neural architectures. For a classification problem with clustered inputs, Chen et al.~\citep{ChenDeepMoE2022} show that a router can learn cluster-center features and divide the problem among two-layer convolutional experts, enabling learning where a single expert struggles. Chowdhury et al.~\citep{ChowdhuryPatchMoE2023} establish sample-complexity improvements from patch-level routing for mixtures of two-layer convolutional networks, showing how routing informative patches to experts can improve generalization. In continual learning, Li et al.~\citep{LiContinualMoE2025} analyze sequential overparameterized linear regression tasks and characterize how expert specialization and routing affect expected forgetting and generalization error. Their analysis also shows that increasing the number of experts can delay convergence without necessarily improving performance. Connecting MoE to transformer adaptation, Le et al.~\citep{LePromptMoE2024} represent attention through linear experts with quadratic gating scores and interpret prefix tuning as adding task-specific experts, motivating a new gating mechanism for prompt-based continual learning. These studies provide theoretical explanations for specialization, sample efficiency, and adaptation under their respective model and data assumptions. Our results complement these perspectives by quantifying how separation between latent expert functions and the number of simultaneously active experts affect statistical recovery rates in Gaussian MoE.

% \paragraph{Specialization, overlap, and routing geometry.}
% Expert distinguishability has been studied in several related settings. Yan et al.~\citep{YanContaminated2025} analyze the effect of overlap between a fixed pretrained expert and a trainable expert in softmax-contaminated MoE. Nielsen et al.~\citep{NielsenClusters2025} use clusters in the input representation to design a router that promotes specialization. Our separation geometry instead concerns distances between expert functions and between the conditional Gaussian means they induce. To our knowledge, existing results do not give separation-explicit functional recovery bounds distinguishing clustered and unstructured dense models, together with retained-region guarantees for hard Top-$K$ routing.

\paragraph{Separation in finite mixture models.}
In finite-mixture models, quantitative identifiability inequalities connect density estimation to recovery of the mixing measure in Wasserstein distance \citep{Nguyen2013,HoNguyen2016EJS,HeinrichKahn2018}. For Gaussian location mixtures, Wu and Yang~\citep{WuYang2020} obtain rates that adapt to separated clusters using moment comparison and polynomial interpolation. Nguyen et al.~\citep{NguyenSeparation2026} further track separation and minimum-weight dependence in single-cluster, multiple-cluster, and unstructured configurations. Extending this perspective to MoE requires an additional step. The mixture atoms now vary with the covariate, and pointwise recovery need not yield a common matching of whole expert functions. Sparse routing also requires control of the probability and geometry of the regions on which particular experts are active.

\subsection{Contributions and main findings}\label{sec:intro-contributions}
Our main contribution is to quantify how separation between expert functions affects their recovery from the aggregate conditional density. The likelihood density error has rate $\sqrt{\log n/n}$ for fixed architecture, without a separation assumption. Recovering the individual experts requires an additional inverse step, whose stability depends on their separation. We make this dependence explicit in upper bounds for the experts' mean absolute errors, showing how the separation penalty changes with cluster structure and the number of simultaneously active experts. Throughout, the Gaussian variance is known and common, and the fitted and true expert counts agree. The rates below hold under the corresponding likelihood, geometry, and restriction hypotheses, with the remaining constants uniform as $\DeltaSep$ varies.

\begin{enumerate}[label=(\roman*),leftmargin=*,itemsep=0.6em]
\item \textbf{Separation controls recovery of globally matched dense experts.}
For dense MoE without a prescribed cluster partition, we obtain the expert recovery bound
\[
\mathcal O_P\!\left(\DeltaSep^{-(2k-2)}\sqrt{\log n/n}\right)
\]
for the maximum mean absolute error of the matched expert functions over the full covariate domain. One global permutation matches all fitted and true experts. The bound makes explicit how the guarantee deteriorates as the minimum separation decreases, even though the density estimation rate remains unchanged. It requires neither fitted localization nor an error condition relative to separation (Theorem~\ref{thm:denseU} and Corollary~\ref{cor:densemle-recovery}(ii)).

Cluster structure refines this dependence. When the true experts form separated clusters with largest size $\smax$, the corresponding bound becomes
\[
\mathcal O_P\!\left(\DeltaSep^{-(2\smax-2)}\sqrt{\log n/n}\right)
\]
under one cluster-preserving matching (Theorem~\ref{thm:densemcfunc} and Corollary~\ref{cor:densemle-recovery}(i)). Thus the largest cluster, rather than the total number of experts, governs the displayed penalty for shrinking within-cluster separation, with the between-cluster scale held fixed. Several-cluster recovery requires stable fitted cluster membership and the stated restriction conditions. These results distinguish configurations with the same minimum separation but different cluster sizes.

\item \textbf{Sparse routing reduces the separation exponent on active regions.}
For Top-$K$ routing with $K\ge2$, we obtain the retained-region expert recovery bound
\[
\mathcal O_P\!\left(C_{K,\lambda}^{-1}\DeltaSep^{-(2K-2)}\sqrt{\log n/n}\right),
\]
where $\lambda$ is a probability threshold and $C_{K,\lambda}$ depends on the conditional geometry and restriction constants. The error sums matched expert discrepancies over retained routing regions under the original covariate distribution. Under single-cluster geometry for the active true experts on those regions, the separation exponent is $2K-2$, compared with $2k-2$ for the dense one-cluster bound (Theorem~\ref{thm:rare} and Corollary~\ref{cor:topmle}). This identifies a way sparse routing can moderate the effect of small separation: fewer Gaussian components need to be distinguished at each input.

The regional guarantees also account for exposure. Extending the loss to all routing regions adds a remainder controlled by the square root of their omitted probability; Proposition~\ref{prop:sieve-rate} balances this remainder against deterioration of the conditional inverse constant. Matchings may differ across regions, and the target concerns expert values where they are active. Consequently, the smaller separation exponent alone does not establish better full-domain recovery of every latent expert.

\item \textbf{Top-1 permits recovery of the selected function without separation.}
With one active expert, the conditional law is a single Gaussian, and the selected expert function has mean absolute error
\[
\mathcal O_P\!\left(\sqrt{\log n/n}\right)
\]
without expert-separation or quantitative restriction assumptions (Corollary~\ref{cor:top1} and Eq.~\eqref{eq:top1-mle-rate}). This guarantee remains valid when latent experts approach one another or coincide. It identifies the endpoint of the separation dependence: distinguishing simultaneously active components is unnecessary for the selected-function target. Recovering every latent expert under one global matching remains a different objective and requires additional exposure and matching conditions.
\end{enumerate}

The deterministic inverse inequalities establish these separation penalties directly for functional recovery losses; the likelihood results transfer them to statistical upper bounds. Verification for linear experts gives concrete settings in which the required assumptions hold. The bounds quantify the effect of separation on the recovery guarantees; matching lower bounds and optimality of the exponents remain open.

Table~\ref{tab:intro-rates} summarizes the likelihood recovery bounds and their expert-error consequences. Use the density scales
\[
\varepsilon_{n,\rm den}:=\sqrt{\frac{d_{\rm den}\log(Cn)}{n}},
\qquad
\varepsilon_{n,K}:=\sqrt{\frac{\mathfrak E_{n,K}}{n}},
\qquad
q_{n,K}:=\DeltaSep^{-(2K-2)}\varepsilon_{n,K},
\]
where $d_{\rm den}=k(d_\eta+d_x+1)$ and $\mathfrak E_{n,K}$ is the routing and smooth-function complexity defined in Eq.~\eqref{eq:top-complexity}. For fixed architecture, both density scales have order $\sqrt{\log n/n}$. The deterministic probability threshold is $\lambda$, and $C_{K,\lambda}$ is the inverse constant on retained regions. Section~\ref{sec:mle} gives the precise hypotheses and probability bounds. Each table entry is an $\mathcal O_P$ upper bound, with the expert-error norms specified in the contributions above. The Top-$K$ rows concern $2\le K\le k$ and fixed architecture. The balanced rate requires $C_{K,\lambda}\ge c_0\lambda^\alpha$ for $\alpha>0$ and $q_{n,K}\to0$.

\begin{table}[!htbp]
\centering
\footnotesize
\caption{Likelihood recovery rates under the corresponding uniform hypotheses. All dense rows control globally matched experts under their respective inverse hypotheses. The unstructured row uses no prescribed cluster partition. Top-$K$ rows use regional matchings; Top-1 concerns the selected function. Fixed constants are suppressed; the balanced row uses the oracle threshold in Proposition~\ref{prop:sieve-rate}.}
\label{tab:intro-rates}
\setlength{\tabcolsep}{4pt}
\renewcommand{\arraystretch}{1.3}
\begin{tabular}{@{}>{\raggedright\arraybackslash}p{0.29\textwidth}
>{\raggedright\arraybackslash}p{0.39\textwidth}
>{\raggedright\arraybackslash}p{0.27\textwidth}@{}}
\toprule
Setting & Rate & Statistical result \\
\midrule
Dense, one cluster & $\DeltaSep^{-(2k-2)}\varepsilon_{n,\rm den}$ & Corollary~\ref{cor:densemle-recovery}(i), $q=1$ \\
Dense, several clusters & $\DeltaSep^{-(2\smax-2)}\varepsilon_{n,\rm den}$ & Corollary~\ref{cor:densemle-recovery}(i) \\
Dense, unstructured & $\DeltaSep^{-(2k-2)}\varepsilon_{n,\rm den}$ & Corollary~\ref{cor:densemle-recovery}(ii) \\
Top-$K$, retained regions & $C_{K,\lambda}^{-1}q_{n,K}$ & Eq.~\eqref{eq:topk-inverse-retained}; Theorem~\ref{thm:topmle} \\
Top-$K$, full partition & $C_{K,\lambda}^{-1}q_{n,K}+\sqrt\lambda$ & Corollary~\ref{cor:topmle} \\
Top-$K$, balanced threshold & $q_{n,K}^{1/(2\alpha+1)}$ & Proposition~\ref{prop:sieve-rate} \\
Top-1 & $\varepsilon_{n,1}$ & Eq.~\eqref{eq:top1-mle-rate} \\
\bottomrule
\end{tabular}
\end{table}

\paragraph{Organization.}
Section~\ref{sec:prelim} collects the notation and common assumptions. Section~\ref{sec:dense} develops dense recovery under clustered and unstructured geometries, and Section~\ref{sec:topk} treats Top-$K$ routing and the Top-1 specialization. Section~\ref{sec:mle} proves likelihood rates and transfers them to the recovery losses. Section~\ref{sec:linear} verifies the assumptions for linear experts, and Section~\ref{sec:discussion} discusses the implications, limitations, and further directions. Appendices~\ref{app:scalar}--\ref{app:linear} contain the inverse and likelihood proofs and the linear verification; Appendix~\ref{app:partialot} explains the partial-transport structure of the proposed losses.

\section{Notation and Universal Assumptions}\label{sec:prelim}
\subsection{Notation}\label{sec:notation}
We use the model notation of Section~\ref{sec:intro-setup} and the following general conventions.

For an integer $m\ge1$, write $[m]=\{1,\ldots,m\}$, and let $\Sigma_m$ be the set of permutations of $[m]$. For finite sets $S,T$ of equal cardinality, $\operatorname{Bij}(S,T)$ is the set of bijections from $S$ to $T$. We use $|S|$ for cardinality and $\1_A$ for the indicator of a set $A$. For real numbers $a,b$, write $a\wedge b=\min\{a,b\}$, $a\vee b=\max\{a,b\}$, and $a_+=\max\{a,0\}$. The notation $\operatorname{conv}(S)$ denotes the convex hull of $S$; $\diam(S)$ and $\dist(S,T)$ denote Euclidean diameter and distance between sets.

For a vector, $\|\cdot\|_2$ is the Euclidean norm. For a measurable scalar function $f$, a probability measure $Q$, a measurable set $A$, and $1\le p<\infty$, define
\[
\|f\|_{L^p(Q)}:=\left(\int |f|^p\,dQ\right)^{1/p},
\qquad
\|f\|_{L^p(A,Q)}:=\left(\int_A |f|^p\,dQ\right)^{1/p}.
\]
The restricted norm is unnormalized: it uses $Q$, not $Q(\cdot\mid A)$. The $L^\infty(Q)$ norm is the essential supremum under $Q$; for continuous functions on the cube, the $L^\infty(P_X)$ norm agrees with the supremum norm because $P_X$ has full support. We use $dy$ for Lebesgue measure on $\R$. For conditional densities, $\|p-q\|_2$ means $\|p-q\|_{L^2(P_X\otimes dy)}$; for scalar convolutions or a fixed covariate, it means $L^2(dy)$. Any other function norm is accompanied by its measure or defined locally.

For conditional densities $p,q$, the integrated Hellinger distance is defined by
\begin{equation}\label{eq:mle-hellinger}
\Hdist^2(p,q):=\frac12\int_{\cX}\int_{\mathbb R}
(\sqrt{p(y\mid x)}-\sqrt{q(y\mid x)})^2\,dy\,dP_X(x).
\end{equation}
With this convention, $0\le\Hdist^2(p,q)\le1$. Every Gaussian mixture density in the model is bounded by $Q_{\max}:=(2\pi\sigma_0)^{-1/2}$, so Lemma~\ref{lem:hell-l2} gives
\begin{equation}\label{eq:mle-L2-Hell}
\|p-q\|_{L^2(P_X\otimes dy)}^2\le8Q_{\max}\Hdist^2(p,q).
\end{equation}
Thus convergence in Hellinger distance also controls the conditional density $L^2$ error used in the inverse inequalities.

For a tuple $G\in\cG_k(\Theta_\beta,\Omega_\eta)$, let $S_G(x)$ and $\pi_{i,G}^{(K)}(x)$ be its active set and weights under the routing convention of Section~\ref{sec:intro-setup}. Its conditional mixing measure is
\begin{equation}\label{eq:topk-mixing-measures}
M_G^{(K)}(x):=\sum_{i\in S_G(x)}\pi_{i,G}^{(K)}(x)\delta_{h(x,\eta_i)},
\end{equation}
where $\delta_t$ is the unit point mass at $t$. In the dense case $K=k$, write
\begin{equation}\label{eq:conditional-measures}
M_G(x):=M_G^{(k)}(x)=\sum_{i=1}^k\pi_{i,G}^{(k)}(x)\delta_{h(x,\eta_i)}.
\end{equation}
The corresponding true mixing measures are $M_{G_*}^{(K)}(x)$ and $M_{G_*}(x)$, obtained by substituting $G_*$ for $G$. These measures describe the scalar mixing laws at a fixed covariate, while the tuples retain expert labels; fitted means may coincide. With convolution in the response variable,
\[
p_G^{(K)}(\cdot\mid x)=\phiSig*M_G^{(K)}(x),
\qquad p_G(\cdot\mid x)=\phiSig*M_G(x),
\]
and the same identities hold for the true model.

For probability measures $\mu,\nu$ on $\R$ with finite first moments, their Wasserstein distance is
\[
W_1(\mu,\nu):=\inf_{\gamma\in\Gamma(\mu,\nu)}
\int_{\R^2}|u-v|\,d\gamma(u,v),
\]
where $\Gamma(\mu,\nu)$ is the set of probability measures with marginals $\mu$ and $\nu$. Expectations and probabilities are taken under the indicated law; $\E_*$ and $\E_*^{(K)}$ denote expectations under $\Pp_*$ and $\Pp_*^{(K)}$, respectively. We write $C(\cX)$ for the continuous real-valued functions on $\cX$, equipped with the supremum norm.

We write $a_n\lesssim b_n$ when $a_n\le Cb_n$ for a fixed constant $C$, $a_n\gtrsim b_n$ for the reverse comparison, and $a_n\asymp b_n$ when both hold. Constants denoted by $c,C>0$ may change between displays; their allowed dependencies are stated with each result. Unless indicated otherwise, they are independent of the sample size and varying separation scales. For a positive deterministic sequence $a_n$, $b_n=\mathcal{O}(a_n)$ means that $|b_n|/a_n$ is eventually bounded, and $Z_n=\mathcal{O}_P(a_n)$ means that $Z_n/a_n$ is bounded in probability under the relevant sampling law. Uniformity over model classes or sequences of truths is asserted only under the stated uniform bounds on constants.

For individual pairs of true experts, write
\begin{equation}\label{eq:functional-separation}
\Delta_{j\ell}:=\|h_j^*-h_\ell^*\|_{L^1(P_X)}.
\end{equation}
Their minimum is $\DeltaSep$, as defined in Section~\ref{sec:intro-setup}. Cluster scales, conditional laws, and recovery losses are introduced with the corresponding results.

\subsection{Universal assumptions}\label{sec:commonassumptions}
The following assumptions concern the parameter spaces and expert family; each result specifies which it requires. Section~\ref{sec:linear} gives verification results for linear experts. Geometry and restriction conditions specific to each recovery target are introduced later.

\begin{assumption}[Compact parameter spaces, bounded experts, and covariate regularity]\label{ass:densecompact}
\begin{enumerate}[label=(\roman*),ref=(\roman*)]
%\textbf{}\\
\item\label{cond:compactspaces} The logit parameter space $\Theta_\beta\subset\R^{d_x}\times\R$ and the expert parameter space $\Omega_\eta\subset\R^{d_\eta}$ are compact.
\item\label{cond:boundedexperts} The expert family is uniformly bounded:
\begin{equation}\label{eq:expert-bound}
\sup_{x\in\cX,\eta\in\Omega_\eta}|h(x,\eta)|\le M<\infty.
\end{equation}
\item\label{cond:analyticexperts} For every $\eta\in\Omega_\eta$, the map $x\mapsto h(x,\eta)$ is real analytic on a connected open neighborhood of $\cX$.
\end{enumerate}
\end{assumption}
Compactness and parameter continuity support finite covering arguments and existence of the dense MLE, while boundedness places all Gaussian means in a common interval \citep{NguyenDemystifying2023,NguyenSoftmax2025}. Compact logits on the cube give $B:=\sup_{x\in\cX,\,\beta\in\Theta_\beta}|\beta_1^\top x+\beta_0|<\infty$ and the dense weight floor
\begin{equation}\label{eq:weight-floor}
\pi_{i,G}^{(k)}(x),\pi_j^*(x)\ge\underline\pi_k:=e^{-2B}/k>0.
\end{equation}
Analyticity ensures that a nonzero expert difference has a null zero set, but does not by itself imply uniform restriction or anti-concentration. General ReLU networks fall outside this analytic framework.

\begin{assumption}[Positive functional separation]\label{ass:densesep}
The quantity $\DeltaSep$ is strictly positive.
\end{assumption}
The true expert functions are distinct, although they may agree at particular covariates. The inverse bounds track $0<\DeltaSep\le2M$ explicitly and allow shrinking separation. Neither the likelihood density rates nor the Top-1 selected-function bound requires this assumption.

\begin{assumption}[Scale-adaptive generalized anti-concentration]\label{ass:denseac}
For every $j\ne\ell$, the scalar variable $h_j^*(X)-h_\ell^*(X)$ has a Lebesgue density $q_{j\ell}$ satisfying
\begin{equation}
\norm{q_{j\ell}}_\infty\le \frac{C_{\rm ac}}{\Delta_{j\ell}}
\label{eq:scale-adaptive-density}
\end{equation}
for a constant $C_{\rm ac}$ independent of the pair and of all separation scales.
\end{assumption}
The normalization by $\Delta_{j\ell}$ requires a uniform density bound as experts approach one another. It implies the small-ball estimate
\begin{equation}\label{eq:smallball-density}
P_X\{|h_j^*(X)-h_\ell^*(X)|\le t\Delta_{j\ell}\}
\le 2C_{\rm ac}t,\qquad t>0,
\end{equation}
used to construct separated good sets in Lemma~\ref{lem:mcgoodset}. The density condition is stronger than this near-zero estimate and is not automatic for analytic experts: $X^2$ for $X\sim\operatorname{Unif}([-1,1])$ has an unbounded density near zero, and a nonzero constant difference has no density. Proposition~\ref{prop:linearac} verifies the condition for homogeneous linear experts.

\begin{assumption}[Differentiability in the expert parameter]\label{ass:expertregularity}
The maps
\[
\eta\mapsto h(\cdot,\eta)\in L^1(P_X)
\quad\text{and}\quad
\eta\mapsto h(\cdot,\eta)\in L^\infty(\cX)
\]
are continuously differentiable on an open neighborhood of $\Omega_\eta$.
\end{assumption}
Differentiability is required in the stated function norms. Together with compactness, its $L^\infty$ part gives the uniform parameter Lipschitz bound and entropy estimate in Lemma~\ref{lem:expert-entropy}, which underlie the likelihood rates. No lower derivative bound is needed for those rates.

\begin{assumption}[Expert parameter identifiability]\label{ass:expertidentifiability}
For $\eta_1,\eta_2\in\Omega_\eta$,
\[
\eta_1\ne\eta_2
\quad\Longrightarrow\quad
\|h(\cdot,\eta_1)-h(\cdot,\eta_2)\|_{L^1(P_X)}>0.
\]
\end{assumption}
This requires injectivity of the individual expert parametrization over $\Omega_\eta$; it does not assert identifiability of the full mixture. Neural-network symmetries can violate it. The functional inverse inequalities and density rates do not require this condition.

\begin{assumption}[Nondegenerate expert parameter derivative]\label{ass:expertnondegenerate}
The derivative in Assumption~\ref{ass:expertregularity} satisfies
\[
m_1:=\inf_{\eta\in\Omega_\eta}\inf_{\|u\|_2=1}
\|D_\eta h(\cdot,\eta)u\|_{L^1(P_X)}>0.
\]
\end{assumption}
This imposes uniform first-order sensitivity of the expert function to parameter perturbations. Injectivity alone does not imply this condition, and the condition does not ensure nonsingularity of the mixture's Fisher information. Linear experts satisfy $m_1\ge b/3$ (Section~\ref{sec:linear}), but the main inverse and likelihood results do not require this condition.

Functional recovery instead uses the quantitative restriction assumptions in Sections~\ref{sec:dense} and \ref{sec:topk} to extend control from informative covariate sets to the full recovery target.

\section{Inverse Bounds for Dense MoE}\label{sec:dense}

In this section, we establish inverse bounds connecting density recovery to expert recovery in dense MoE through the expert separation quantity $\Delta_{\mathrm{sep}}$. We consider two main regimes: (i) the multi-cluster regime, in which the true experts are partitioned into separated clusters, with the single-cluster case included as a specialization; and (ii) the unstructured regime, in which no cluster partition or cluster geometry is prescribed. The latter retains the separation, anti-concentration, and restriction assumptions stated below.

\subsection{Multi-cluster regime}\label{sec:denseclusters}\label{sec:densesingle}\label{sec:densemc}
We first consider a prescribed partition of the true expert functions into $q\in\{1,\ldots,k\}$ clusters. The within-cluster scale is the minimum functional separation $\DeltaSep$, while a separate scale $\DeltaMacro$ describes separation between distinct clusters. The case $q=1$ requires only the within-cluster condition and imposes no localization of the fitted experts.

\begin{definition}[Clustered functional geometry]\label{def:mcgeom}
Let $\mathcal C_1^*,\ldots,\mathcal C_q^*$ be a partition of $[k]$ into nonempty blocks, and put
\[
s_m:=|\mathcal C_m^*|,\qquad \smax:=\max_{m\le q}s_m.
\]
For a fixed $C_0\ge1$, assume
\begin{equation}\label{eq:cluster-functional-diameter}
\|h_j^*-h_\ell^*\|_{L^1(P_X)}\le C_0\DeltaSep,
\qquad j,\ell\in\mathcal C_m^*,\quad m\in[q].
\end{equation}
When $q\ge2$, fix a macroscopic scale $\DeltaMacro>0$ and assume additionally
\[
\|h_j^*-h_\ell^*\|_{L^1(P_X)}\ge\DeltaMacro
\quad(j\in\mathcal C_m^*,\ \ell\in\mathcal C_{m'}^*,\ m\ne m'),
\qquad \DeltaSep\le\frac{\DeltaMacro}{8C_0}.
\]
For $q=1$, Eq.~\eqref{eq:cluster-functional-diameter} is precisely the single-cluster diameter condition, and no $\DeltaMacro$ is specified. Singleton blocks are permitted. These functional conditions are distinguished from the pointwise geometry below.
\end{definition}

Definition~\ref{def:mcgeom} describes clustering after averaging over the covariate distribution. The inverse argument, however, first compares Gaussian mixtures at a fixed covariate $x$, where the expert values are scalar component means. We therefore need to describe how each functional cluster appears in this scalar mixture. For $m\in[q]$, define its set of true means and the interval they span by
\[
\mathcal T_m(x):=\{h_j^*(x):j\in\mathcal C_m^*\},
\qquad I_m^*(x):=\operatorname{conv}\mathcal T_m(x).
\]
The set $\mathcal T_m(x)$ records the means belonging to cluster $m$, while $I_m^*(x)$ records their full range. Using intervals allows us to require that different clusters occupy separate blocks in the scalar ordering: distinct means from different clusters can still interlace, in which case their spanning intervals overlap. For $q\ge2$, define the smallest gap between these intervals by
\[
D_x:=\min_{m\ne m'}\dist(I_m^*(x),I_{m'}^*(x)).
\]
Thus $D_x>0$ ensures that the true cluster intervals are disjoint. This gap also provides a scale for localizing fitted means around their assigned true intervals without allowing fitted clusters to overlap. The following assumption requires the within-cluster spacings to be comparable to $\DeltaSep$ and, when $q\ge2$, the interval gaps to be comparable from below to $\DeltaMacro$, on a set of positive probability. On that same set, it requires a fixed assignment of fitted labels to the true clusters.

\begin{assumption}[Pointwise cluster geometry and stable fitted membership]\label{ass:mccompat}
There are constants $\rho_{\rm cl}\in(0,1]$ and $0<a_{\rm cl}\le A_{\rm cl}<\infty$, uniform over the admissible model pairs and separation scales, and a measurable set $E_{\rm cl}=E_{\rm cl}(G,G_*)\subseteq\cX$ of probability at least $\rho_{\rm cl}$, with the following properties.
\begin{enumerate}[label=(\roman*)]
\item For every $x\in E_{\rm cl}$ and distinct $j,\ell\in\mathcal C_m^*$,
\[
a_{\rm cl}\DeltaSep\le|h_j^*(x)-h_\ell^*(x)|\le A_{\rm cl}\DeltaSep.
\]
\item If $q\ge2$, there is a fixed $d_{\rm cl}>0$ such that, for every $x\in E_{\rm cl}$,
\[
D_x\ge d_{\rm cl}\DeltaMacro,
\qquad A_{\rm cl}\DeltaSep\le d_{\rm cl}\DeltaMacro/8.
\]
\item There is a partition $\widehat{\mathcal C}_1,\ldots,\widehat{\mathcal C}_q$ of the fitted labels, independent of $x\in E_{\rm cl}$, with $|\widehat{\mathcal C}_m|=s_m$. If $q\ge2$, every $i\in\widehat{\mathcal C}_m$ satisfies
\[
\dist(h_i(x),I_m^*(x))<D_x/8,\qquad x\in E_{\rm cl}.
\]
The partition may depend on $(G,G_*)$. For $q=1$, it is simply $\widehat{\mathcal C}_1=[k]$, with no restriction on the fitted locations beyond the common support bound.
\end{enumerate}
\end{assumption}
Assumption~\ref{ass:mccompat} has three roles. The first condition makes the within-cluster scalar spacings comparable to $\DeltaSep$ on $E_{\rm cl}$. The second separates the true cluster intervals at the macroscopic scale. The third assigns exactly $s_m$ fitted labels to each true block, keeps that assignment fixed throughout $E_{\rm cl}$, and, for $q\ge2$, confines each assigned fitted mean to the stated neighborhood of its true interval. Although the fitted partition may depend on the model pair, it cannot change with the covariate on this set. This consistency is necessary for the subsequent construction of a common matching of expert functions.

For $q\ge2$, the compatibility neighborhoods are disjoint, and the full intervals
\[
I_m(x):=\operatorname{conv}\bigl(\mathcal T_m(x)\cup\{h_i(x):i\in\widehat{\mathcal C}_m\}\bigr)
\]
satisfy
\begin{equation}\label{eq:full-cluster-gap}
\min_{m\ne m'}\dist(I_m(x),I_{m'}(x))
\ge3D_x/4\ge3d_{\rm cl}\DeltaMacro/4,\qquad x\in E_{\rm cl}.
\end{equation}
For $q=1$, Lemma~\ref{lem:mcgoodset} constructs the required set from functional clustering and anti-concentration, with $P_X(E_{\rm cl})\ge1/2$, for every fitted tuple in the common model class. For $q\ge2$, the lemma supplies sufficient conditions for the true interval geometry under its additional scale requirement. Stable fitted membership and localization remain separate hypotheses; they do not follow from separation of the true experts alone.

Pointwise recovery on $E_{\rm cl}$ is insufficient for recovery of whole expert functions unless discrepancies on this set control their global norms. The next assumption provides this extension separately for expert and gate errors.

\begin{assumption}[Uniform quantitative restriction]\label{ass:denseR}
Define the expert-discrepancy class
\begin{equation}
\cQ_h
=
\left\{
x\mapsto \pi_j^*(x)\big[h(x,\eta)-h(x,\eta_j^*)\big]:
j\in[k],\ \eta\in\Omega_\eta
\right\}\setminus\{0\}
\label{eq:expert-discrepancy-class}
\end{equation}
and the gating-discrepancy class
\begin{equation}
\cQ_\pi
=
\left\{
x\mapsto\pi_i(x;G)-\pi_j^*(x):
i,j\in[k],\ G\in\mathcal G_k(\Theta_\beta,\Omega_\eta)
\right\}\setminus\{0\}.
\label{eq:gate-discrepancy-class}
\end{equation}
For every fixed $\rho>0$, there is $c_{\rm res}(\rho)>0$, uniform over the admissible true models, such that, for every measurable $A\subset\cX$ with $P_X(A)\ge\rho$,
\begin{equation}
\norm{q}_{L^2(A,P_X)}
\ge c_{\rm res}(\rho)\norm{q}_{L^2(P_X)},\qquad q\in\cQ_h,
\label{eq:expert-restriction}
\end{equation}
and
\begin{equation}
\norm{r}_{L^1(A,P_X)}
\ge c_{\rm res}(\rho)\norm{r}_{L^1(P_X)},\qquad r\in\cQ_\pi.
\label{eq:gate-restriction}
\end{equation}
The inequalities are interpreted trivially for the zero function.
\end{assumption}
Assumption~\ref{ass:denseR} requires any set of probability at least $\rho$ to retain a fixed fraction of each discrepancy's global norm. The expert inequality uses $L^2(P_X)$, matching the covariate norm in the density error, whereas the gate inequality uses $L^1(P_X)$ to measure average absolute probability error. The constants must be uniform over the specified discrepancy classes and admissible truths. Real analyticity alone does not establish this quantitative uniformity for expert differences. For the bounded affine logits used here, Lemma~\ref{lem:affine-gate-restriction} verifies the gate inequality. Verification for linear experts is given in Section~\ref{sec:linear}.

The fixed partitions determine the admissible permutations: define
\[
\Sigma_{\mathcal C}:=\{\sigma\in\Sigma_k:
\sigma(\mathcal C_m^*)=\widehat{\mathcal C}_m\text{ for every }m\},
\qquad N_{\mathcal C}:=|\Sigma_{\mathcal C}|=\prod_{m=1}^q s_m!.
\]
Using the fixed label sets, define on all of $\cX$
\[
\Pi_m(x;G):=\sum_{i\in\widehat{\mathcal C}_m}\pi_i(x),
\qquad \Pi_m^*(x):=\sum_{j\in\mathcal C_m^*}\pi_j^*(x).
\]
A block-preserving permutation satisfies $\sum_{j\in\mathcal C_m^*}\pi_{\sigma(j)}=\Pi_m$, so the aggregate masses are independent of the within-block matching. These functions are defined on all of $\cX$ using the same label blocks, even though fitted localization is required only on $E_{\rm cl}$. For $q=1$, both aggregate masses equal one. For several clusters, their discrepancies require a separate restriction condition.

\begin{assumption}[Uniform restriction for cluster-mass errors]\label{ass:clusterR}
When $q\ge2$, for each fixed $\rho\in(0,1]$ there is $c_{\rm cl}(\rho)>0$, uniform over the admissible model pairs and their fixed partitions, such that for every measurable $A\subseteq\cX$ with $P_X(A)\ge\rho$ and every $m$,
\[
\|\Pi_m(\cdot;G)-\Pi_m^*\|_{L^1(A,P_X)}
\ge c_{\rm cl}(\rho)\|\Pi_m(\cdot;G)-\Pi_m^*\|_{L^1(P_X)}.
\]
The inequality is trivial for a zero error. No additional restriction is required when $q=1$.
\end{assumption}
Assumption~\ref{ass:clusterR} extends control of net cluster-mass errors from an informative set to the full domain. It is imposed separately because a cluster error is a sum of component errors, and cancellation within that sum prevents a direct deduction from the componentwise inequalities in Assumption~\ref{ass:denseR}. This condition is used to retain control of mass transferred between clusters at the macroscopic scale. Lemma~\ref{lem:affine-gate-restriction} verifies it for the bounded affine softmax gates of this paper, uniformly over fixed label blocks.

\begin{definition}[Clustered labelled functional loss]\label{def:Dmc}\label{def:Dfunc}
Set
\[
\omega_q:=\begin{cases}0,&q=1,\\ \DeltaMacro,&q\ge2.\end{cases}
\]
Relative to the true and fitted partitions above, define
\begin{equation}\label{eq:clustered-loss}
\begin{aligned}
\Dmc(G,G_*):=\min_{\sigma\in\Sigma_{\mathcal C}}\Bigg\{&
\sum_{j=1}^k\Bigl[
\|\pi_j^*(h_{\sigma(j)}-h_j^*)\|_{L^2(P_X)}
+\DeltaSep\|\pi_{\sigma(j)}-\pi_j^*\|_{L^1(P_X)}\Bigr]\\
&+\omega_q\sum_{m=1}^q\|\Pi_m(\cdot;G)-\Pi_m^*\|_{L^1(P_X)}\Bigg\}.
\end{aligned}
\end{equation}
For $q=1$, $\Sigma_{\mathcal C}=\Sigma_k$ and this is exactly
\[
\Dfunc(G,G_*):=\min_{\sigma\in\Sigma_k}\sum_{j=1}^k\Bigl[
\|\pi_j^*(h_{\sigma(j)}-h_j^*)\|_{L^2(P_X)}
+\DeltaSep\|\pi_{\sigma(j)}-\pi_j^*\|_{L^1(P_X)}\Bigr].
\]
Thus $\Dfunc$ denotes the one-cluster specialization of $\Dmc$. We also use this unrestricted-permutation formula in the unstructured regime; defining the loss itself does not require the one-cluster diameter condition.
\end{definition}
The structure of these losses can be understood through partial transport along a fixed expert matching. Fix $x$ and an admissible permutation $\sigma$, and write
\[
a_j:=\pi_{\sigma(j)}(x),\qquad b_j:=\pi_j^*(x),\qquad
r_j:=|h_{\sigma(j)}(x)-h_j^*(x)|,\qquad m_j:=\min\{a_j,b_j\}.
\]
The largest mass that can be transported along the matched pair $j\mapsto\sigma(j)$ is $m_j$, at cost $m_jr_j$. Transporting all this common mass leaves $a_j+b_j-2m_j=|a_j-b_j|$ unmatched across the two sides. Charging $\DeltaSep$ per unit of unmatched mass therefore gives the pointwise cost
\[
\sum_{j=1}^k m_jr_j+\DeltaSep\sum_{j=1}^k|a_j-b_j|.
\]
The dense weight floor implies $\underline\pi_k b_j\le m_j\le b_j$, so replacing the common-mass displacement $m_jr_j$ by the true-mass displacement $b_jr_j$ changes this cost by at most a fixed multiplicative factor. This explains the first two terms in Definition~\ref{def:Dmc}: they measure matched expert displacement and residual component-mass discrepancy. The factor $\DeltaSep$ reflects the transport scale for reallocating mass among nearby experts. This interpretation prescribes transport of the maximal common mass under the matching; allowing additional mass to be discarded would instead truncate the displacement cost, as detailed in Appendix~\ref{app:partialot}.

For several clusters, the additional term measures net mass discrepancy at the macroscopic scale $\DeltaMacro$. Redistribution within a cluster can change its component weights while leaving its total mass unchanged. A net transfer between clusters, by contrast, crosses an interval gap of order $\DeltaMacro$ on the informative set. The aggregate term retains control of this discrepancy even as $\DeltaSep$ decreases. It supplements the componentwise penalty; the two terms are not asserted to decompose the cost of an optimal transport plan.

The functional norms and the placement of the minimization specify the recovery target. The $L^2(P_X)$ expert terms match the covariate norm used for density error, while the $L^1(P_X)$ mass terms measure average absolute gate errors. Taking these norms before minimizing forces one permutation to match whole expert functions throughout $\cX$. The positive weight floor then gives control of the unweighted expert errors. Proposition~\ref{prop:partial-functional} makes the common-mass comparison precise for these same functional norms and the cluster correction. For $q=1$, the aggregate mass discrepancy vanishes and all permutations are admissible, yielding $\Dfunc$.

The following theorem combines the geometric and restriction conditions. It uses the restriction assumptions only at level $\rho_0=\rho_{\rm cl}/N_{\mathcal C}$: among the $N_{\mathcal C}$ admissible permutations, at least one attains the smallest pointwise matching cost on a subset of $E_{\rm cl}$ having probability at least $\rho_0$.

\begin{theorem}[Dense clustered functional recovery]\label{thm:densemcfunc}
Suppose the clustered geometry of Definition~\ref{def:mcgeom} and Assumptions~\ref{ass:densecompact}, \ref{ass:densesep}, \ref{ass:denseac}, and \ref{ass:mccompat} hold. Impose Assumption~\ref{ass:denseR} at level
\[
\rho_0:=\frac{\rho_{\rm cl}}{N_{\mathcal C}}
=\frac{\rho_{\rm cl}}{\prod_m s_m!}.
\]
If $q\ge2$, also impose Assumption~\ref{ass:clusterR} at this level. Then
\begin{equation}\label{eq:clustered-inverse}
\|p_G-p_{G_*}\|_{L^2(P_X\otimes dy)}
\ge C_{\rm cluster}\DeltaSep^{2\smax-2}\Dmc(G,G_*).
\end{equation}
Here $C_{\rm cluster}>0$ depends only on $k,M,\sigma_0,B,C_0,C_{\rm ac},\rho_{\rm cl},a_{\rm cl},A_{\rm cl}$ and $c_{\rm res}(\rho_0)$; when $q\ge2$, it may also depend on $\DeltaMacro,d_{\rm cl}$ and $c_{\rm cl}(\rho_0)$. It is independent of $G,G_*$ and $\DeltaSep$ when these constants are uniform.
\end{theorem}
The proof of Theorem~\ref{thm:densemcfunc} is provided in Appendix~\ref{app:densemcfuncproof}. Theorem~\ref{thm:densemcfunc} bounds the clustered labelled loss $\Dmc$ by the conditional density error, with separation factor $\DeltaSep^{-(2\smax-2)}$. The loss compares whole expert functions and component gate probabilities under one cluster-preserving permutation, and also measures aggregate gate discrepancies at the macroscopic scale $\DeltaMacro$. The largest cluster size determines the separation exponent, while the macroscopic gap and the geometry and restriction constants enter $C_{\rm cluster}$. This dependence is uniform in $\DeltaSep$ when those quantities remain fixed.

The proof obtains a pointwise matching on $E_{\rm cl}$, selects one permutation on a subset of probability at least $\rho_0$, and uses restriction to extend control to the full domain. For several clusters, stable fitted membership and localization are hypotheses of the theorem; a small density error alone does not establish them. The bound will be used to establish expert convergence rates in Section~\ref{sec:densemle}.

\begin{corollary}[One-cluster specialization]\label{thm:densemain}
Suppose Assumptions~\ref{ass:densecompact}, \ref{ass:densesep}, and \ref{ass:denseac} hold, and Definition~\ref{def:mcgeom} holds with $q=1$. If Assumption~\ref{ass:denseR} holds at level $1/(2k!)$, then
\[
\|p_G-p_{G_*}\|_{L^2(P_X\otimes dy)}
\ge C_{\rm dense}\DeltaSep^{2k-2}\Dfunc(G,G_*).
\]
Here $C_{\rm dense}>0$ depends only on $k,M,\sigma_0,B,C_0,C_{\rm ac}$ and $c_{\rm res}(1/(2k!))$.
\end{corollary}
For a single cluster, $\smax=k$ and every permutation is admissible. Lemma~\ref{lem:mcgoodset} supplies a set of probability at least $1/2$ on which all true means have spacings comparable to $\DeltaSep$; the restriction level is therefore $1/(2k!)$. This establishes the corollary without a fitted-localization assumption or a cluster-mass restriction condition. The proof is given in Appendix~\ref{appendix:proof_thm:densemain}.

The corollary controls $\Dfunc$ with separation exponent $2k-2$. It applies to every fitted tuple satisfying the common model bounds and the stated uniform restriction condition, including tuples whose means coincide at some covariates. Both the theorem and its corollary concern functional recovery; recovery of the expert parameters requires additional identifiability and inverse regularity of the expert parametrization.

\begin{remark}[Cluster size and the separation exponent]\label{rem:countadapt}
A block containing $s_m$ true and $s_m$ fitted labels contributes at most $2s_m$ interpolation nodes. The first divided difference is controlled by Lipschitz continuity; each subsequent order can incur one inverse power of $\DeltaSep$. Divided differences spanning distinct blocks have denominators bounded below by a fixed gap. The maximal power is therefore $2\smax-2$. The scalar bound in Appendix~\ref{app:clusterRKHS} also allows unequal fitted block counts, although the labelled theorem requires equal counts and stable membership.
\end{remark}

\subsection{Unstructured regime}\label{sec:denseU}
We next consider configurations in which pairwise functional distances need not be comparable to the minimum separation and no cluster partition is prescribed. We retain the unrestricted permutation loss $\Dfunc$ from Definition~\ref{def:Dmc}; its formula is meaningful without the one-cluster diameter condition. A direct scalar density-to-matching inequality yields the same separation exponent $2k-2$ for this loss. Quantitative restriction then converts a pointwise matching into one matching of whole expert functions.

\begin{definition}[Unstructured functional geometry]\label{def:ugeom}
No cluster partition or upper diameter bound proportional to $\DeltaSep$ is prescribed. Boundedness, positive functional separation, and anti-concentration are retained.
\end{definition}
The direct density argument is necessary because a uniform comparison of labelled matching with Wasserstein distance can fail. For example, let $0<\delta<1/2$ and consider
\[
\mu_*:=\tfrac13\delta_0+\tfrac13\delta_\delta+\tfrac13\delta_1,
\qquad
\mu:=\tfrac23\delta_0+\tfrac16\delta_1+\tfrac16\delta_1.
\]
Here $W_1(\mu,\mu_*)=\delta/3$, while every bijection of the three labels has true-weighted displacement at least $(1-\delta)/3$. This rules out a separation-uniform transport-to-matching comparison for arbitrary configurations, but not a density-to-matching inequality. Lemma~\ref{lem:scalar-labelled} proves the latter: when a true atom lacks a nearby fitted label, a Gaussian RKHS test function directly detects its discrepancy.

We keep conditional transport as an auxiliary quantity. Define
\begin{equation}\label{eq:OT-loss}
w_G(x):=W_1(M_G(x),M_{G_*}(x)),
\qquad \DOT(G,G_*):=\|w_G\|_{L^2(P_X)}.
\end{equation}
The coupling in $w_G(x)$ may split mass and vary with $x$, so $\DOT$ itself does not specify one global label matching. Nevertheless, restriction of this error over the full product class implies restriction of every weighted expert discrepancy, by replacing one true expert at a time while retaining the true gates.

\begin{assumption}[Uniform norm restriction for the Wasserstein error]\label{ass:OTR}
For every fixed $\rho\in(0,1]$ there is $c_{\rm OT}(\rho)>0$ such that, uniformly over $G\in\cG_k(\Theta_\beta,\Omega_\eta)$ and every measurable $A\subseteq\cX$ with $P_X(A)\ge\rho$,
\[
\|w_G\|_{L^2(A,P_X)}
\ge c_{\rm OT}(\rho)\|w_G\|_{L^2(P_X)}.
\]
Here $w_G$ is defined in Eq.~\eqref{eq:OT-loss}, and the inequality is interpreted trivially when $w_G=0$ almost surely. When truths vary, the constants are required to be uniform over the admissible truths.
\end{assumption}
The theorem uses this assumption at $\rho_0=1/(2k!)$. The class-wide quantifier is essential: Lemma~\ref{lem:OT-expert-restriction} applies the assumption to tuples formed by replacing a single true expert with an arbitrary fitted expert. It thereby supplies the expert inequality in Assumption~\ref{ass:denseR}. For bounded affine softmax logits on the cube, the gate inequality follows from the multivariate Tur\'an--Nazarov estimate; Lemma~\ref{lem:affine-gate-restriction} gives the proof. Neither conclusion requires a cluster diameter condition.

\begin{theorem}[Dense unstructured labelled functional recovery]\label{thm:denseU}
Suppose Assumptions~\ref{ass:densecompact}, \ref{ass:densesep}, \ref{ass:denseac}, and \ref{ass:OTR} hold on the full product class $\cG_k(\Theta_\beta,\Omega_\eta)$. Then, for every $G$ in this class,
\begin{equation}\label{eq:unstructured-functional-inverse}
\|p_G-p_{G_*}\|_{L^2(P_X\otimes dy)}
\ge C_{\rm U,func}\DeltaSep^{2k-2}\Dfunc(G,G_*).
\end{equation}
Here $C_{\rm U,func}>0$ depends only on $k,M,\sigma_0,B,C_{\rm ac},d_x$ and $c_{\rm OT}(1/(2k!))$. It is independent of $G,G_*$ and $\DeltaSep$ when these quantities are uniform. No cluster partition, fitted localization, or small-error condition is required.
\end{theorem}
The proof of Theorem~\ref{thm:denseU} is provided in Appendix~\ref{app:denseUproof}. Theorem~\ref{thm:denseU} bounds $\Dfunc$ by the conditional density error with separation factor $\DeltaSep^{-(2k-2)}$. It controls whole expert functions and separation-weighted component gate discrepancies under one global permutation, without prescribing clusters, localizing fitted means, or requiring the error to be small relative to separation. Thus the labelled inverse bound extends beyond configurations whose pairwise distances are comparable to $\DeltaSep$.

The proof establishes direct scalar labelled inversion on a separated set of probability at least $1/2$. One of the $k!$ permutations minimizes the pointwise cost on a subset of probability at least $1/(2k!)$, and restriction extends its discrepancies to the full domain. The constant therefore depends on $c_{\rm OT}(1/(2k!))$, with class-wide transport restriction supplying the expert inequality and bounded affine softmax logits supplying gate restriction.

The same conclusion holds if the expert inequality \eqref{eq:expert-restriction} is imposed directly at $1/(2k!)$ in place of Assumption~\ref{ass:OTR}; the constant then depends on that expert restriction constant. This formulation covers restricted fitted classes that are not closed under single-expert replacement. Alternatively, transport restriction may be imposed on an ambient full product class. Restriction only for the original pair, or only at probability level $1/2$, is insufficient for this argument. The bound will be used to establish expert convergence rates in Section~\ref{sec:densemle}.

\section{Inverse Bounds for Sparse MoE}\label{sec:topk}

This section establishes deterministic inverse inequalities for active expert recovery under Top-$K$ routing. We first partition the covariate domain into regions on which the true and fitted active sets are fixed, and define a loss that matches experts within each region. For $K\ge2$, conditional separation and restriction assumptions yield an inverse bound with separation exponent $2K-2$ on regions of probability at least $\tau$; the remaining regions contribute a remainder controlled by their total probability. We then treat Top-1, for which the selected expert function admits a separation-free bound. Statistical rates follow by combining these inequalities with the likelihood results in Section~\ref{sec:mle}.

\subsection{Routing regions and recovery losses}\label{sec:topk-regions-losses}
We define recovery losses by matching the fitted and true experts on regions where their active sets are fixed. Throughout this subsection, fix $1\le K\le k$ and a model pair $(G,G_*)$ from the sparse model in Section~\ref{sec:intro-setup}.

For $K$-element subsets $S,T\subset[k]$, the joint routing region
\begin{equation}\label{eq:topk-regions}
\mathcal R_{S,T}:=\{x\in\cX:S_{G_*}(x)=S,\ S_G(x)=T\}
\end{equation}
contains the covariates at which the true active labels are $S$ and the fitted active labels are $T$. These regions form a finite measurable partition of $\cX$. Let $\mathcal R_+$ denote the regions of positive probability. For each $C\in\mathcal R_+$, define its probability and conditional covariate law by
\begin{equation}\label{eq:topk-conditional-law}
w_C:=P_X(C),
\qquad P_C(A):=\frac{P_X(A\cap C)}{w_C}.
\end{equation}
Both the partition and its probabilities depend on $(G,G_*)$. Zero-probability regions make no contribution to the losses and are omitted.

On a region $C=\mathcal R_{S,T}$, we compare the two sets of $K$ active experts under a bijection $\sigma:S\to T$. The common logit bound $B$ gives the active-weight floor
\begin{equation}\label{eq:topk-weight-floor}
\pi_{i,G}^{(K)}(x)\ge\underline\pi_K:=e^{-2B}/K
\quad(i\in T),
\qquad
\pi_{j,G_*}^{(K)}(x)\ge\underline\pi_K
\quad(j\in S),\qquad x\in C.
\end{equation}
Using the global separation scale $\DeltaSep$, define the cost of a matching and its minimum by
\begin{equation}\label{eq:topk-cell-cost}
\begin{aligned}
L_C(\sigma)&:=\sum_{j\in S}\Bigl[
\|\pi_{j,G_*}^{(K)}(h_{\sigma(j)}-h_j^*)\|_{L^2(P_C)}+\DeltaSep\|\pi_{\sigma(j),G}^{(K)}-\pi_{j,G_*}^{(K)}\|_{L^1(P_C)}\Bigr],\\
D_C&:=\min_{\sigma\in\operatorname{Bij}(S,T)}L_C(\sigma).
\end{aligned}
\end{equation}
The first term measures expert displacement and the second measures gate-probability error. Both norms use the conditional law $P_C$, so $D_C$ measures error within the region. We next aggregate these regional errors using their original probabilities $w_C$.

\begin{definition}[Top-$K$ recovery loss]\label{def:topD}
The full regionwise recovery loss is
\begin{equation}\label{eq:topk-loss}
\Dtop(G,G_*):=\left(\sum_{C\in\mathcal R_+}w_CD_C^2\right)^{1/2}.
\end{equation}
For $\tau\in(0,1]$, define
\begin{equation}\label{eq:topk-retained-loss}
\begin{aligned}
\mathcal R_\tau&:=\{C\in\mathcal R_+:w_C\ge\tau\},\\
D_{K,\tau}(G,G_*)&:=\left(\sum_{C\in\mathcal R_\tau}w_CD_C^2\right)^{1/2},\\
w_{\rm rare}(\tau)&:=\sum_{C\in\mathcal R_+\setminus\mathcal R_\tau}w_C.
\end{aligned}
\end{equation}
An empty sum is interpreted as zero.
\end{definition}
The structure of $D_C$ separates errors in expert values from errors in their gate probabilities. Weighting an expert difference by the true active probability reflects its contribution to the conditional mixture; the positive floor in Eq.~\eqref{eq:topk-weight-floor} also permits control of the unweighted expert error. The factor $\DeltaSep$ expresses component-mass discrepancy at the separation scale, as in the dense matching loss. The expert terms use $L^2(P_C)$ to align with the density norm in the inverse inequality, while the gate terms use $L^1(P_C)$ to measure average absolute probability error. Taking these norms before minimizing requires one matching throughout each region, although different regions may use different matchings.

The full loss $\Dtop$ measures recovery over all active regions, with each region weighted by its probability of being observed. These weights convert conditional errors back to the original covariate distribution: for any square-integrable $u$,
\[
\|u\|_{L^2(C,P_X)}^2=w_C\|u\|_{L^2(P_C)}^2.
\]
The square root of the weighted sum of $D_C^2$ therefore has the same aggregation structure as the global density $L^2$ norm. This permits the regional inverse inequalities to be combined directly. The resulting target controls active experts under regional matchings; it does not impose a common permutation of all $k$ expert functions.

The retained loss $D_{K,\tau}$ uses the same structure but includes only regions with $w_C\ge\tau$. It isolates the part of the recovery problem on which uniform conditional separation and restriction assumptions are imposed. The threshold does not alter the estimator or renormalize the retained probabilities. Lowering $\tau$ includes more regions and reduces the omitted probability $w_{\rm rare}(\tau)$, but can weaken the conditional inverse constants. Boundedness controls the remaining regional errors, whose quadratic aggregation produces the $\sqrt{w_{\rm rare}(\tau)}$ remainder in Theorem~\ref{thm:rare}. Thus $D_{K,\tau}$ supports the inverse analysis on retained regions, while $\Dtop$ records recovery across the full routing partition.

\subsection{Recovery bounds on routing regions}\label{sec:topk-recovery}
We now establish recovery bounds on the retained routing regions, quantify the contribution of rare regions, and derive the separation-free Top-1 specialization. For $K\ge2$, the argument applies scalar Gaussian inversion and a matching comparison under each conditional law $P_C$, then uses restriction inequalities to extend control across the region.

Fix $2\le K\le k$ and $\tau\in(0,1]$. The following constants are uniform over the admissible model pairs and all their retained regions, and independent of $\DeltaSep$; they may depend on $\tau$.

\begin{assumption}[Separation on retained routing regions]\label{ass:topR1}
There are constants $0<a_\tau\le A_\tau<\infty$ and $\rho_\tau\in(0,1]$ such that every $C=\mathcal R_{S,T}\in\mathcal R_\tau$ contains a measurable set $E_C$ with
\begin{equation}\label{eq:topk-good-set}
P_C(E_C)\ge\rho_\tau,
\end{equation}
on which, for $P_C$-almost every $x$,
\begin{equation}\label{eq:topk-cell-geometry}
a_\tau\DeltaSep\le|h_j^*(x)-h_\ell^*(x)|\le A_\tau\DeltaSep,
\qquad j,\ell\in S,\quad j\ne\ell.
\end{equation}
\end{assumption}
Assumption~\ref{ass:topR1} places the active true means at spacings comparable to $\DeltaSep$ on a subset of conditional probability at least $\rho_\tau$. The lower bound permits Gaussian inversion, and the upper bound permits comparison of transport with a labelled matching. Only the active experts on each retained region must satisfy this geometry; a single cluster containing all $k$ experts is unnecessary.

\begin{assumption}[Uniform restriction on retained routing regions]\label{ass:topR2}
There are constants $c_{h,\tau},c_{\pi,\tau}>0$ such that, for every $C=\mathcal R_{S,T}\in\mathcal R_\tau$, every measurable $A\subseteq C$ with $P_C(A)\ge\rho_\tau/K!$, and every $j\in S$, $i\in T$,
\begin{equation}\label{eq:topk-expert-restriction}
\|\pi_{j,G_*}^{(K)}(h_i-h_j^*)\|_{L^2(A,P_C)}
\ge c_{h,\tau}\|\pi_{j,G_*}^{(K)}(h_i-h_j^*)\|_{L^2(P_C)},
\end{equation}
and
\begin{equation}\label{eq:topk-gate-restriction}
\|\pi_{i,G}^{(K)}-\pi_{j,G_*}^{(K)}\|_{L^1(A,P_C)}
\ge c_{\pi,\tau}\|\pi_{i,G}^{(K)}-\pi_{j,G_*}^{(K)}\|_{L^1(P_C)}.
\end{equation}
The inequalities include the zero-error case trivially.
\end{assumption}
Assumption~\ref{ass:topR2} ensures that a sufficiently large subset of $C$ retains a fixed fraction of each expert and gate error. The level $\rho_\tau/K!$ arises because one of the $K!$ bijections minimizes the pointwise matching cost on a subset of at least that conditional probability. Restriction then extends its error bounds to all of $C$. The expert inequality requires verification under the conditional law. The gate inequality follows from Lemma~\ref{lem:affine-gate-restriction}, since positive-probability joint regions agree almost everywhere with convex polytopes. Proposition~\ref{prop:conditionalR1} supplies a sufficient condition for Assumption~\ref{ass:topR1}.

\begin{remark}[Dependence on the probability threshold]\label{rem:top-threshold}
Conditioning a global probability bound on $C$ can introduce a factor $w_C^{-1}$. For example, Assumption~\ref{ass:denseac} implies
\[
P_C\{|h_j^*-h_\ell^*|\le t\DeltaSep\}
\le\frac{2C_{\rm ac}t}{w_C}
\le\frac{2C_{\rm ac}t}{\tau},\qquad C\in\mathcal R_\tau.
\]
This explains why the conditional constants may deteriorate as $\tau\downarrow0$. Neither this estimate nor unconditional separation by itself supplies all of Assumptions~\ref{ass:topR1} and \ref{ass:topR2}.
\end{remark}

The following theorem combines these conditional assumptions and aggregates the regionwise bounds using the original probabilities $w_C$.

\begin{theorem}[Top-$K$ inverse bound with rare routing regions]\label{thm:rare}
Let $2\le K\le k$ and $\tau\in(0,1]$. Suppose Assumption~\ref{ass:densecompact}\ref{cond:compactspaces}--\ref{cond:boundedexperts} and Assumption~\ref{ass:densesep} hold. Impose Assumptions~\ref{ass:topR1} and \ref{ass:topR2} on $\mathcal R_\tau$. Then
\begin{equation}\label{eq:topk-inverse-retained}
\|p_G^{(K)}-p_{G_*}^{(K)}\|_{L^2(P_X\otimes dy)}
\ge C_{K,\tau}\DeltaSep^{2K-2}D_{K,\tau}(G,G_*).
\end{equation}
Moreover,
\begin{equation}\label{eq:topk-inverse-robust}
\begin{aligned}
\Dtop(G,G_*)\le{}&C_{K,\tau}^{-1}\DeltaSep^{-(2K-2)}
\|p_G^{(K)}-p_{G_*}^{(K)}\|_{L^2(P_X\otimes dy)}+4KM\sqrt{w_{\rm rare}(\tau)},
\end{aligned}
\end{equation}
and
\begin{equation}\label{eq:topk-rare-mass}
w_{\rm rare}(\tau)\le\min\left\{1,\binom{k}{K}^{\!2}\tau\right\}.
\end{equation}
Here $C_{K,\tau}>0$ depends only on
\[
K,M,\sigma_0,B,a_\tau,A_\tau,\rho_\tau,c_{h,\tau},c_{\pi,\tau},
\]
and is independent of the individual retained-region probabilities $w_C$, of the model pair, and of $\DeltaSep$ when the displayed constants are fixed.
\end{theorem}
The proof of Theorem~\ref{thm:rare} is provided in Appendix~\ref{app:rareproof}. Theorem~\ref{thm:rare} bounds the retained-region loss $D_{K,\tau}$ by the conditional density error with separation factor $\DeltaSep^{-(2K-2)}$. The regional bounds are aggregated using the original probabilities $w_C$, so no further factor depending on an individual retained-region probability is needed. Conditional geometry and restriction enter through the common constant $C_{K,\tau}$. Each matching is fixed within its routing region, but different regions may use different matchings; the result does not provide one global correspondence between all latent experts.

For the full loss $\Dtop$, the theorem adds the remainder $4KM\sqrt{w_{\rm rare}(\tau)}$. Boundedness controls this contribution without separation or restriction assumptions on omitted regions. The finite partition gives $w_{\rm rare}(\tau)\le\binom{k}{K}^2\tau$, while decreasing $\tau$ can weaken the conditional inverse constant. Thus the bound makes explicit both the separation dependence governed by $K$ and the effect of retaining regions with smaller probabilities. These bounds will be used to establish expert convergence rates in Section~\ref{sec:topmle}.

\begin{remark}[An explicit inverse constant]\label{rem:tauconstant}
Let $p_0:=\underline\pi_K=e^{-2B}/K$ and $c_{\rm res,\tau}:=\min\{c_{h,\tau},c_{\pi,\tau}\}$. A valid choice of the matching constant in Lemma~\ref{lem:assign}, with $m=K$, is
\begin{equation}\label{eq:assign-constant-bound}
C_A(a,A):=\max\left\{
\frac{6}{5p_0}+\frac{8}{3a},\;
\frac{1}{p_0}+\frac{8(A+2)}{p_0a}
\right\}.
\end{equation}
If $c_{\rm sc}=c_{\rm sc}(K,M,\sigma_0,p_0)>0$ is the constant of Lemma~\ref{lem:scalar}, then one may take
\begin{equation}\label{eq:tauconstant-expanded}
C_{K,\tau}:=\frac{c_{\rm sc}a_\tau^{2K-2}c_{\rm res,\tau}}
{2K\,C_A(a_\tau,A_\tau)}.
\end{equation}
The restriction constants are evaluated at the relative probability level $\rho_\tau/K!$. Decreasing $\tau$ retains more regions but can weaken their geometry and restriction bounds. Formula~\eqref{eq:tauconstant-expanded} makes this dependence available for the threshold calculation in Section~\ref{sec:topmle}.
\end{remark}

When $K=1$, no comparison between simultaneously active experts is needed. Each conditional response law is a single Gaussian, so we can control the selected function directly over the full covariate domain, without a probability threshold or either conditional assumption. Denote the selected labels by $i_G(x)$ and $i_*(x)$. The active weights equal one, so
\[
p_G^{(1)}(y\mid x)=\phiSig(y-h_{i_G(x)}(x)),
\qquad p_{G_*}^{(1)}(y\mid x)=\phiSig(y-h_{i_*(x)}^*(x)).
\]
Define the selected-function loss
\begin{equation}\label{eq:Dsel}
D_{\rm sel}(G,G_*):=\|h_{i_G(\cdot)}(\cdot)-h_{i_*(\cdot)}^*(\cdot)\|_{L^2(P_X)}.
\end{equation}
\begin{corollary}[Separation-free Top-1 recovery]\label{cor:top1}
Under Assumption~\ref{ass:densecompact}\ref{cond:boundedexperts},
\begin{equation}\label{eq:top1-equivalence}
c_-D_{\rm sel}(G,G_*)
\le\|p_G^{(1)}-p_{G_*}^{(1)}\|_{L^2(P_X\otimes dy)}
\le c_+D_{\rm sel}(G,G_*).
\end{equation}
Here $0<c_-\le c_+<\infty$ depend only on $M,\sigma_0$.
\end{corollary}
The proof of Corollary~\ref{cor:top1} is provided in Appendix~\ref{app:top1proof}.
Corollary~\ref{cor:top1} shows that conditional density error and the $L^2(P_X)$ error of the selected expert function are equivalent up to constants depending only on $M$ and $\sigma_0$. With one active expert, the conditional law is a single Gaussian, so no separation of mixture components is required. Each joint region has only one bijection and no gate error, giving $\Dtop=D_{\rm sel}$.

Write $f_G(x):=h_{i_G(x)}(x)$ for the function selected by a tuple $G$, and $f_{G_*}(x):=h_{i_*(x)}^*(x)$ for its true counterpart. Under the likelihood hypotheses of Theorem~\ref{thm:topmle} for $K=1$, including existence of a measurable global MLE, a fixed architecture gives
\begin{equation}\label{eq:top1-selected-expert-rate}
\|f_{\widehat G_n^{(1)}}-f_{G_*}\|_{L^1(P_X)}
\le D_{\rm sel}(\widehat G_n^{(1)},G_*)
=\mathcal O_P\!\left(\sqrt{\frac{\log n}{n}}\right).
\end{equation}
The key finding is that Top-1 removes the separation penalty for recovery of the selected expert function. Under the stated likelihood conditions and fixed model bounds, its mean absolute error has rate $\sqrt{\log n/n}$ even as true experts approach one another or coincide, without a routing-region threshold or quantitative restriction assumption. This matters because selecting one expert eliminates the need to distinguish several simultaneously active Gaussian components: density accuracy directly controls the function used at each input. The result provides a statistical reason to consider Top-1 when this selected function is the recovery target. Recovering every latent expert over the full domain remains a different objective, requiring information about inactive experts and consistency of their label matching across regions.

\section{From Inverse Bounds to Statistical Rates for MLE}\label{sec:mle}

We now combine likelihood density estimation with the inverse inequalities of Sections~\ref{sec:dense} and \ref{sec:topk} to obtain statistical recovery rates. Section~\ref{sec:densemle} treats dense softmax models, and Section~\ref{sec:topmle} treats hard Top-$K$ routing. In each case, we first bound the density error and then identify the additional geometric conditions needed to control the corresponding recovery loss.

Throughout this section, assume the iid, correctly specified sampling model of Section~\ref{sec:intro-setup} and Assumptions~\ref{ass:densecompact} and \ref{ass:expertregularity}. Use the MLEs defined there and write $\widehat p_n:=p_{\widehat G_n}$ and $\widehat p_n^{(K)}:=p_{\widehat G_n^{(K)}}^{(K)}$ for their fitted densities. The density rates require no separation assumptions; recovery requires the relevant inverse hypotheses uniformly over the fitted class. Unless stated otherwise, the architecture is fixed.

\subsection{Density estimation and expert recovery in dense MoE}\label{sec:densemle}

We first derive covering bounds for the expert and logit classes and use them to establish a finite-sample density rate for the dense MLE. We then apply the inverse inequalities to obtain recovery rates for globally matched expert functions under clustered and unstructured geometries. Both transfers are direct deterministic comparisons; neither requires an additional localization event for the estimator.

\paragraph{Complexity of the expert and logit classes.}

Let $N(\varepsilon,\mathscr F,\|\cdot\|_\infty)$ denote the covering number of a function class $\mathscr F$ in the uniform norm. Set $d_h:=d_\eta$ and
\[
\mathscr H:=\{h(\cdot,\eta):\eta\in\Omega_\eta\},
\qquad
\mathscr L:=\{x\mapsto\beta_1^\top x+\beta_0:(\beta_1,\beta_0)\in\Theta_\beta\}.
\]

\begin{lemma}[Uniform entropy bounds from parameter regularity]\label{lem:expert-entropy}
Under Assumptions~\ref{ass:densecompact} and \ref{ass:expertregularity}, for every $0<\varepsilon\le1$,
\begin{equation}\label{eq:expert-entropy}
\log N(\varepsilon,\mathscr H,\|\cdot\|_\infty)
\le d_h\log(A_h/\varepsilon),
\end{equation}
and
\begin{equation}\label{eq:logit-entropy}
\log N(\varepsilon,\mathscr L,\|\cdot\|_\infty)
\le(d_x+1)\log(A_\ell/\varepsilon).
\end{equation}
Here $A_h,A_\ell\ge e$ depend only on the fixed parameter sets, the covariate domain, and the expert regularity bounds.
\end{lemma}

The proof of Lemma~\ref{lem:expert-entropy} is provided in Appendix~\ref{app:entropyproof}.
Assumption~\ref{ass:expertregularity} yields a uniform Lipschitz bound in the expert parameter on the compact parameter set, so a finite parameter net supplies a uniform function cover. Convexity of the parameter set is unnecessary. The likelihood arguments below use the resulting entropy bound and therefore also apply to other bounded, covariate-continuous expert classes satisfying Eq.~\eqref{eq:expert-entropy}, with the corresponding exponent $d_h$.

\paragraph{Likelihood density rate.}
For the dense log-likelihood $\ell_n(G):=\sum_{a=1}^n\log p_G(Y_a\mid X_a)$, parameter continuity and compactness ensure attainment of the maximum. Lemma~\ref{lem:dense-mle-existence} supplies the measurable selection $\widehat G_n$ used below. The product cover for the $k$ experts and logits has dimension exponent

\begin{equation}
d_{\rm den}=k(d_h+d_x+1).
\label{eq:dense-model-dimension}
\end{equation}

\begin{theorem}[Finite-sample dense density rate]\label{thm:densemle}
Under Assumptions~\ref{ass:densecompact} and \ref{ass:expertregularity}, and the dense sampling model of Section~\ref{sec:intro-setup}, let $\widehat p_n$ be the dense global MLE.
For every $n\ge2$ and $t\ge1$,
\begin{equation}
\Pp_*\left\{
\Hdist^2(\widehat p_n,p_{G_*})
>
C\frac{d_{\rm den}\log(Cn)+t}{n}
\right\}
\le e^{-t}.
\label{eq:dense-density-rate}
\end{equation}
Here $C>0$ depends only on the fixed model class. Consequently,
\begin{equation}
\Hdist(\widehat p_n,p_{G_*})
=\mathcal{O}_P\left(\sqrt{\frac{d_{\rm den}\log n}{n}}\right),
\label{eq:dense-hellinger-rate}
\end{equation}
and the same rate holds for $\norm{\widehat p_n-p_{G_*}}_{L^2(P_X\otimes dy)}$.
\end{theorem}

The proof of Theorem~\ref{thm:densemle} is provided in Appendix~\ref{app:densemleproof}.
The theorem gives a finite-sample bound with an exponentially decreasing tail in $t$. For fixed architecture, both Hellinger and density $L^2$ errors have order $\sqrt{\log n/n}$. The proof compares likelihoods through a finite collection of pointwise upper functions, rather than a local parameter expansion, and remains valid at expert collisions without nonsingularity of the Fisher information.

\paragraph{Recovery under the dense inverse hypotheses.}
The inverse hypotheses must hold uniformly over the fitted class so that the bound applies to the data-dependent MLE. If necessary, one may restrict the likelihood class, provided it contains the truth. Its covering bounds remain valid, and compactness of the restricted class preserves dense MLE existence. Write
\[
\varepsilon_{n,\rm den}:=\sqrt{\frac{d_{\rm den}\log(Cn)}{n}}.
\]
To obtain clustered recovery, impose the hypotheses of Theorem~\ref{thm:densemcfunc} uniformly over the fitted class. In particular, each candidate must satisfy the required fitted cluster compatibility and restriction conditions; the density bound alone does not establish them. For unstructured labelled recovery, impose instead the hypotheses of Theorem~\ref{thm:denseU}, including class-wide Wasserstein restriction on the ambient product class, or use the direct expert-restriction formulation described after that theorem. For a nonempty set of admissible permutations $\Sigma\subseteq\Sigma_k$, define
\begin{equation}\label{eq:dense-matched-expert-error}
\mathcal E_\Sigma(G,G_*):=
\min_{\sigma\in\Sigma}\max_{j\in[k]}
\|h_{\sigma(j)}-h_j^*\|_{L^1(P_X)}.
\end{equation}
This error uses one permutation for all experts and all covariates. In the clustered setting, take $\Sigma=\Sigma_{\mathcal C}$ for the partitions associated with the model pair; without prescribed clusters, take $\Sigma=\Sigma_k$.

\begin{corollary}[Dense likelihood recovery across geometries]\label{cor:densemle-recovery}
Assume the hypotheses of Theorem~\ref{thm:densemle}. If the indicated inverse hypotheses hold uniformly over the fitted class, then:
\begin{enumerate}[label=(\roman*)]
\item Under the hypotheses of Theorem~\ref{thm:densemcfunc},
\[
\Dmc(\widehat G_n,G_*)
=\mathcal{O}_P\bigl(C_{\rm cluster}^{-1}\DeltaSep^{-(2\smax-2)}\varepsilon_{n,\rm den}\bigr).
\]
Consequently, under one cluster-preserving matching,
\begin{equation}\label{eq:dense-clustered-expert-rate}
\begin{aligned}
\mathcal E_{\Sigma_{\mathcal C}}(\widehat G_n,G_*)
&\le\underline\pi_k^{-1}\Dmc(\widehat G_n,G_*)\\
&=\mathcal O_P\!\left(
\underline\pi_k^{-1}C_{\rm cluster}^{-1}
\DeltaSep^{-(2\smax-2)}\varepsilon_{n,\rm den}\right).
\end{aligned}
\end{equation}

\item Under the hypotheses of Theorem~\ref{thm:denseU},
\begin{equation}\label{eq:dense-unstructured-functional-rate}
\Dfunc(\widehat G_n,G_*)
=\mathcal O_P\!\left(C_{\rm U,func}^{-1}
\DeltaSep^{-(2k-2)}\varepsilon_{n,\rm den}\right).
\end{equation}
Consequently, under one unrestricted global matching,
\begin{equation}\label{eq:dense-unstructured-expert-rate}
\begin{aligned}
\mathcal E_{\Sigma_k}(\widehat G_n,G_*)
&\le\underline\pi_k^{-1}\Dfunc(\widehat G_n,G_*)\\
&=\mathcal O_P\!\left(\underline\pi_k^{-1}C_{\rm U,func}^{-1}
\DeltaSep^{-(2k-2)}\varepsilon_{n,\rm den}\right).
\end{aligned}
\end{equation}
No cluster diameter, fitted membership, or small-error condition is required. For a restricted likelihood class, impose transport restriction on the ambient full product class, or impose the expert restriction inequality directly at $1/(2k!)$ as described after Theorem~\ref{thm:denseU}; in the latter formulation, replace the restriction dependence in $C_{\rm U,func}$ accordingly.

\end{enumerate}
These bounds also apply to sequences of truths with shrinking separation, provided the other model, geometry, and restriction constants remain uniform. In part~(ii), this includes the restriction constant at probability level $1/(2k!)$.
\end{corollary}

The proof of Corollary~\ref{cor:densemle-recovery} is provided in Appendix~\ref{app:mletransfer}.
In both cases, the expert rate follows from a deterministic loss comparison for every admissible model pair. For one cluster, $\Dmc=\Dfunc$ and $\smax=k$, giving separation penalty $\DeltaSep^{-(2k-2)}$. For several clusters, the largest block determines the displayed exponent, with the macroscopic gap and other inverse constants held fixed. The unstructured result has exponent $2k-2$ without prescribing clusters; it obtains a global matching through direct scalar labelled inversion and restriction at $1/(2k!)$.

For fixed positive separation and fixed architecture, both expert rates have order $\sqrt{\log n/n}$ under their respective hypotheses. With $\DeltaSep=n^{-\gamma}$ and the other constants uniform, $\gamma<1/(4k-4)$ is sufficient for consistency of $\Dfunc$ and of the maximum matched-expert error in part~(ii). No additional requirement that the transport error be $o_P(\DeltaSep)$ is needed. The unscaled component gate errors, whose loss coefficients are $\DeltaSep$, require the stronger condition $\DeltaSep^{-(2k-1)}\varepsilon_{n,\rm den}\to0$ for consistency under this bound.

The corollary isolates the two contributions to the recovery rate: $\varepsilon_{n,\rm den}$ describes density estimation, while the inverse factor describes deterioration as experts approach one another. Uniformity of the remaining constants is essential when applying this interpretation to sequences with shrinking separation.

\subsection{Density estimation and expert recovery in sparse MoE}\label{sec:topmle}

We establish a density rate for the Top-$K$ MLE by combining a count of realizable routing patterns with covers of the expert and active-weight functions. For $K\ge2$, we transfer this rate to recovery on retained routing regions and choose a probability threshold that balances the inverse bound with the contribution of rare regions. We conclude with the separation-free selected-function rate for Top-1.

\paragraph{Routing complexity.}
Near a routing tie, a small perturbation of the slope parameters can change the active set, so a uniform parameter cover need not approximate the routed density. We therefore keep routing exact and count its active-set patterns on finitely many covariates, while using the covers from Section~\ref{sec:densemle} for the remaining smooth quantities.

Define

\begin{equation}
q_k=\binom{k}{2},
\qquad
d_R=kd_x\,\1_{\{K<k\}},
\label{eq:routing-parameter-dimension}
\end{equation}

and

\begin{equation}
d_S=kd_h+k(d_x+1)\1_{\{K\ge2\}}.
\label{eq:sparse-smooth-dimension}
\end{equation}

Here $d_R$ counts routing parameters, and $d_S$ is the dimension exponent for the smooth covers. For $K=1$, the active weight is one and no logit cover is needed; for $K=k$, routing is constant and $d_R=0$. For covariates $x_1,\ldots,x_m$, let

\begin{equation}
\Gamma_K(m)=\sup_{x_{1:m}}
\left|
\left\{
(S_G(x_1),\ldots,S_G(x_m)):G
\right\}
\right|.
\label{eq:active-pattern-count}
\end{equation}

\begin{lemma}[Polynomial active-set pattern growth]\label{lem:routingtrace}
Under the routing convention of Section~\ref{sec:topk}, for every $m\ge1$ and $1\le K<k$,
\begin{equation}
\Gamma_K(m)\le(1+2mq_k)^{kd_x}.
\label{eq:routingtrace}
\end{equation}
For $K=k$, $\Gamma_K(m)=1$.
\end{lemma}

The proof of Lemma~\ref{lem:routingtrace} is provided in Appendix~\ref{app:routingtraceproof}.

The bound counts only patterns realizable by a common router. Each comparison $(\beta_{1i}-\beta_{1j})^\top x_a$ defines a hyperplane in router-parameter space, and counting all faces of the resulting arrangement includes routing ties. Thus the number of patterns grows polynomially with $m$, allowing routing complexity to enter the likelihood rate through its logarithm.

\paragraph{Likelihood density rate.}
Let $\ell_n^{(K)}(G):=\sum_{a=1}^n\log p_G^{(K)}(Y_a\mid X_a)$. As in Section~\ref{sec:intro-setup}, assume a measurable global maximizer exists and write

\begin{equation}\label{eq:top-mle}
\widehat G_n^{(K)}\in\operatorname*{arg\,max}_{G\in\cG_k(\Theta_\beta,\Omega_\eta)}\ell_n^{(K)}(G),
\qquad \widehat p_n^{(K)}:=p_{\widehat G_n^{(K)}}^{(K)}.
\end{equation}

For $K<k$, this existence condition is substantive: discontinuities at ties mean that compactness alone does not ensure attainment. For $K=k$, take $\widehat G_n^{(k)}=\widehat G_n$, recovering the dense estimator.

With $A=\max\{A_h,A_\ell,e\}$, combine the smooth-cover and routing-pattern contributions in

\begin{equation}
\mathfrak E_{n,K}
=
1+d_S\log(An^2)+d_R\log(1+4nq_k).
\label{eq:top-complexity}
\end{equation}

\begin{theorem}[Finite-sample Top-$K$ density rate]\label{thm:topmle}
Impose Assumptions~\ref{ass:densecompact} and \ref{ass:expertregularity}, and the Top-$K$ sampling model of Section~\ref{sec:intro-setup}. Assume that a measurable global maximizer $\widehat G_n^{(K)}$ satisfying Eq.~\eqref{eq:top-mle} exists. For every $n\ge2$ and $t\ge1$,
\begin{equation}
\Pp_*^{(K)}\left\{
\Hdist^2(\widehat p_n^{(K)},p_{G_*}^{(K)})
>
C\frac{\mathfrak E_{n,K}+t}{n}
\right\}
\le3e^{-t}.
\label{eq:top-density-rate}
\end{equation}
Here $C>0$ depends only on the fixed model class.
\end{theorem}

The proof of Theorem~\ref{thm:topmle} is provided in Appendix~\ref{app:topmleproof}.
The density rate requires no expert separation, positive routing margin, or lower bound on routing-region probabilities. With $d_h=d_\eta$, it implies

\begin{equation}
\Hdist(\widehat p_n^{(K)},p_{G_*}^{(K)})
=
\mathcal{O}_P\left[
\sqrt{\frac{k(d_x+d_\eta+1)\log(Cnk)}{n}}
\right].
\label{eq:sparse-hellinger-rate}
\end{equation}

The same order holds for the density $L^2$ error by Eq.~\eqref{eq:mle-L2-Hell}. Counting routing slopes and covering smooth logits separately changes only a constant in this dimension bound. For fixed architecture the rate is again $\sqrt{\log n/n}$, despite the discontinuous selection rule. Define the density scale
\[
\varepsilon_{n,K}:=\sqrt{\frac{\mathfrak E_{n,K}}{n}}.
\]

\paragraph{Recovery on retained routing regions.}
For $K\ge2$, apply Theorem~\ref{thm:rare} at a deterministic probability threshold $\lambda\in(0,1]$, corresponding to $\tau$ in Section~\ref{sec:topk}. The additional assumptions are positive functional separation and uniform retained-region geometry and restriction over the fitted class. Uniformity covers the joint regions induced by every candidate and the truth, since the estimator's routing partition is random. The constant $C_{K,\lambda}$ may deteriorate as $\lambda$ decreases. A restricted likelihood class may be used to enforce these conditions if it contains the truth and admits a measurable global maximizer; the same complexity bounds continue to apply.

To state the expert errors, choose the first minimizer $\sigma_C$ of $L_C$ in a fixed ordering of the bijections on each positive-probability joint region and define
\begin{equation}\label{eq:topk-retained-expert-error}
\mathcal E_{K,\lambda}(G,G_*):=
\sum_{C=\mathcal R_{S,T}\in\mathcal R_\lambda}\sum_{j\in S}
\|h_{\sigma_C(j)}-h_j^*\|_{L^1(C,P_X)}.
\end{equation}
This error sums the matched expert discrepancies over retained regions under the original covariate law. Let $\mathcal E_K(G,G_*)$ denote the same sum over all $C\in\mathcal R_+$. Each matching is fixed within its region and may differ between regions.

\begin{corollary}[Rare-region Top-$K$ likelihood recovery]\label{cor:topmle}
Fix $2\le K\le k$ and a deterministic threshold $\lambda\in(0,1]$. Assume the hypotheses of Theorem~\ref{thm:topmle} and Assumption~\ref{ass:densesep}. Impose Assumptions~\ref{ass:topR1} and \ref{ass:topR2} uniformly over the fitted class at this threshold. Then, for every $n\ge2$ and $t\ge1$, with probability at least $1-3e^{-t}$,
\begin{equation}\label{eq:topmle-recovery-tail}
\begin{aligned}
\Dtop(\widehat G_n^{(K)},G_*)\le{}&
C C_{K,\lambda}^{-1}\DeltaSep^{-(2K-2)}
\sqrt{\frac{\mathfrak E_{n,K}+t}{n}}\\
&+4KM\sqrt{w_{\rm rare}(\lambda;\widehat G_n^{(K)},G_*)}.
\end{aligned}
\end{equation}
In particular, for fixed architecture,
\begin{equation}\label{eq:topmle-recovery-rate}
\Dtop(\widehat G_n^{(K)},G_*)
=\mathcal{O}_P\bigl(C_{K,\lambda}^{-1}\DeltaSep^{-(2K-2)}\varepsilon_{n,K}
+\sqrt\lambda\bigr).
\end{equation}
Under the regional matchings defining Eq.~\eqref{eq:topk-retained-expert-error}, the expert errors satisfy
\begin{equation}\label{eq:topk-retained-expert-rate}
\begin{aligned}
\mathcal E_{K,\lambda}(\widehat G_n^{(K)},G_*)
&\le\underline\pi_K^{-1}D_{K,\lambda}(\widehat G_n^{(K)},G_*)\\
&=\mathcal O_P\!\left(\underline\pi_K^{-1}C_{K,\lambda}^{-1}
\DeltaSep^{-(2K-2)}\varepsilon_{n,K}\right),
\end{aligned}
\end{equation}
and
\begin{equation}\label{eq:topk-full-expert-rate}
\begin{aligned}
\mathcal E_K(\widehat G_n^{(K)},G_*)
&\le\underline\pi_K^{-1}\Dtop(\widehat G_n^{(K)},G_*)\\
&=\mathcal O_P\!\left(\underline\pi_K^{-1}
\left[C_{K,\lambda}^{-1}\DeltaSep^{-(2K-2)}\varepsilon_{n,K}
+\sqrt\lambda\right]\right).
\end{aligned}
\end{equation}
The same statements apply to deterministic threshold sequences $\lambda_n$ when the corresponding inverse assumptions and displayed constants hold at each $\lambda_n$.
\end{corollary}

The proof of Corollary~\ref{cor:topmle} is provided in Appendix~\ref{app:topmletransfer}.
Corollary~\ref{cor:topmle} gives expert convergence rates on both the retained regions and the full routing partition. On retained regions, the density scale $\varepsilon_{n,K}$ is multiplied by the separation penalty $\DeltaSep^{-(2K-2)}$ and the conditional inverse factor $C_{K,\lambda}^{-1}$. For fixed positive separation, fixed architecture, and a fixed threshold with uniform inverse constants, the retained expert errors have order $\sqrt{\log n/n}$. Reducing $K$ lowers the separation exponent, while the total expert count and routing geometry still affect the complexity and inverse constants.

The full-partition bound includes an additional contribution from rare regions. The finite-sample inequality tracks their actual total probability, while Eq.~\eqref{eq:topk-full-expert-rate} uses the uniform remainder $\sqrt\lambda$. No separation or restriction assumptions are imposed on omitted regions. A fixed threshold therefore gives a retained-region convergence guarantee, but the full-partition bound requires a decreasing threshold and control of $C_{K,\lambda}$ to make both terms vanish. The next proposition quantifies this balance.

The expert errors are integrated under the original covariate distribution and concern values where the matched experts are active. The regions and matchings are induced by the estimator and the truth, and uniformity of the hypotheses allows the bounds to apply to this random partition. Regional matchings need not agree across regions, so the corollary does not establish full-domain recovery of every latent expert under one global permutation.

\paragraph{Choosing the probability threshold.}
A fixed threshold yields the retained-region guarantee, but the bound with remainder $\sqrt\lambda$ does not by itself establish consistency over the full partition. To make that remainder vanish, let the threshold decrease while controlling the resulting inverse constant. Write
\[
q_{n,K}:=\DeltaSep^{-(2K-2)}\varepsilon_{n,K}.
\]
The following proposition assumes a polynomial lower bound on $C_{K,\lambda}$ that is uniform over the thresholds used. This assumption quantifies how much conditional recovery can worsen as more regions are retained.

\begin{proposition}[Choice of the retained-region threshold]\label{prop:sieve-rate}
Fix $2\le K\le k$. Suppose the hypotheses of Corollary~\ref{cor:topmle} hold for all sufficiently small deterministic thresholds, with
\begin{equation}\label{eq:C-polynomial}
C_{K,\lambda}\ge c_0\lambda^\alpha
\end{equation}
for constants $c_0>0$ and $\alpha>0$ independent of $n,\lambda$. If $q_{n,K}\to0$, choose
\begin{equation}\label{eq:lambda-opt}
\lambda_n\asymp q_{n,K}^{2/(2\alpha+1)}.
\end{equation}
Then
\begin{equation}\label{eq:sieve-rate}
\Dtop(\widehat G_n^{(K)},G_*)
=\mathcal{O}_P\bigl(q_{n,K}^{1/(2\alpha+1)}\bigr).
\end{equation}
Consequently, under the regional matchings defining $\mathcal E_K$ in Corollary~\ref{cor:topmle}, the summed active expert error over the full routing partition satisfies
\begin{equation}\label{eq:sieve-expert-rate}
\begin{aligned}
\mathcal E_K(\widehat G_n^{(K)},G_*)
&\le\underline\pi_K^{-1}\Dtop(\widehat G_n^{(K)},G_*)=\mathcal O_P\!\left(\underline\pi_K^{-1}
q_{n,K}^{1/(2\alpha+1)}\right).
\end{aligned}
\end{equation}
\end{proposition}

The proof of Proposition~\ref{prop:sieve-rate} is provided in Appendix~\ref{app:mlethreshold}.
Proposition~\ref{prop:sieve-rate} gives convergence of both the full regionwise loss and the summed active expert error, at order $q_{n,K}^{1/(2\alpha+1)}$ when the active-weight floor is fixed. The expert error integrates discrepancies over all joint routing regions under the original covariate law. Its matchings remain region-specific, so the conclusion concerns expert values where they are active and does not establish one global matching of all latent expert functions.

Corollary~\ref{cor:topmle} gives bounds at any prescribed deterministic threshold, including a threshold sequence for which its hypotheses hold. Its full-partition bound retains two contributions: the inverse error $C_{K,\lambda}^{-1}q_{n,K}$ and the rare-region remainder $\sqrt\lambda$. Proposition~\ref{prop:sieve-rate} adds uniform inverse hypotheses over sufficiently small thresholds and the polynomial bound \eqref{eq:C-polynomial}. These stronger assumptions permit a threshold choice that balances $q_{n,K}\lambda^{-\alpha}$ with $\sqrt\lambda$, making both contributions vanish when $q_{n,K}\to0$. Thus the proposition specializes the corollary to an explicit full-partition convergence rate. The threshold is used only in the analysis and does not change the MLE.

The exponent $1/(2\alpha+1)$ records the cost of extending recovery to regions with weaker conditional geometry: a larger $\alpha$ gives a slower guaranteed rate. For fixed positive separation and fixed architecture, the full expert-error rate is $(\log n/n)^{1/[2(2\alpha+1)]}$. The threshold choice is oracle when the separation scale or the deterioration of the inverse constant is unknown; the proposition does not supply a data-driven rule or assert that this rate is optimal.

The exponent $\alpha$ must be obtained from the geometry and restriction constants, rather than from the likelihood bound. For example, Eqs.~\eqref{eq:assign-constant-bound} and \eqref{eq:tauconstant-expanded} show that $a_\lambda\asymp\lambda$, $A_\lambda\lesssim\lambda^{-1}$, and a uniform positive lower bound on $c_{\rm res,\lambda}$ give $C_{K,\lambda}\gtrsim\lambda^{2K}$. The resulting rate is $q_{n,K}^{1/(4K+1)}$. If instead $A_\lambda\lesssim1$ under the same other conditions, then $C_{K,\lambda}\gtrsim\lambda^{2K-1}$ and

\begin{equation}\label{eq:linear-sieve-rate}
\lambda_n\asymp q_{n,K}^{2/(4K-1)},
\qquad
\Dtop(\widehat G_n^{(K)},G_*)
=\mathcal{O}_P\bigl(q_{n,K}^{1/(4K-1)}\bigr).
\end{equation}

The same order controls $\mathcal E_K$ through Eq.~\eqref{eq:sieve-expert-rate}. For fixed positive separation and fixed architecture, it becomes $(\log n/n)^{1/[2(4K-1)]}$. Appendix~\ref{app:mlethreshold} gives the inverse-constant calculation; Section~\ref{sec:linear} verifies the geometric scaling and, using Lemma~\ref{lem:affine-gate-restriction}, the required restriction bounds for linear experts.

\paragraph{Top-1 selected-function recovery.}
For $K=1$, combine Theorem~\ref{thm:topmle} directly with Corollary~\ref{cor:top1} to obtain

\begin{equation}\label{eq:top1-mle-rate}
D_{\rm sel}(\widehat G_n^{(1)},G_*)=\mathcal{O}_P(\varepsilon_{n,1}).
\end{equation}
No separation, restriction, or probability-threshold condition is needed for this transfer. The selected-function $L^1(P_X)$ error is bounded by $D_{\rm sel}$, as in Eq.~\eqref{eq:top1-selected-expert-rate}; the target remains the function selected at each input.
Table~\ref{tab:intro-rates} summarizes these rates and the expert-error consequences of the corresponding recovery losses.

\section{Assumptions Verification for Linear Experts}\label{sec:linear}

This section gives concrete sufficient conditions for applying the recovery results to homogeneous linear experts. We first relate coefficient distances to functional separation and verify anti-concentration under the cube design. We then establish restriction for expert errors and give an elementary proof for binary softmax gates; Lemma~\ref{lem:affine-gate-restriction} covers general active counts. Coefficient clustering supplies dense and sparse geometry and an explicit threshold dependence for the sparse inverse constant. Fitted cluster compatibility remains a separate requirement for several-cluster recovery. Unstructured labelled recovery follows using the direct expert-restriction formulation of Theorem~\ref{thm:denseU}.

\subsection{Coefficient separation and anti-concentration}\label{sec:linear-separation}
We begin with two properties of linear projections under the cube design. The first makes Euclidean coefficient distance a proxy for the functional separation scale. The second controls how often a pair of distinct experts can give nearly equal predictions, uniformly as their coefficients approach one another. Take $d_\eta=d_x$ and
\begin{equation}\label{eq:linear-family}
h(x,\eta)=\eta^\top x,\qquad
\Omega_\eta\subseteq\{\eta\in\R^{d_x}:\|\eta\|_2\le R_\eta\},
\end{equation}
where $\Omega_\eta$ is compact, and retain the cube design and compact logit parameter space of the common model. The coefficient bound gives $|h(x,\eta)|\le M:=b\sqrt{d_x}R_\eta$. Linearity also gives analyticity in $x$ and continuous differentiability in $\eta$ in the required function norms. Hence Assumptions~\ref{ass:densecompact} and \ref{ass:expertregularity} hold, with the common logit bound $B$. Let
\[
\delta_\eta:=\min_{j\ne\ell}\|\eta_j^*-\eta_\ell^*\|_2.
\]

\begin{proposition}[Functional and coefficient separation]\label{prop:linearsep}
For every $a\in\R^{d_x}$ under the cube design,
\begin{equation}\label{eq:linear-norm-equivalence}
\frac b3\|a\|_2\le\E|a^\top X|\le\frac b{\sqrt3}\|a\|_2.
\end{equation}
Consequently, for the linear family \eqref{eq:linear-family},
\begin{equation}\label{eq:linear-separation-equivalence}
\frac b3\delta_\eta\le\DeltaSep\le\frac b{\sqrt3}\delta_\eta.
\end{equation}
\end{proposition}
The proof of Proposition~\ref{prop:linearsep} is provided in Appendix~\ref{app:linearsepproof}.
Proposition~\ref{prop:linearsep} shows that positive functional separation is equivalent to distinct true coefficient vectors. Both comparison constants are independent of the particular pair and its distance, so coefficient and functional separation remain comparable along sequences of colliding experts. The same lower bound verifies Assumptions~\ref{ass:expertidentifiability} and \ref{ass:expertnondegenerate}, with $m_1\ge b/3$ because $D_\eta h(\cdot,\eta)a=a^\top x$.

Norm comparability alone does not give the density bound required by Assumption~\ref{ass:denseac}. For a nonzero projection $a^\top X$, conditioning on all but a coordinate with maximal $|a_r|$ leaves a uniform variable on an interval of length $2b|a_r|$. This yields the following bound.

\begin{proposition}[Scale-adaptive anti-concentration]\label{prop:linearac}
For $a\ne0$, the variable $Z=a^\top X$ under the cube design has a density satisfying
\begin{equation}\label{eq:linear-ac}
\|f_Z\|_\infty\le\frac{\sqrt{d_x}}{2b\|a\|_2}
\le\frac{\sqrt{d_x}}{2\sqrt3\,\E|Z|}.
\end{equation}
In particular, distinct true linear coefficients satisfy
\begin{equation}\label{eq:linear-ac-pair}
\|q_{j\ell}\|_\infty\le\frac{\sqrt{d_x}}{2\sqrt3\,\Delta_{j\ell}},
\qquad j\ne\ell.
\end{equation}
\end{proposition}
The proof of Proposition~\ref{prop:linearac} is provided in Appendix~\ref{app:linearacproof}.
Taking $a=\eta_j^*-\eta_\ell^*$ verifies Assumption~\ref{ass:denseac} with $C_{\rm ac}=\sqrt{d_x}/(2\sqrt3)$ whenever $\delta_\eta>0$. The density bound scales with each pair's own functional distance $\Delta_{j\ell}$, while its constant stays fixed as that distance decreases. This supplies anti-concentration at both the within-cluster and between-cluster scales used in Lemma~\ref{lem:mcgoodset}; no common direction for the coefficient differences is required.

\subsection{Restriction of expert and gate errors}\label{sec:linear-restriction}
We next verify that control on a subset of a routing region extends to the whole region. For linear expert differences, this follows from a restriction inequality whose constant is independent of the region's size and shape. Binary softmax errors admit the same reduction; gates with more active labels are handled by the exponential-sum estimate in Lemma~\ref{lem:affine-gate-restriction}.

Under the affine routing scores, every positive-probability joint region agrees with a convex polytope up to a null set. Its conditional law $P_C$ is therefore normalized Lebesgue measure on that polytope. More generally, on any bounded convex body $C$, the degree-one Remez inequality gives, for $1\le p<\infty$ and $P_C(A)\ge\rho>0$,
\begin{equation}\label{eq:aff-remez}
\|L\|_{L^p(A,P_C)}\ge c_{\rm aff}(d_x,p,\rho)\|L\|_{L^p(P_C)}
\end{equation}
for every affine function $L$, with $c_{\rm aff}(d_x,p,\rho)>0$ independent of the volume and shape of $C$ \citep{BrudnyiGanzburg1973}. Thus the constant depends on the relative probability of $A$ within $C$, rather than the absolute probability $P_X(C)$. Appendix~\ref{app:linearrestrictionproof} derives this norm form from the usual supremum estimate.

The active-weight floor transfers Eq.~\eqref{eq:aff-remez} to the weighted expert discrepancies. For binary softmax, each active probability is a logistic function of an affine logit contrast. Bounded logits keep the logistic derivative bounded away from zero on the relevant interval, so the gate discrepancy is comparable to an affine difference. These two observations yield the next proposition.

\begin{proposition}[Restriction for linear experts and binary softmax]\label{prop:linearrestriction}
Assume Eq.~\eqref{eq:linear-family} and the uniform logit bound $B$. On every positive-probability joint Top-$K$ routing region $C=\mathcal R_{S,T}$, for every measurable $A\subseteq C$ with $P_C(A)\ge\rho>0$ and every $i\in T$, $j\in S$,
\begin{equation}\label{eq:linear-expert-restriction}
\begin{aligned}
\|\pi_{j,G_*}^{(K)}(h_i-h_j^*)\|_{L^2(A,P_C)}
&\ge c_h\|\pi_{j,G_*}^{(K)}(h_i-h_j^*)\|_{L^2(P_C)},\\
c_h&=\underline\pi_Kc_{\rm aff}(d_x,2,\rho).
\end{aligned}
\end{equation}
For $K=2$, also
\begin{equation}\label{eq:linear-binary-restriction}
\|\pi_{i,G}^{(2)}-\pi_{j,G_*}^{(2)}\|_{L^1(A,P_C)}
\ge c_\pi\|\pi_{i,G}^{(2)}-\pi_{j,G_*}^{(2)}\|_{L^1(P_C)},
\end{equation}
where $c_\pi>0$ depends only on $B,d_x,\rho$. The expert conclusion holds for dense weights on $\cX$ with $K$ replaced by $k$; the gate conclusion also holds for dense softmax with $k=2$.
\end{proposition}
The proof of Proposition~\ref{prop:linearrestriction} is provided in Appendix~\ref{app:linearrestrictionproof}.
The expert inequality verifies the expert part of Assumption~\ref{ass:topR2} for every active count, and the expert part of Assumption~\ref{ass:denseR} on the full cube. For Top-2, evaluating both inequalities at $\rho=\rho_\tau/2!$ verifies all of Assumption~\ref{ass:topR2}. If $\rho_\tau$ is fixed, these constants remain uniform as $\tau\downarrow0$, even when the retained regions become small or narrow. The proposition also verifies both parts of dense restriction when $k=2$. Separation is not needed for these restriction conclusions; zero discrepancies satisfy the inequalities trivially.

For $K\ge3$, a component gate error need not be comparable to a single affine difference. Cross multiplication instead reduces the question to restriction of an exponential sum. On $C=\mathcal R_{S,T}$, put
\[
Z_T(x):=\sum_{a\in T}e^{\ell_a(x)},\qquad
Z_S^*(x):=\sum_{b\in S}e^{\ell_b^*(x)}.
\]
Then
\begin{equation}\label{eq:linear-gate-numerator}
\begin{aligned}
\pi_{i,G}^{(K)}-\pi_{j,G_*}^{(K)}&=\frac{F_{ij}}{Z_TZ_S^*},\\
F_{ij}&:=\sum_{b\in S}e^{\ell_i+\ell_b^*}
-\sum_{a\in T}e^{\ell_j^*+\ell_a}.
\end{aligned}
\end{equation}
The numerator has at most $2K$ exponential terms, while
$K^2e^{-2B}\le Z_TZ_S^*\le K^2e^{2B}$. A sufficient condition for gate restriction is therefore the uniform estimate
\begin{equation}\label{eq:exp-remez}
\|F\|_{L^1(A,P_C)}\ge c_{\exp}(\rho)\|F\|_{L^1(P_C)},
\qquad P_C(A)\ge\rho,
\end{equation}
uniformly over the relevant numerators and regions, including zero functions trivially. The denominator bounds then give the gate-restriction constant $e^{-4B}c_{\exp}(\rho)$. The same reduction applies to dense component weights with $K=k$.

The multivariate Tur\'an--Nazarov inequality supplies Eq.~\eqref{eq:exp-remez} for a fixed number of exponential terms on every convex routing region \citep[Theorem~3.2]{BrudnyiYomdin2016}; Lemma~\ref{lem:affine-gate-restriction} gives the reduction to the $L^1$ form. Its constant is uniform over region volumes, shapes, and collisions of logit slopes. Thus, together with Proposition~\ref{prop:linearrestriction}, both parts of Assumption~\ref{ass:topR2} hold for every fixed $K$ at a fixed relative probability level. The same result verifies dense component and aggregate gate restriction.

\subsection{Cluster geometry and separation on retained regions}\label{sec:lineargeom}
We now translate clustering of the true coefficients into the geometry used in Sections~\ref{sec:dense} and \ref{sec:topk}. For dense recovery, this verifies the functional cluster scales while leaving fitted compatibility to be checked. For sparse recovery, a single coefficient cluster gives a uniform pointwise diameter bound and an explicit lower separation bound on every retained region. Throughout this subsection, assume $\delta_\eta>0$ and keep the model bounds fixed.

\paragraph{Dense cluster geometry.}
For a partition $\mathcal C_1^*,\ldots,\mathcal C_q^*$ and a fixed $C_\eta\ge1$, suppose
\begin{equation}\label{eq:linear-coefficient-blocks}
\max_m\max_{j,\ell\in\mathcal C_m^*}\|\eta_j^*-\eta_\ell^*\|_2
\le C_\eta\delta_\eta.
\end{equation}
This condition bounds every within-block coefficient distance by a fixed multiple of the minimum distance. Proposition~\ref{prop:linearsep} then gives the within-cluster condition of Definition~\ref{def:mcgeom} with $C_0=\sqrt3C_\eta$. If $q\ge2$ and coefficients in distinct blocks are at least $D_\eta>0$ apart, one may take $\DeltaMacro=bD_\eta/3$. The scale relation
\begin{equation}\label{eq:linear-cluster-scale}
\delta_\eta\le\frac{D_\eta}{24C_\eta}
\end{equation}
ensures $\DeltaSep\le\DeltaMacro/(8C_0)$, completing Definition~\ref{def:mcgeom}.

These comparisons establish functional clustering. For $q=1$, Proposition~\ref{prop:linearac} and Lemma~\ref{lem:mcgoodset} also supply the pointwise geometry needed by Corollary~\ref{thm:densemain}; the gate restriction is as discussed in Section~\ref{sec:linear-restriction}. In particular, for $k=2$ the single-cluster diameter condition is automatic and binary gate restriction is verified.

For $q\ge2$, the sufficient construction of true interval gaps still requires the stronger scale condition specified after Lemma~\ref{lem:mcgoodset}. Fitted localization and stable membership in Assumption~\ref{ass:mccompat} must also hold. Lemma~\ref{lem:affine-gate-restriction} verifies cluster-mass restriction through aggregate numerators with at most $2k^2$ terms. For unstructured labelled recovery, no coefficient clustering or fitted localization is needed: Proposition~\ref{prop:linearrestriction} supplies expert restriction, so the direct expert-restriction version of Theorem~\ref{thm:denseU} applies. This argument does not assert Wasserstein restriction for linear experts; it makes that separate property unnecessary for this version of the labelled conclusion.

\paragraph{Geometry on retained routing regions.}
For sparse recovery, impose the sufficient condition that all true coefficients form one cluster:
\begin{equation}\label{eq:eta-cluster}
\max_{j\ne\ell}\|\eta_j^*-\eta_\ell^*\|_2\le C_\eta\delta_\eta.
\end{equation}
This global condition is stronger than the active-region geometry required by Theorem~\ref{thm:rare}, but makes that geometry verifiable uniformly over the fitted routers. It implies
\begin{equation}\label{eq:function-cluster}
\begin{aligned}
\max_{j\ne\ell}\|h_j^*-h_\ell^*\|_{L^1(P_X)}&\le\sqrt3C_\eta\DeltaSep,\\
\max_{j\ne\ell}\|h_j^*-h_\ell^*\|_\infty&\le3\sqrt{d_x}C_\eta\DeltaSep.
\end{aligned}
\end{equation}
The second inequality bounds the active-support diameter on every routing region, regardless of its probability. Lower separation requires a different argument: conditioning the anti-concentration estimate on a region with probability at least $\tau$ can increase its small-ball probability by a factor $1/\tau$. A union bound over active pairs then gives the following uniform geometry.

\begin{proposition}[Linear separation on retained routing regions]\label{prop:heavylinear}
Assume Eq.~\eqref{eq:linear-family}, $\delta_\eta>0$, and Eq.~\eqref{eq:eta-cluster}. For $2\le K\le k$ and $\tau\in(0,1]$, Assumption~\ref{ass:topR1} holds uniformly over all joint routing regions with probability at least $\tau$, with
\begin{equation}\label{eq:heavy-constants}
\rho_\tau=\frac34,\qquad
 a_\tau=\min\left\{1,\frac{\tau}{4\binom K2\sqrt{d_x/3}}\right\},\qquad
 A_\tau=3\sqrt{d_x}C_\eta.
\end{equation}
If the constants in Assumption~\ref{ass:topR2} are bounded below uniformly in $\tau$, then, with the other model bounds fixed,
\begin{equation}\label{eq:linear-Ctau}
C_{K,\tau}\gtrsim\tau^{2K-1}
\end{equation}
for all sufficiently small $\tau$.
\end{proposition}
The proof of Proposition~\ref{prop:heavylinear} is provided in Appendix~\ref{app:heavylinearproof}.
Proposition~\ref{prop:heavylinear} supplies a good set occupying at least three quarters of each retained region. Its upper spacing constant $A_\tau$ is fixed, while $a_\tau\asymp\tau$ for small thresholds. Thus the guaranteed lower spacing decreases to order $\tau\DeltaSep$ as smaller regions are included. With uniform restriction constants, substitution into Eq.~\eqref{eq:tauconstant-expanded} yields $C_{K,\tau}\gtrsim\tau^{2K-1}$. The exponent accounts for both scalar inversion and the transport-to-matching comparison; it is a sufficient bound for the threshold analysis, not a claim of sharpness.

\paragraph{Consequence for likelihood recovery.}
For Top-2, Proposition~\ref{prop:linearrestriction} applies at the fixed relative level $\rho=\rho_\tau/2!=3/8$, so both restriction constants are uniform in $\tau$. Under Eq.~\eqref{eq:eta-cluster}, this gives $C_{2,\tau}\gtrsim\tau^3$ with no further conditional restriction assumption. Assuming a measurable global Top-2 MLE exists, the likelihood hypotheses are satisfied by the compact linear model. If $q_{n,2}=\DeltaSep^{-2}\varepsilon_{n,2}\to0$, Proposition~\ref{prop:sieve-rate} therefore gives
\[
\lambda_n\asymp q_{n,2}^{2/7},\qquad
D_2(\widehat G_n^{(2)},G_*)=\mathcal{O}_P(q_{n,2}^{1/7}).
\]
For fixed positive separation and fixed architecture, this is $\mathcal{O}_P((\log n/n)^{1/14})$ for the full regionwise loss. The same order controls the summed active expert errors through the comparison in Section~\ref{sec:topmle}. The threshold remains the oracle choice of Section~\ref{sec:topmle}. For every fixed $K\ge2$, Proposition~\ref{prop:linearrestriction} and Lemma~\ref{lem:affine-gate-restriction} give uniform expert and gate restriction at $\rho=3/(4K!)$. Thus, under the same coefficient-cluster and MLE-existence conditions, the general rate \eqref{eq:linear-sieve-rate} follows as well.

The homogeneous form of Eq.~\eqref{eq:linear-family} matters for anti-concentration. Adding an intercept preserves the affine Remez argument, but can produce nonzero constant expert differences without Lebesgue densities. Extensions beyond homogeneous linear experts therefore require separate verification of the density condition.

\section{Discussion}\label{sec:discussion}

The goal of this paper is to understand when an accurate mixture-of-experts model also provides accurate estimates of its individual expert functions. This question matters whenever the experts themselves, and the distinctions between them, are part of the scientific or modeling objective. We study how expert separation and routing sparsity affect recovery from the aggregate conditional distribution. Our analysis concerns Gaussian experts with a known common variance and a correctly specified number of components, and measures recovery directly through errors in the expert functions.

The main contributions describe how the arrangement and activation of experts influence their estimation. For dense models, we show that separated clusters can improve the dependence of the recovery bound on expert separation: the largest cluster governs the displayed separation penalty. We also obtain recovery under one global matching without a prescribed cluster partition. For sparse models, the number of simultaneously active experts determines the separation exponent on sufficiently informative routing regions, while the total probability of omitted regions controls their contribution to the error. Combining these bounds with likelihood density rates gives expert convergence guarantees under the corresponding geometric and restriction assumptions. Verification for linear experts provides concrete settings in which the results apply.

These findings help explain why predictive accuracy alone may give an incomplete account of expert learning. Experts that produce similar outputs can be difficult to distinguish even when their combined distribution is estimated accurately. By making the dependence on separation explicit, the results quantify this source of instability and show how cluster structure can affect it. They also reveal a tradeoff in sparse routing: activating fewer experts reduces the difficulty of distinguishing components at a given input, but changes the information available about each expert across the input domain. Thus the choice of active count should reflect both the recovery objective and where the router exposes the experts to data.

Top-1 gives a particularly clear implication. With one active expert, the conditional response law is a single Gaussian, so density accuracy directly controls the function selected at each input. Under the stated likelihood conditions, this selected function achieves the baseline convergence rate without a penalty for expert separation, even when latent experts approach one another or coincide. This provides a statistical reason to consider Top-1 when recovery of the selected function is the objective. The guarantee concerns that selected function; recovering every expert over its entire domain would additionally require sufficient exposure and a consistent correspondence between fitted and true experts across routing regions. The result therefore informs the choice of routing rule without establishing Top-1 as universally optimal for all expert-recovery tasks.

Several directions follow from this distinction. One is to develop conditions that turn recovery on active regions into recovery of all expert functions under a common matching. Such conditions could clarify how routing coverage and overlap between active sets affect the information available about individual experts. Another is to develop data-driven methods for choosing the probability threshold used in the sparse recovery analysis, and eventually for selecting the active count itself. The present threshold calculation depends on separation and conditional geometry that may be unknown. Empirical studies could examine how well these theoretical quantities explain expert recovery in fitted models and whether they can guide practical routing choices.

Further work should investigate the sharpness and computational scope of the guarantees. Matching lower bounds would determine whether the separation penalties and full-partition rates can be improved. Extending the analysis from global likelihood maximizers to approximate solutions would connect recovery accuracy to the behavior of practical training procedures. Broader model classes also merit study, including an unknown or over-specified number of experts, unequal or unknown variances, and more flexible expert and routing networks. These extensions would test how far the connection between separation, exposure, and functional recovery persists beyond the setting considered here.

% \section*{Declaration of AI Assistance}
% The authors used ChatGPT to assist with manuscript drafting and revision, organization of proofs, and literature searches. The tool also assisted in checking parts of the mathematical arguments. The authors take full responsibility for the mathematical results, proofs, references, and final manuscript.

\clearpage
\appendix
\addtocontents{toc}{\protect\setcounter{tocdepth}{10}}
\begin{center}
{}\textbf{ \Large{Appendices for
``When Does Sparse Routing Help? \vspace{0.5em}\\ The Geometry of Expert Separation in Mixture of Experts''} }
\end{center}
\tableofcontents
\section{Separation Complexity of Gaussian Mixtures}\label{app:scalar}

At each fixed covariate, recovery in the MoE model reduces to inversion of a univariate Gaussian location mixture. This appendix proves the scalar inverse inequalities used in the functional arguments, with explicit dependence on the separation of the true means. We combine Gaussian RKHS identities \citep{BerlinetThomasAgnan2004,Sriperumbudur2010}, Wasserstein duality \citep{Villani2003}, and Newton interpolation \citep{DeBoor2001}.

The proof constructs an RKHS function whose integral against the difference of the mixing measures controls their Wasserstein distance. Interpolation bounds the norm of this function in terms of true-support spacing. Lipschitz continuity controls the first divided difference without a separation factor; the higher orders determine the inverse exponent. For several support clusters, only divided differences within a cluster incur powers of the shrinking separation.

Throughout this appendix, $\phiSig$ is the density of $N(0,\sigma_0)$, with the variance convention in Eq.~\eqref{eq:gaussian-density}. Unsubscripted $L^2$ norms of Gaussian convolutions are taken with respect to $dy$. Supremum and derivative bounds for interpolation are taken on $[-M,M]$ unless another domain is specified. Constants have the dependencies stated in each result and are independent of the varying separation scale.

\subsection{Gaussian convolution identities and RKHS test functions}

The convolution kernel represents the inner product of two Gaussian translates. Its RKHS therefore provides a direct comparison between integrals against a signed mixing measure and the observable density error.

Define
\begin{equation}
K_{\sigma_0}(s,t)
:=\int_{\mathbb R}\phi_{\sigma_0}(y-s)\phi_{\sigma_0}(y-t)\,dy.
\label{eq:convolution-kernel-definition}
\end{equation}
Let $\mathcal H_{\sigma_0}$ be the RKHS of $K_{\sigma_0}$. For a finite signed measure $\nu$ of finite support, write
\[
m_\nu:=\int K_{\sigma_0}(t,\cdot)\,d\nu(t).
\]

\begin{lemma}[Gaussian convolution and kernel-mean identity]\label{lem:rkhsidentity}
For $\sigma_0>0$,
\begin{equation}
K_{\sigma_0}(s,t)
=\kappa_{\sigma_0}\exp\left\{-\frac{(s-t)^2}{4\sigma_0}\right\},
\qquad
\kappa_{\sigma_0}=\frac{1}{2\sqrt{\pi\sigma_0}}.
\label{eq:convolution-kernel-formula}
\end{equation}
For every finite signed measure $\nu$ of finite support,
\begin{equation}
\|m_\nu\|_{\mathcal H_{\sigma_0}}^2
=\|\phi_{\sigma_0}*\nu\|_{L^2(dy)}^2.
\label{eq:kernel-mean-isometry}
\end{equation}
Consequently, for every $F\in\mathcal H_{\sigma_0}$,
\begin{equation}
\left|\int F(t)\,d\nu(t)\right|
\le
\|F\|_{\mathcal H_{\sigma_0}}
\|\phi_{\sigma_0}*\nu\|_{L^2(dy)}.
\label{eq:rkhs-duality-app}
\end{equation}
\end{lemma}

\begin{proof}

By definition,
\[
\begin{aligned}
K_{\sigma_0}(s,t)
&=\frac{1}{2\pi\sigma_0}
\int_{\mathbb R}
\exp\left\{-\frac{(y-s)^2+(y-t)^2}{2\sigma_0}\right\}dy.
\end{aligned}
\]
The elementary identity
\[
(y-s)^2+(y-t)^2
=2\left(y-\frac{s+t}{2}\right)^2+\frac{(s-t)^2}{2}
\]
gives
\[
\begin{aligned}
K_{\sigma_0}(s,t)
&=\frac{e^{-(s-t)^2/(4\sigma_0)}}{2\pi\sigma_0}
\int_{\mathbb R}
\exp\left\{-\frac{(y-(s+t)/2)^2}{\sigma_0}\right\}dy\\
&=\frac{1}{2\sqrt{\pi\sigma_0}}
\exp\left\{-\frac{(s-t)^2}{4\sigma_0}\right\},
\end{aligned}
\]
which is Eq.~\eqref{eq:convolution-kernel-formula}.

Since $\nu$ has finite support, the reproducing property can be applied term by term to give
\begin{equation}
\begin{aligned}
\|m_\nu\|_{\mathcal H_{\sigma_0}}^2
&=\left\langle
\int K_{\sigma_0}(s,\cdot)d\nu(s),
\int K_{\sigma_0}(t,\cdot)d\nu(t)
\right\rangle_{\mathcal H_{\sigma_0}}\\
&=\iint K_{\sigma_0}(s,t)d\nu(s)d\nu(t).
\end{aligned}
\label{eq:kernel-mean-quadratic-form}
\end{equation}
On the other hand,
\begin{equation}
\begin{aligned}
\|\phi_{\sigma_0}*\nu\|_2^2
&=\int_{\mathbb R}
\left\{\int\phi_{\sigma_0}(y-s)d\nu(s)\right\}
\left\{\int\phi_{\sigma_0}(y-t)d\nu(t)\right\}dy\\
&=\iint
\left\{\int\phi_{\sigma_0}(y-s)\phi_{\sigma_0}(y-t)dy\right\}
d\nu(s)d\nu(t)\\
&=\iint K_{\sigma_0}(s,t)d\nu(s)d\nu(t).
\end{aligned}
\label{eq:convolution-quadratic-form}
\end{equation}
Comparing this expression with Eq.~\eqref{eq:kernel-mean-quadratic-form} proves the isometry \eqref{eq:kernel-mean-isometry}.

For $F\in\mathcal H_{\sigma_0}$, the reproducing property gives
\[
\int F(t)d\nu(t)
=\left\langle F,m_\nu\right\rangle_{\mathcal H_{\sigma_0}}.
\]
Cauchy--Schwarz in the Hilbert space and Eq.~\eqref{eq:kernel-mean-isometry} yield Eq.~\eqref{eq:rkhs-duality-app}.
\end{proof}

Interpolation requires explicit RKHS test functions. Gaussian-weighted polynomials form a finite-dimensional class with a norm determined by their coefficients, as the next lemma shows.

\begin{lemma}[Finite-degree Gaussian RKHS test functions]\label{lem:rkhspoly}
Let $D$ be a nonnegative integer and
\begin{equation}
P(t)=\sum_{r=0}^{D}c_rt^r,
\qquad
F_P(t)=e^{-t^2/(4\sigma_0)}P(t).
\label{eq:gaussian-polynomial-test}
\end{equation}
Then $F_P\in\mathcal H_{\sigma_0}$ and
\begin{equation}
\|F_P\|_{\mathcal H_{\sigma_0}}^2
=\kappa_{\sigma_0}^{-1}
\sum_{r=0}^{D}c_r^2(2\sigma_0)^rr!.
\label{eq:gaussian-polynomial-rkhs-norm}
\end{equation}
For $D\le D_{\max}$,
\begin{equation}
\|F_P\|_{\mathcal H_{\sigma_0}}
\le C(D_{\max},\sigma_0)\max_{0\le r\le D}|c_r|
\label{eq:gaussian-polynomial-max-bound}
\end{equation}
and
\begin{equation}
\|F_P\|_{\mathcal H_{\sigma_0}}
\le C(D_{\max},\sigma_0)\sum_{r=0}^D|c_r|.
\label{eq:gaussian-polynomial-sum-bound}
\end{equation}
\end{lemma}

\begin{proof}

Expanding the cross term in Eq.~\eqref{eq:convolution-kernel-formula} gives
\begin{equation}
\begin{aligned}
K_{\sigma_0}(s,t)
&=\kappa_{\sigma_0}
e^{-s^2/(4\sigma_0)}e^{-t^2/(4\sigma_0)}
e^{st/(2\sigma_0)}\\
&=\sum_{r=0}^{\infty}
\varphi_r(s)\varphi_r(t),
\end{aligned}
\label{eq:gaussian-kernel-feature-expansion}
\end{equation}
where
\begin{equation}
\varphi_r(t)
=\sqrt{\kappa_{\sigma_0}}
\frac{t^r e^{-t^2/(4\sigma_0)}}{\sqrt{(2\sigma_0)^rr!}}.
\label{eq:gaussian-kernel-features}
\end{equation}

The feature expansion represents RKHS functions by square-summable coefficient sequences. To verify uniqueness of this representation, multiply $\sum_r a_r\varphi_r(t)$ by $e^{t^2/(4\sigma_0)}$. For $a\in\ell_2$, the resulting power series converges locally uniformly by Cauchy--Schwarz; if it vanishes identically, uniqueness of power-series coefficients gives $a_r=0$ for every $r$. Hence the feature system is orthonormal in the RKHS. Rewriting Eq.~\eqref{eq:gaussian-polynomial-test} as
\[
F_P=\sum_{r=0}^D
c_r\sqrt{\frac{(2\sigma_0)^rr!}{\kappa_{\sigma_0}}}\,\varphi_r
\]
proves both membership and the exact norm formula \eqref{eq:gaussian-polynomial-rkhs-norm}.

Since $D\le D_{\max}$ and the finite collection $\{(2\sigma_0)^rr!:0\le r\le D_{\max}\}$ has a finite maximum, Eqs.~\eqref{eq:gaussian-polynomial-max-bound}--\eqref{eq:gaussian-polynomial-sum-bound} follow immediately.
\end{proof}

% \begin{remark}[Comparison with inverse-heat moment extraction]\label{rem:rkhs-vs-hell}
% Inverse-heat arguments extract moments by transforming a polynomial on the latent support into a test function on the response space. The present argument instead multiplies the interpolation polynomial by a Gaussian factor, placing it directly in the RKHS. Lemma~\ref{lem:rkhsidentity} then bounds its integral by the density $L^2(dy)$ error.
% \end{remark}

\subsection{Interpolation bounds from spacing and smoothness}

The interpolation argument uses two estimates. Separation of every three consecutive nodes controls divided differences of a Lipschitz function; derivative bounds provide an alternative when fitted nodes collide. For a polynomial $P$, write $[t^r]P$ for the coefficient of $t^r$ and $\|P\|_{\mathrm{coef},1}:=\sum_r|[t^r]P|$ for its coefficient norm. These are distinct from function norms on the support interval.

\begin{lemma}[Separated triples control Newton divided differences]\label{lem:tripleDD}
Let $z_1<\cdots<z_N$ lie in $[-M,M]$ and suppose that
\begin{equation}
z_{r+2}-z_r\ge\gamma>0,
\qquad 1\le r\le N-2.
\label{eq:separated-triple-condition}
\end{equation}
Let $g$ be a Lipschitz function on $[-M,M]$, with
\begin{equation}
\|g\|_\infty+\operatorname{Lip}(g)\le L.
\label{eq:interpolation-lipschitz-bound}
\end{equation}
Then, for $1\le q\le N-1$ and $1\le i\le N-q$,
\begin{equation}
|[z_i,\ldots,z_{i+q}]g|
\le C_q L\,(1\wedge\gamma)^{-(q-1)}.
\label{eq:separated-divided-difference-bound}
\end{equation}
If $P$ is the Newton interpolant of $g$ at all $N$ nodes and $2\le N\le N_0$, then
\begin{equation}
\sum_{r=0}^{N-1}|[t^r]P(t)|
\le C(N_0,M)L(1\wedge\gamma)^{-(N-2)}.
\label{eq:separated-interpolant-coefficients}
\end{equation}
\end{lemma}

\begin{proof}

The value bound controls divided differences of order zero. Lipschitz continuity gives the first-order estimate
\begin{equation}
|[z_i,z_{i+1}]g|
=\frac{|g(z_{i+1})-g(z_i)|}{z_{i+1}-z_i}
\le\operatorname{Lip}(g)\le L.
\label{eq:first-divided-difference-bound}
\end{equation}
Thus the first-order estimate is uniform even when two adjacent nodes approach one another.

For $q\ge2$, the Newton recursion is
\begin{equation}
[z_i,\ldots,z_{i+q}]g
=
\frac{[z_{i+1},\ldots,z_{i+q}]g-[z_i,\ldots,z_{i+q-1}]g}{z_{i+q}-z_i}.
\label{eq:divided-difference-recursion}
\end{equation}
The interval from $z_i$ to $z_{i+q}$ contains at least three nodes, hence by Eq.~\eqref{eq:separated-triple-condition}
\begin{equation}
z_{i+q}-z_i\ge z_{i+2}-z_i\ge\gamma.
\label{eq:divided-difference-span}
\end{equation}
Suppose inductively that every divided difference of order $q-1$ is bounded by
\[
C_{q-1}L(1\wedge\gamma)^{-(q-2)}.
\]
Then Eqs.~\eqref{eq:divided-difference-recursion}--\eqref{eq:divided-difference-span} imply
\[
|[z_i,\ldots,z_{i+q}]g|
\le2C_{q-1}L(1\wedge\gamma)^{-(q-1)}.
\]
Starting from Eq.~\eqref{eq:first-divided-difference-bound} proves Eq.~\eqref{eq:separated-divided-difference-bound}.

The Newton form is
\begin{equation}
P(t)=\sum_{q=0}^{N-1}a_q\prod_{r=1}^{q}(t-z_r),
\qquad
a_q=[z_1,\ldots,z_{q+1}]g.
\label{eq:newton-interpolation-form}
\end{equation}
If $Q_q(t)=\prod_{r=1}^q(t-z_r)$, then the coefficient $\ell_1$ norm satisfies
\begin{equation}
\|Q_q\|_{\mathrm{coef},1}
\le\prod_{r=1}^q(1+|z_r|)
\le(1+M)^q.
\label{eq:newton-basis-coefficients}
\end{equation}
Therefore
\begin{equation}
\sum_{r=0}^{N-1}|[t^r]P(t)|
\le\sum_{q=0}^{N-1}|a_q|(1+M)^q.
\label{eq:newton-coefficient-sum}
\end{equation}
By Eq.~\eqref{eq:separated-divided-difference-bound}, the largest inverse power is $(1\wedge\gamma)^{-(N-2)}$, corresponding to $q=N-1$. Since $N\le N_0$, the remaining finite sum is bounded by a constant depending only on $N_0$ and $M$, which proves Eq.~\eqref{eq:separated-interpolant-coefficients}.
\end{proof}

\begin{lemma}[Smooth divided differences without node separation]\label{lem:smoothDD}
Let $g\in C^{N-1}([-M,M])$ and let $z_1<\cdots<z_N$ be arbitrary distinct nodes in $[-M,M]$. Then
\begin{equation}
|[z_i,\ldots,z_{i+q}]g|
\le\frac{\|g^{(q)}\|_\infty}{q!},
\quad 0\le q\le N-1,\quad 1\le i\le N-q.
\label{eq:smooth-divided-difference-bound}
\end{equation}
If, for some $\alpha>0$,
\begin{equation}
\|g\|_\infty\le L_0,
\qquad
\|g^{(q)}\|_\infty\le L_q\alpha^{-(q-1)},\quad 1\le q\le N-1,
\label{eq:interpolation-derivative-condition}
\end{equation}
then the interpolation polynomial $P$ at the $2\le N\le N_0$ nodes satisfies
\begin{equation}
\sum_r|[t^r]P(t)|
\le C(N_0,M,L_0,\ldots,L_{N_0-1})
(1\wedge\alpha)^{-(N-2)}.
\label{eq:smooth-interpolant-coefficients}
\end{equation}
\end{lemma}

\begin{proof}

For distinct nodes, the generalized mean-value theorem for divided differences gives a point $\xi$ in the convex hull of the nodes such that
\begin{equation}
[z_i,\ldots,z_{i+q}]g=\frac{g^{(q)}(\xi)}{q!}.
\label{eq:divided-difference-mean-value}
\end{equation}
This proves Eq.~\eqref{eq:smooth-divided-difference-bound}.

The coefficient bound follows from the Newton expansion and Eqs.~\eqref{eq:newton-interpolation-form}--\eqref{eq:newton-coefficient-sum}. The derivative estimates replace the spacing estimates used in Lemma~\ref{lem:tripleDD}.
\end{proof}

\subsection{Gaussian inversion with separated true support}

\begin{lemma}[Gaussian inverse bound with explicit separation]\label{lem:scalar}
Fix an integer $m\ge2$, constants $M>0$, $\sigma_0>0$, and $p_0\in(0,1/m]$. Let
\begin{equation}
\mu=\sum_{i=1}^{m}a_i\delta_{s_i},
\qquad
\mu_*=\sum_{j=1}^{m}b_j\delta_{t_j}
\label{eq:scalar-mixture-pair}
\end{equation}
be probability measures with
\begin{equation}
a_i,b_j\ge p_0,
\qquad
s_i,t_j\in[-M,M].
\label{eq:scalar-weight-support-bounds}
\end{equation}
The fitted locations $s_i$ are allowed to coincide. Assume that the true locations $t_j$ are distinct and set
\begin{equation}
\delta=\min_{j\ne\ell}|t_j-t_\ell|>0.
\label{eq:scalar-true-spacing}
\end{equation}
Then
\begin{equation}
\|\phi_{\sigma_0}*\mu-\phi_{\sigma_0}*\mu_*\|_{L^2(dy)}
\ge
c_G\delta^{2m-2}W_1(\mu,\mu_*).
\label{eq:scalar-gaussian-inverse}
\end{equation}
Here $c_G>0$ depends only on $m,M,\sigma_0,p_0$.
\end{lemma}

\begin{proof}[Proof of Lemma~\ref{lem:scalar}]
Set
\begin{equation}
\nu=\mu-\mu_*,
\qquad
\bar\delta=1\wedge\delta.
\label{eq:scalar-signed-measure}
\end{equation}
The claim is immediate if $W_1(\mu,\mu_*)=0$, so we assume $W_1(\mu,\mu_*)>0$. By the $W_1$ dual formula on the compact interval $[-M,M]$, there exists a $1$-Lipschitz function $\psi$ such that
\begin{equation}
\int\psi(t)d\nu(t)=W_1(\mu,\mu_*).
\label{eq:wasserstein-dual-witness}
\end{equation}
Because $\nu(\mathbb R)=0$, adding a constant to $\psi$ does not change the integral. Thus, we can normalize
\begin{equation}
\psi(0)=0,
\label{eq:dual-witness-normalization}
\end{equation}
so that $|\psi(t)|\le M$ on $[-M,M]$. Extend $\psi$ to a $1$-Lipschitz function on $\R$; this leaves its values on both supports unchanged.

We distinguish two regimes. Small Wasserstein error forces one fitted label near each true atom, allowing the separated-node interpolation bound. For larger error, smoothing the dual function supplies derivative bounds without requiring fitted separation.

\medskip\noindent\textbf{Step 1: Localization at small transport cost.}
Suppose first that
\begin{equation}
W_1(\mu,\mu_*)<\frac{p_0\delta}{8}.
\label{eq:scalar-small-transport}
\end{equation}
For $j\in[m]$, define
\begin{equation}
I_j=(t_j-\delta/4,t_j+\delta/4).
\label{eq:true-atom-neighborhoods}
\end{equation}
The intervals are pairwise disjoint. We claim that each $I_j$ contains at least one fitted location $s_i$.

If $I_j$ contains no fitted location, then every coupling $\gamma=(\gamma_{i\ell})$ of $(a_i)$ and $(b_\ell)$ sends mass to $t_j$ from points satisfying
\[
|s_i-t_j|\ge\delta/4.
\]
Since $t_j$ receives mass $b_j\ge p_0$, the coupling cost is at least
\[
\sum_i\gamma_{ij}|s_i-t_j|
\ge\frac{\delta}{4}\sum_i\gamma_{ij}
=\frac{b_j\delta}{4}
\ge\frac{p_0\delta}{4},
\]
contradicting Eq.~\eqref{eq:scalar-small-transport} after taking the infimum over $\gamma$. Thus every $I_j$ contains a fitted component.

The $m$ disjoint intervals contain all $m$ fitted labels, so each interval contains exactly one label. Relabeling the fitted components gives
\begin{equation}
s_j\in I_j,
\qquad j=1,\ldots,m,
\label{eq:fitted-atom-localization}
\end{equation}
with no fitted labels outside $\bigcup_j I_j$.

\medskip\noindent\textbf{Step 2: Spacing of the union support.}
Let
\begin{equation}
z_1<\cdots<z_N,
\qquad N\le2m,
\label{eq:union-support-nodes}
\end{equation}
be the distinct union of the numerical support points $\{s_i\}\cup\{t_j\}$. Each interval $I_j$ contains its true point $t_j$ and at most one distinct fitted point $s_j$, hence at most two distinct union nodes. Therefore any three consecutive nodes cannot all lie in the same interval. The first and third lie in two distinct intervals, whose distance is at least
\[
\delta-2(\delta/4)=\delta/2.
\]
Thus
\begin{equation}
z_{r+2}-z_r\ge\delta/2,
\qquad 1\le r\le N-2.
\label{eq:localized-triple-spacing}
\end{equation}
This is the only node-separation fact needed in the small-transport interpolation.

\medskip\noindent\textbf{Step 3: The weighted dual function.}
Define
\begin{equation}
g(t)=e^{t^2/(4\sigma_0)}\psi(t),
\qquad |t|\le M.
\label{eq:weighted-dual-function}
\end{equation}
For $s,t\in[-M,M]$, we have
\begin{equation}
\begin{aligned}
|g(s)-g(t)|
&\le e^{M^2/(4\sigma_0)}|\psi(s)-\psi(t)|+
|\psi(t)|\left|e^{s^2/(4\sigma_0)}-e^{t^2/(4\sigma_0)}\right|\\
&\le C_1(M,\sigma_0)|s-t|,
\end{aligned}
\label{eq:weighted-dual-lipschitz}
\end{equation}
where $C_1(M,\sigma_0)>0$ depends only on the support bound and variance. Thus $\operatorname{Lip}(g)\le C_1(M,\sigma_0)$. Since $g(0)=0$,
\begin{align*}
|g(s)|\leq C_1(M,\sigma_0)|s|\leq MC_1(M,\sigma_0),
\end{align*}
for $s\in[-M,M]$. It follows that $\|g\|_\infty\le MC_1(M,\sigma_0)$, and hence
\begin{equation}
\|g\|_\infty+\operatorname{Lip}(g)
\le (M+1)C_1(M,\sigma_0):= C(M,\sigma_0).
\label{eq:weighted-dual-uniform-bound}
\end{equation}

\medskip\noindent\textbf{Step 4: An interpolating RKHS witness.}
Let $P$ denote the unique polynomial of degree at most $N-1$ satisfying
\begin{equation}
P(z_r)=g(z_r),
\qquad r=1,\ldots,N.
\label{eq:dual-interpolation-values}
\end{equation}
Apply Lemma~\ref{lem:tripleDD} with $\gamma=\delta/2$ and $N_0=2m$. Its coefficient bound yields
\begin{equation}
\sum_{q=0}^{N-1}|[t^q]P(t)|
\le C(m,M,\sigma_0)\bar\delta^{-(N-2)}
\le C(m,M,\sigma_0)\bar\delta^{-(2m-2)}.
\label{eq:dual-interpolant-coefficients}
\end{equation}
The exponent $2m-2$ is uniform over all admissible pairs. Shared support locations reduce $N$ and can yield a smaller power for a particular pair.

Set
\begin{equation}
F(t)=e^{-t^2/(4\sigma_0)}P(t).
\label{eq:rkhs-interpolation-witness}
\end{equation}
Lemma~\ref{lem:rkhspoly} and Eq.~\eqref{eq:dual-interpolant-coefficients} give
\begin{equation}
\|F\|_{\mathcal H_{\sigma_0}}
\le C(m,M,\sigma_0)\bar\delta^{-(2m-2)}.
\label{eq:rkhs-witness-norm}
\end{equation}
For every atom $z$ of either $\mu$ or $\mu_*$, we have $P(z)=g(z)$ by interpolation, and therefore
\begin{equation}
F(z)=e^{-z^2/(4\sigma_0)}g(z)=\psi(z).
\label{eq:rkhs-witness-support-values}
\end{equation}
Consequently,
\begin{equation}
\int Fd\nu=\int\psi d\nu=W_1(\mu,\mu_*).
\label{eq:rkhs-witness-transport-identity}
\end{equation}
Using the RKHS duality \eqref{eq:rkhs-duality-app}, we have
\begin{equation}
W_1(\mu,\mu_*)
\le C(m,M,\sigma_0)\bar\delta^{-(2m-2)}
\|\phi_{\sigma_0}*\nu\|_2.
\label{eq:small-transport-inverse}
\end{equation}
The conversion from $\bar\delta$ to $\delta$ is made for both regimes in Step~8.

\medskip\noindent\textbf{Step 5: Smoothing at large transport cost.}
Suppose instead that
\begin{equation}
W_1(\mu,\mu_*)\ge\frac{p_0\delta}{8}.
\label{eq:scalar-large-transport}
\end{equation}
This regime permits fitted collisions, so the union support need not satisfy the three-node spacing condition. We smooth the same dual function and apply the derivative-based interpolation estimate.

Choose a nonnegative $C^\infty$-function $\zeta$ supported on $[-1,1]$ with $\int\zeta=1$, and put
\begin{equation}
\zeta_\varepsilon(t)=\varepsilon^{-1}\zeta(t/\varepsilon),
\qquad
\psi_\varepsilon=\psi*\zeta_\varepsilon,
\qquad
\varepsilon=\frac{p_0\bar\delta}{64}.
\label{eq:dual-mollification}
\end{equation}
For every $t$, we have
\begin{equation}
\begin{aligned}
|\psi_\varepsilon(t)-\psi(t)|
&=\left|\int\{\psi(t-z)-\psi(t)\}\zeta_\varepsilon(z)dz\right|\le\int|z|\zeta_\varepsilon(z)dz
\le\varepsilon.
\end{aligned}
\label{eq:dual-mollification-error}
\end{equation}
The total variation norm of $\nu$ is at most two, so
\begin{equation}
\left|\int(\psi_\varepsilon-\psi)d\nu\right|
\le2\varepsilon=\frac{p_0\bar\delta}{32}\leq \frac{1}{4}W_1(\mu,\mu_*),
\label{eq:mollified-witness-integral-error}
\end{equation}
where the final inequality follows from Eq.~\eqref{eq:scalar-large-transport}. By Eq.~\eqref{eq:wasserstein-dual-witness},
\begin{equation}
\int\psi_\varepsilon d\nu
\ge\frac{3}{4}W_1(\mu,\mu_*).
\label{eq:mollified-witness-transport-bound}
\end{equation}

\medskip\noindent\textbf{Step 6: Derivative bounds.}
The $1$-Lipschitz function $\psi$ is absolutely continuous on compact intervals and has a weak derivative with $\|\psi'\|_{L^\infty(\R)}\le1$. The convolution estimates in this step use Lebesgue norms on $\R$. For $q\ge1$,
\begin{equation}
\psi_\varepsilon^{(q)}
=\psi'*\zeta_\varepsilon^{(q-1)}.
\label{eq:mollified-dual-derivative}
\end{equation}
Thus, Young's inequality gives
\begin{equation}
\|\psi_\varepsilon^{(q)}\|_\infty
\le\|\psi'\|_\infty
\|\zeta_\varepsilon^{(q-1)}\|_1
\le C_q\varepsilon^{-(q-1)}
\le C_q\bar\delta^{-(q-1)}.
\label{eq:mollified-dual-derivative-bound}
\end{equation}
Also $\|\psi_\varepsilon\|_{L^\infty([-M,M])}\le M+\varepsilon\le M+1$. Constants in the derivative bounds may depend on $p_0$, as allowed in the lemma.

Set
\begin{equation}
g_\varepsilon(t)=e^{t^2/(4\sigma_0)}\psi_\varepsilon(t).
\label{eq:weighted-mollified-dual}
\end{equation}
For each $q\le2m-1$, Leibniz's rule gives
\begin{equation}
g_\varepsilon^{(q)}
=\sum_{r=0}^q\binom qr
\left(e^{t^2/(4\sigma_0)}\right)^{(r)}
\psi_\varepsilon^{(q-r)}.
\label{eq:weighted-mollified-leibniz}
\end{equation}
All derivatives of the Gaussian multiplier are uniformly bounded on $[-M,M]$. If $q-r=0$ use the uniform bound on $\psi_\varepsilon$; otherwise use Eq.~\eqref{eq:mollified-dual-derivative-bound}. Since $\bar\delta\le1$, every smaller inverse power is dominated by $\bar\delta^{-(q-1)}$. Thus
\begin{equation}
\|g_\varepsilon^{(q)}\|_{L^\infty([-M,M])}
\le C_q\bar\delta^{-(q-1)},
\qquad 1\le q\le2m-1.
\label{eq:weighted-mollified-derivative-bound}
\end{equation}

\medskip\noindent\textbf{Step 7: Interpolation without fitted separation.}
Let $z_1<\cdots<z_N$ again denote the distinct union of the two numerical supports; now no spacing property is assumed. Let $P_\varepsilon$ interpolate $g_\varepsilon$ at these nodes. Lemma~\ref{lem:smoothDD} and Eq.~\eqref{eq:weighted-mollified-derivative-bound} yield
\begin{equation}
\sum_q|[t^q]P_\varepsilon(t)|
\le C\bar\delta^{-(2m-2)}.
\label{eq:mollified-interpolant-coefficients}
\end{equation}
Define
\[
F_\varepsilon(t)=e^{-t^2/(4\sigma_0)}P_\varepsilon(t).
\]
Then Lemma~\ref{lem:rkhspoly} gives
\begin{equation}
\|F_\varepsilon\|_{\mathcal H_{\sigma_0}}
\le C\bar\delta^{-(2m-2)}.
\label{eq:mollified-rkhs-witness-norm}
\end{equation}
Interpolation makes $F_\varepsilon$ agree with $\psi_\varepsilon$ on both supports. Hence
\begin{equation}
\int F_\varepsilon d\nu
=\int\psi_\varepsilon d\nu
\ge\frac{3}{4}W_1(\mu,\mu_*).
\label{eq:mollified-rkhs-transport-bound}
\end{equation}
The RKHS duality therefore yields
\begin{equation}
\frac{3}{4}W_1(\mu,\mu_*)
\le C\bar\delta^{-(2m-2)}
\|\phi_{\sigma_0}*\nu\|_2.
\label{eq:large-transport-inverse}
\end{equation}
This proves the inverse inequality in the large-transport regime.

\medskip\noindent\textbf{Step 8: Removal of the spacing truncation.}
Combining Eqs.~\eqref{eq:small-transport-inverse} and \eqref{eq:large-transport-inverse},
\begin{equation}
\|\phi_{\sigma_0}*\nu\|_2
\ge c\bar\delta^{2m-2}W_1(\mu,\mu_*).
\label{eq:truncated-spacing-inverse}
\end{equation}
Because all true locations lie in $[-M,M]$,
\[
0<\delta\le2M.
\]
If $\delta\le1$ then $\bar\delta=\delta$. If $\delta>1$, then
\[
1=\bar\delta^{2m-2}
\ge\frac{\delta^{2m-2}}{\max\{1,2M\}^{2m-2}}.
\]
Thus
\begin{equation}
\bar\delta^{2m-2}
\ge C(M,m)^{-1}\delta^{2m-2}.
\label{eq:spacing-truncation-comparison}
\end{equation}
Absorb this fixed factor into $c_G$ in Eq.~\eqref{eq:truncated-spacing-inverse} to obtain Eq.~\eqref{eq:scalar-gaussian-inverse}.
\end{proof}

\begin{remark}[Origin of the exponent $2m-2$]\label{rem:scalar-exponent}
There are at most $2m$ union support nodes, so interpolation uses divided differences through order $2m-1$. Lipschitz continuity bounds the first-order difference without inverse spacing. The remaining orders contribute at most $2m-2$ inverse powers. This accounts for the exponent in Lemma~\ref{lem:scalar}.
\end{remark}

% \begin{remark}[The large-$W_1$ regime]\label{rem:largeW}
% Small transport cost and the true weight floor force one fitted label into each disjoint neighborhood of a true atom. This conclusion can fail at larger transport cost, where fitted atoms may collide. Smoothing the dual function controls the divided differences through derivatives and removes the need for fitted separation.
% \end{remark}

\subsection{Gaussian inversion with separated support clusters}\label{app:clusterRKHS}

The scalar bound can be sharpened when the union support occupies separated blocks. Divided differences within a block depend on its local spacing, whereas those spanning blocks have denominators bounded below by the macroscopic gap. This distinction yields an exponent determined by the largest block rather than the total number of components, in accordance with the localization used in finite-mixture separation bounds \citep{NguyenSeparation2026}.

We first establish the interpolation estimate for consecutive support blocks and then apply it to Gaussian mixtures. The statement includes one block, for which the gap conditions disappear and Lemma~\ref{lem:scalar} applies directly. With several blocks, the fitted and true labels must have the stated allocation to separated full-support intervals.

\subsubsection{Divided differences within and between support blocks}

\begin{lemma}[Divided differences across separated blocks]\label{lem:blockDD}
Let
\[
z_1<\cdots<z_N\in[-M,M]
\]
be partitioned into consecutive nonempty blocks $B_1,\ldots,B_q$. When $q\ge2$, assume that distinct blocks are separated by at least $D>0$. No gap is specified when $q=1$. Put $N_m=|B_m|$ and
\[
r_*:=\max_m(N_m-2)_+.
\]
Fix $0<\gamma\le1$. Suppose $g$ is specified at the nodes and satisfies one of the following two conditions.
\begin{enumerate}[label=(\alph*)]
\item Within every block, each three consecutive nodes have span at least $\gamma$, and $|g|$ and every first divided difference are bounded by $L$.
\item The node values are restrictions of a $C^{N-1}$ function on $[-M,M]$ satisfying $\|g\|_\infty\le L_0$ and, for every $m$ and $1\le r\le N_m-1$,
\[
\|g^{(r)}\|_\infty\le L_r\gamma^{-(r-1)}.
\]
\end{enumerate}
Then, for $1\le i\le j\le N$,
\[
|[z_i,\ldots,z_j]g|
\le C\gamma^{-r_*}.
\]
Consequently the Newton interpolant $P$ of $g$ on the $N$ nodes satisfies
\[
\sum_{r=0}^{N-1}|[t^r]P(t)|
\le C'\gamma^{-r_*}.
\]
Here $C,C^{\prime}>0$ depend only on $N,M$ and the displayed $L$-constants, and additionally on $D$ when $q\ge2$.
\end{lemma}

\begin{proof}
Condition (a) uses node spacing as in the small-transport argument; condition (b) uses derivative bounds as in the smoothed argument. In both cases, we first control divided differences within a block and then extend the bound across blocks.

Consider first a divided difference $[z_i,\ldots,z_j]g$ whose nodes all lie in one block $B_m$. For a zeroth-order divided difference, the value bound applies directly. For positive order, under condition (a), Lemma~\ref{lem:tripleDD} applied inside that block yields
\[
|[z_i,\ldots,z_j]g|
\le C_m\gamma^{-(j-i-1)_+}
\le C_m\gamma^{-r_*},
\]
because $j-i+1\le N_m$. Under condition (b), the generalized mean-value formula gives
\[
[z_i,\ldots,z_j]g
=\frac{g^{(j-i)}(\xi)}{(j-i)!}
\]
for some $\xi$ in the convex hull of the nodes. If $j=i$ the claim follows from the zeroth-order bound; if $j>i$, the derivative assumption gives
\[
|[z_i,\ldots,z_j]g|
\le C\gamma^{-(j-i-1)}
\le C\gamma^{-r_*}.
\]

We argue by induction on the order $r=j-i$. The claim is already established when the nodes lie in one block. Otherwise $z_i$ and $z_j$ belong to different blocks because the blocks are consecutive in the global ordering. Hence
\[
z_j-z_i\ge D.
\]
The divided-difference recursion gives
\[
[z_i,\ldots,z_j]g
=
\frac{[z_{i+1},\ldots,z_j]g-[z_i,\ldots,z_{j-1}]g}{z_j-z_i}.
\]
Both numerator divided differences have order $r-1$. By the induction hypothesis each is at most $C_{r-1}\gamma^{-r_*}$, and therefore
\[
|[z_i,\ldots,z_j]g|
\le \frac{2C_{r-1}}{D}\gamma^{-r_*}.
\]
For $q=1$, the within-block argument already covers every divided difference. For $q\ge2$, since $r\le N-1$, iterating this recursion changes only a constant depending on $N$ and $D$; it never introduces another inverse power of $\gamma$.

Write the Newton interpolant as
\[
P(t)=\sum_{r=0}^{N-1}[z_1,\ldots,z_{r+1}]g\prod_{h=1}^r(t-z_h).
\]
Every node has absolute value at most $M$, so the sum of absolute monomial coefficients of $\prod_{h=1}^r(t-z_h)$ is bounded by $(1+M)^r$. Combining this deterministic coefficient bound with the preceding divided-difference estimate and summing over $r\le N-1$ proves the final assertion.
\end{proof}

\subsubsection{Clustered inversion and stable component counts}

\begin{lemma}[Clustered scalar Gaussian stability]\label{lem:scalarMC}
Fix $n_0\ge2$, $M>0$, $\sigma_0>0$, $p_0\in(0,1/n_0]$, $A\ge1$, and a scale $0<\delta\le2M$. Let
\[
\mu=\sum_{i=1}^{n_0} a_i\delta_{s_i},
\qquad
\mu_*=\sum_{j=1}^{n_0} b_j\delta_{t_j},
\qquad a_i,b_j\ge p_0,
\]
be probability measures with all locations in $[-M,M]$. Partition the true indices into nonempty blocks $J_1,\ldots,J_q$, where $1\le q\le n_0$. Suppose that for each block $r$,
\[
\delta\le |t_j-t_\ell|\le A\delta
\quad\text{for distinct }j,\ell\in J_r.
\]
When $q\ge2$, fix $D>0$ and assume that the true block intervals satisfy
\begin{equation}
\operatorname{dist}\!\left(
\operatorname{conv}\{t_j:j\in J_r\},
\operatorname{conv}\{t_j:j\in J_{r'}\}
\right)\ge D,
\qquad r\ne r',
\label{eq:scalar-true-block-gap}
\end{equation}
and
\begin{equation}
A\delta\le D/8.
\label{eq:scalar-block-scale-condition}
\end{equation}
Assume also that the fitted indices admit disjoint (possibly empty) subsets $I_1,\ldots,I_q$ whose union is $[n_0]$. For
\[
K_r:=\operatorname{conv}\bigl(\{t_j:j\in J_r\}\cup\{s_i:i\in I_r\}\bigr),
\]
when $q\ge2$ assume $\dist(K_r,K_{r'})\ge3D/4$ for $r\ne r'$. When $q=1$, $I_1=J_1=[n_0]$; all conditions involving $D$ are omitted, and the fitted support is unrestricted within $[-M,M]$. Put
\[
r_*:=\max_{1\le r\le q}\{|I_r|+|J_r|-2\}_+.
\]
Then
\[
\|\phi_{\sigma_0}*\mu-\phi_{\sigma_0}*\mu_*\|_{L^2(dy)}
\ge c_{\rm mc}\delta^{r_*}W_1(\mu,\mu_*).
\]
Here $c_{\rm mc}>0$ depends only on $n_0,M,\sigma_0,p_0,A$ and, when $q\ge2$, $D$.
\end{lemma}

The full-support intervals are disjoint and hence form consecutive blocks after sorting the union support. For $q\ge2$, true interval gaps of at least $D$ and fitted localization within $D/8$ of the assigned true interval ensure the required full-block gap. The proof uses only that gap and the stated component counts.

\begin{proof}
If $q=1$, the actual minimum true spacing $\delta_*:=\min_{j\ne\ell}|t_j-t_\ell|$ is at least $\delta$, and $r_*=2n_0-2$. Lemma~\ref{lem:scalar} therefore gives
\[
\|\phiSig*\mu-\phiSig*\mu_*\|_2
\ge c_G\delta_*^{2n_0-2}W_1(\mu,\mu_*)
\ge c_G\delta^{r_*}W_1(\mu,\mu_*).
\]
This case requires no condition on the fitted locations beyond boundedness.

Suppose henceforth $q\ge2$. Since $A\ge1$ and $A\delta\le D/8$, the minimum spacing across all true atoms is at least $\delta$, including pairs in different blocks. We follow Lemma~\ref{lem:scalar}, replacing the global interpolation estimate by Lemma~\ref{lem:blockDD}.

Let $\nu=\mu-\mu_*$, $W=W_1(\mu,\mu_*)$, and choose a $1$-Lipschitz dual function for $W_1$ $\psi$ normalized by $\psi(0)=0$. Thus
\[
\int\psi\,d\nu=W,
\qquad
|\psi|\le M
\quad\text{on }[-M,M].
\]
Put $\bar\delta=1\wedge\delta$.

Assume
\[
W<\frac{p_0\delta}{8}.
\]
The localization argument in Lemma~\ref{lem:scalar} shows that every interval $(t_j-\delta/4,t_j+\delta/4)$ contains one fitted label. Since there are exactly $n_0$ fitted labels, there is exactly one fitted label in each such interval. A fitted label within $\delta/4$ of a true atom must be in its block, because every other full block is at distance at least $3D/4>\delta/4$. Consequently the fitted block counts equal the true block counts in this regime and each full block contains at most $2|J_m|$ distinct nodes. Inside a block, every three consecutive union nodes have span at least $\delta/2$; different blocks are separated by a fixed multiple of $D$.

Define
\[
g(t)=e^{t^2/(4\sigma_0)}\psi(t).
\]
As in equations \eqref{eq:weighted-dual-lipschitz}--\eqref{eq:weighted-dual-uniform-bound}, $g$ is uniformly bounded and Lipschitz. Lemma~\ref{lem:blockDD}(a), with $\gamma=\bar\delta/2$, gives an interpolating polynomial $P$ satisfying
\[
\sum_r|[t^r]P(t)|
\le C\bar\delta^{-r_*}.
\]
The number of distinct nodes may be smaller than the number of labels, so $r_*$ is a uniform upper bound for the exponent. Set
\[
F(t)=e^{-t^2/(4\sigma_0)}P(t).
\]
Lemma~\ref{lem:rkhspoly} yields
\[
\|F\|_{\mathcal H_{\sigma_0}}
\le C\bar\delta^{-r_*}.
\]
Since $F=\psi$ on both supports,
\[
W=\int F\,d\nu
\le C\bar\delta^{-r_*}\|\phi_{\sigma_0}*\nu\|_2.
\]

Assume now
\[
W\ge\frac{p_0\delta}{8}.
\]
Choose the same mollifier as in Lemma~\ref{lem:scalar} with bandwidth
\[
\varepsilon=\frac{p_0\bar\delta}{64},
\qquad
\psi_\varepsilon=\psi*\zeta_\varepsilon.
\]
The estimates \eqref{eq:dual-mollification-error}, \eqref{eq:mollified-witness-integral-error}, and \eqref{eq:mollified-witness-transport-bound} give
\[
\int\psi_\varepsilon\,d\nu\ge\frac{3}{4}W.
\]
Moreover, for each derivative order $r\ge1$,
\[
\|\psi_\varepsilon^{(r)}\|_\infty
\le C_r\bar\delta^{-(r-1)}.
\]
For
\[
g_\varepsilon(t)=e^{t^2/(4\sigma_0)}\psi_\varepsilon(t),
\]
Leibniz's rule gives the same derivative bounds
\[
\|g_\varepsilon^{(r)}\|_\infty
\le C_r\bar\delta^{-(r-1)},\qquad r\ge1.
\]

Also $\|g_\varepsilon\|_{L^\infty([-M,M])}\le C$, since $|\psi_\varepsilon|\le M+\varepsilon$ on this interval. Interpolate $g_\varepsilon$ at the distinct union support points. A divided difference entirely inside block $m$ has order at most $|I_m|+|J_m|-1$, so Lemma~\ref{lem:blockDD}(b), with $\gamma=\bar\delta$, bounds it by $C\bar\delta^{-r_*}$. A divided difference spanning several blocks is obtained recursively by dividing differences of lower order by a denominator at least $D/2$; hence it costs only a fixed factor and no further power of $\bar\delta$. Therefore the interpolating polynomial $P_\varepsilon$ satisfies
\[
\sum_r|[t^r]P_\varepsilon(t)|
\le C\bar\delta^{-r_*}.
\]
Set $F_\varepsilon=e^{-t^2/(4\sigma_0)}P_\varepsilon$. Then
\[
\|F_\varepsilon\|_{\mathcal H_{\sigma_0}}
\le C\bar\delta^{-r_*},
\qquad
\int F_\varepsilon d\nu
=\int\psi_\varepsilon d\nu
\ge\frac{3}{4}W.
\]
RKHS duality gives
\[
\|\phi_{\sigma_0}*\nu\|_2
\ge c\bar\delta^{r_*}W.
\]

Since $\delta\le2M$, the same calculation as equation \eqref{eq:spacing-truncation-comparison} turns $\bar\delta^{r_*}$ into $\delta^{r_*}$ at the price of a fixed factor depending only on $M$ and $n_0$. This proves the lemma.
\end{proof}

\begin{corollary}[Stable cluster counts]\label{cor:scalarMCstable}
Under the hypotheses of Lemma~\ref{lem:scalarMC}, assume $|I_m|=|J_m|=s_m$ for every $m$, and put $\smax:=\max_m s_m$. Then
\[
\|\phi_{\sigma_0}*\mu-\phi_{\sigma_0}*\mu_*\|_2
\ge c\delta^{2s_{\max}-2}W_1(\mu,\mu_*).
\]
The constant $c>0$ has the same dependencies as in Lemma~\ref{lem:scalarMC}.
\end{corollary}
This follows by substituting $r_*=2\smax-2$ into the lemma.

\section{From Wasserstein Distance to Component Matching}\label{app:denseproof}

The scalar inverse inequalities control Wasserstein distance between conditional mixing measures. Functional recovery requires an additional comparison with a labelled matching cost that records expert displacement and gate error. This appendix establishes that comparison for one true cluster and for several separated clusters. The argument complements quantitative mixture-identification bounds in Wasserstein distance \citep{Nguyen2013,HoNguyen2016EJS,HoNguyen2016AOS}.

When matched weights differ, a permutation alone does not define a transport coupling. We therefore begin with an optimal labelled coupling \citep{Villani2003}. Small transport cost forces a local matching and controls its weight discrepancies through off-diagonal mass. At larger cost, the cluster diameter bounds the displacement of any admissible pairing.

\subsection{Matching within one true cluster}

For the labelled probability measures
\begin{equation}
\mu=\sum_{i=1}^{m}a_i\delta_{s_i},
\qquad
\mu_*=\sum_{j=1}^{m}b_j\delta_{t_j}
\label{eq:matching-mixture-pair}
\end{equation}
and a scale $\DeltaSep>0$, define
\begin{equation}
\mathcal A_{\DeltaSep}(\mu,\mu_*)
=
\min_{\sigma\in\Sigma_m}
\sum_{j=1}^m
\left[
b_j|s_{\sigma(j)}-t_j|
+\DeltaSep|a_{\sigma(j)}-b_j|
\right].
\label{eq:pointwise-assignment-loss}
\end{equation}

\begin{lemma}[Matching loss controlled by $W_1$]\label{lem:assign}
Assume the hypotheses of Lemma~\ref{lem:scalar} and, for $\DeltaSep>0$ and $0<a_0\le A_0$,
\begin{equation}
\delta:=\min_{j\ne\ell}|t_j-t_\ell|\ge a_0\DeltaSep,
\qquad
\max_{j,\ell}|t_j-t_\ell|\le A_0\DeltaSep.
\label{eq:matching-cluster-geometry}
\end{equation}
Then
\begin{equation}
\mathcal A_{\DeltaSep}(\mu,\mu_*)
\le C_AW_1(\mu,\mu_*),
\label{eq:wasserstein-assignment-bound}
\end{equation}
where $C_A>0$ depends only on $m,p_0,a_0,A_0$.
\end{lemma}

\begin{proof}
Put
\[
W=W_1(\mu,\mu_*).
\]
Choose an optimal labelled coupling
\begin{equation}
\gamma=(\gamma_{ij})_{1\le i,j\le m},
\qquad
\gamma_{ij}\ge0,
\label{eq:labelled-optimal-coupling}
\end{equation}
with marginals
\begin{equation}
\sum_{j=1}^m\gamma_{ij}=a_i,
\qquad
\sum_{i=1}^m\gamma_{ij}=b_j,
\label{eq:labelled-coupling-marginals}
\end{equation}
and cost
\begin{equation}
W=\sum_{i,j}\gamma_{ij}|s_i-t_j|.
\label{eq:labelled-coupling-cost}
\end{equation}
Compactness of the feasible transport polytope ensures existence of an optimal coupling. Labels remain distinct even when their locations coincide; splitting mass among those labels preserves the transport cost.

\medskip\noindent\textbf{Step 1: Matching at small transport cost.}
Assume
\begin{equation}
W<\frac{p_0\delta}{8}.
\label{eq:matching-small-transport}
\end{equation}
By the localization argument used in the proof of Lemma~\ref{lem:scalar}, the disjoint intervals
\[
I_j=(t_j-\delta/4,t_j+\delta/4)
\]
contain one fitted label each. Relabel the fitted indices so that
\begin{equation}
|s_j-t_j|<\delta/4,
\qquad j=1,\ldots,m.
\label{eq:matching-localization}
\end{equation}
We use this pairing in the assignment loss.

\medskip\noindent\textbf{Step 2: Weight error from off-diagonal transport.}
Define the total off-diagonal coupling mass
\begin{equation}
O:=\sum_{i\ne j}\gamma_{ij}.
\label{eq:off-diagonal-mass}
\end{equation}
For $i\ne j$, the reverse triangle inequality and Eq.~\eqref{eq:matching-localization} give
\begin{equation}
|s_i-t_j|
\ge |t_i-t_j|-|s_i-t_i|
\ge\delta-\delta/4
=3\delta/4.
\label{eq:off-diagonal-distance}
\end{equation}
Therefore the off-diagonal part of the optimal transport cost satisfies
\[
W
\ge\sum_{i\ne j}\gamma_{ij}|s_i-t_j|
\ge\frac{3\delta}{4}O,
\]
and hence
\begin{equation}
O\le\frac{4W}{3\delta}.
\label{eq:off-diagonal-mass-bound}
\end{equation}
Combining with Eq.~\eqref{eq:matching-small-transport},
\begin{equation}
O<\frac{p_0}{6}.
\label{eq:off-diagonal-mass-floor}
\end{equation}

The row and column constraints in Eq.~\eqref{eq:labelled-coupling-marginals} imply the exact identity
\begin{equation}
\begin{aligned}
a_i-b_i
&=\left(\gamma_{ii}+\sum_{j\ne i}\gamma_{ij}\right)
-\left(\gamma_{ii}+\sum_{j\ne i}\gamma_{ji}\right)\\
&=\sum_{j\ne i}\gamma_{ij}-\sum_{j\ne i}\gamma_{ji}.
\end{aligned}
\label{eq:coupling-weight-balance}
\end{equation}
Taking absolute values, summing in $i$, and using the triangle inequality,
\begin{equation}
\sum_i|a_i-b_i|
\le
\sum_i\sum_{j\ne i}\gamma_{ij}
+
\sum_i\sum_{j\ne i}\gamma_{ji}
=2O.
\label{eq:weight-error-off-diagonal-bound}
\end{equation}
Thus
\begin{equation}
\sum_i|a_i-b_i|
\le\frac{8W}{3\delta}.
\label{eq:weight-error-spacing-bound}
\end{equation}
Using $\delta\ge a_0\DeltaSep$,
\begin{equation}
\DeltaSep\sum_i|a_i-b_i|
\le\frac{8}{3a_0}W.
\label{eq:scaled-weight-error-bound}
\end{equation}
This controls the mass-discrepancy term.

\medskip\noindent\textbf{Step 3: Location error from diagonal transport.}
By Eq.~\eqref{eq:labelled-coupling-marginals},
\[
\gamma_{ii}
=a_i-\sum_{j\ne i}\gamma_{ij}
\ge a_i-O
\ge p_0-O.
\]
Equation \eqref{eq:off-diagonal-mass-floor} therefore gives
\begin{equation}
\gamma_{ii}\ge\frac{5p_0}{6}.
\label{eq:diagonal-coupling-floor}
\end{equation}
The single diagonal term of the optimal cost yields
\[
W\ge\gamma_{ii}|s_i-t_i|,
\]
so
\begin{equation}
|s_i-t_i|
\le\frac{6W}{5p_0}.
\label{eq:matched-location-bound}
\end{equation}
Weighting by the true probabilities and using $\sum_i b_i=1$,
\begin{equation}
\sum_i b_i|s_i-t_i|
\le\frac{6W}{5p_0}.
\label{eq:true-weighted-location-bound}
\end{equation}
Combining Eqs.~\eqref{eq:scaled-weight-error-bound} and \eqref{eq:true-weighted-location-bound}, the cost of this specific permutation is at most
\begin{equation}
\left(\frac{6}{5p_0}+\frac{8}{3a_0}\right)W.
\label{eq:small-transport-matching-constant}
\end{equation}
Since $\mathcal A_{\DeltaSep}$ is the minimum over permutations, Eq.~\eqref{eq:wasserstein-assignment-bound} follows in the small-$W$ regime.

\medskip\noindent\textbf{Step 4: The cluster scale at large transport cost.}
Assume now
\begin{equation}
W\ge\frac{p_0\delta}{8}.
\label{eq:matching-large-transport}
\end{equation}
The lower separation bound in Eq.~\eqref{eq:matching-cluster-geometry} gives
\begin{equation}
W\ge\frac{p_0a_0\DeltaSep}{8},
\qquad\text{hence}\qquad
\DeltaSep\le\frac{8W}{p_0a_0}.
\label{eq:matching-scale-transport-bound}
\end{equation}
Together with the diameter assumption, this inequality bounds the true-support diameter by a fixed multiple of $W$.

\medskip\noindent\textbf{Step 5: The large-transport matching bound.}
For a fixed fitted label $i$, every unit of its source mass must be transported to some true atom. Therefore
\[
\sum_j\gamma_{ij}|s_i-t_j|
\ge a_i\min_j|s_i-t_j|.
\]
Since the left side is bounded by the total optimal cost $W$ and $a_i\ge p_0$,
\begin{equation}
\operatorname{dist}(s_i,\{t_1,\ldots,t_m\})
\le\frac{W}{a_i}
\le\frac{W}{p_0}.
\label{eq:fitted-to-support-bound}
\end{equation}
Choose $t_{j(i)}$ attaining this distance. For any true index $j$, the diameter condition in Eq.~\eqref{eq:matching-cluster-geometry} gives
\begin{equation}
\begin{aligned}
|s_i-t_j|
&\le|s_i-t_{j(i)}|+|t_{j(i)}-t_j|\\
&\le\frac{W}{p_0}+A_0\DeltaSep\\
&\le\left(\frac{1}{p_0}+\frac{8A_0}{p_0a_0}\right)W,
\end{aligned}
\label{eq:fitted-to-true-distance-bound}
\end{equation}
where the last step uses Eq.~\eqref{eq:matching-scale-transport-bound}.

Hence the location cost of \emph{any} permutation $\sigma$ satisfies
\begin{equation}
\sum_jb_j|s_{\sigma(j)}-t_j|
\le C W\sum_jb_j
=CW.
\label{eq:large-transport-displacement-bound}
\end{equation}
The $\ell_1$ distance between two probability vectors is at most two, so
\begin{equation}
\DeltaSep\sum_j|a_{\sigma(j)}-b_j|
\le2\DeltaSep
\le\frac{16}{p_0a_0}W.
\label{eq:large-transport-mass-bound}
\end{equation}
The displacement and mass bounds hold for every permutation in this regime. Taking the minimum and combining with the small-transport estimate proves the lemma.
\end{proof}

% \begin{remark}[True-weighted displacement in the functional loss]\label{rem:trueweights}
% The true-weighted displacement becomes $\pi_j^*(x)|h_{\sigma(j)}(x)-h_j^*(x)|$ in the functional loss. Lemma~\ref{lem:assign} controls it by the symmetric Wasserstein distance using an actual coupling. Under the same geometry, analogous comparisons hold for fitted-weighted and common-mass displacements.
% \end{remark}

\subsection{Matching within separated clusters and controlling cluster masses}

In the notation of Corollary~\ref{cor:scalarMCstable}, set $\lambda_q=0$ when $q=1$ and $\lambda_q=D$ when $q\ge2$. Let $\Sigma_{\rm block}$ be the permutations mapping each true block to its fitted block and define
\[
\begin{aligned}
\mathcal A_{\rm mc}(\mu,\mu_*)
:=\min_{\sigma\in\Sigma_{\rm block}}
\Bigg\{&\sum_j\left[b_j|s_{\sigma(j)}-t_j|+\delta|a_{\sigma(j)}-b_j|\right]\\
&+\lambda_q\sum_m\left|\sum_{j\in J_m}(a_{\sigma(j)}-b_j)\right|\Bigg\}.
\end{aligned}
\]

\begin{lemma}[Cluster-aware matching cost controlled by $W_1$]\label{lem:hierassign}
Under the hypotheses of Corollary~\ref{cor:scalarMCstable},
\[
\mathcal A_{\rm mc}(\mu,\mu_*)
\le C W_1(\mu,\mu_*).
\]
The constant $C$ depends only on the number of atoms, $M,p_0,A$ and, when $q\ge2$, the gap $D$.
\end{lemma}

For $q=1$, the aggregate term vanishes and no gap is specified.

\begin{proof}
When $q=1$, $\Sigma_{\rm block}=\Sigma_{n_0}$ and the aggregate term is zero. Lemma~\ref{lem:assign}, with its scale $\DeltaSep=\delta$, $a_0=1$ and $A_0=A$, proves the claim directly. In that lemma the actual minimum spacing is at least $\delta$, as required.

Assume henceforth $q\ge2$, so $\lambda_q=D$. Let $W=W_1(\mu,\mu_*)$ and $\gamma$ be an optimal labelled coupling.

Let $O_{\rm cross}$ be the total coupling mass transported between distinct blocks. Because distinct full blocks have distance at least $3D/4$,
\[
W\ge(3D/4)O_{\rm cross}.
\]
For each block, flow conservation gives
\[
A_m-B_m
=\text{outflow}_m-\text{inflow}_m,
\]
where $A_m$ and $B_m$ are the fitted and true block masses. Hence
\[
\sum_m|A_m-B_m|
\le2O_{\rm cross},
\]
and therefore
\[
D\sum_m|A_m-B_m|\le(8/3)W.
\]

If
\[
W<\frac{p_0\delta}{8},
\]
the localization argument of Lemma~\ref{lem:assign} yields exactly one fitted label in the $\delta/4$ neighborhood of every true atom. The resulting pairing is automatically block preserving. The remaining transport comparison in the proof of Lemma~\ref{lem:assign} then gives
\[
\sum_j\left[b_j|s_{\sigma(j)}-t_j|+\delta|a_{\sigma(j)}-b_j|\right]
\le C W.
\]
Together with the cluster-mass estimate above, this proves the claim in this regime.

Assume
\[
W\ge\frac{p_0\delta}{8}.
\]
We compare $W$ with the fixed block-gap scale. If
\[
W\ge \frac{p_0D}{16},
\]
bounded support, $\delta\le2M$, and the fixed number of blocks give a uniform upper bound for $\mathcal A_{\rm mc}$. Since $W$ is bounded below by $p_0D/16$, this implies $\mathcal A_{\rm mc}\le CW$.

It remains to consider
\[
\frac{p_0\delta}{8}\le W<\frac{p_0D}{16}.
\]
For every fitted label $i$, the row-mass argument in equation \eqref{eq:fitted-to-support-bound} gives
\[
\operatorname{dist}(s_i,\{t_j\})\le W/p_0<D/16.
\]
Every true atom in a different full block is at distance at least $3D/4$ from $s_i$. Since the nearest true atom is at distance less than $D/16$, it must belong to the same block. Hence
\[
\operatorname{dist}(s_i,\{t_j:j\in J_{m(i)}\})\le W/p_0.
\]
Within that block the true diameter is at most $A\delta$. Thus for any block-preserving permutation,
\[
|s_{\sigma(j)}-t_j|
\le W/p_0+A\delta
\le C W,
\]
where the last inequality uses $\delta\le8W/p_0$. Summing with the true weights gives a location cost at most $CW$. The component mass term satisfies
\[
\delta\sum_j|a_{\sigma(j)}-b_j|
\le2\delta
\le16W/p_0.
\]
Adding the preceding cluster-mass estimate proves the result.
\end{proof}

\section{Proofs for Results in Section~\ref{sec:dense}}\label{app:geometricproofs}
This appendix transfers the scalar estimates to the dense labelled functional targets. We first construct a set of positive covariate probability on which the true expert values satisfy the required separation bounds. Under clustered geometry, scalar transport comparisons, a finite-permutation argument, and quantitative restriction yield one global matching. For unstructured recovery, a direct scalar density-to-matching bound replaces the diameter-dependent transport comparison. Class-wide transport restriction and an exponential Remez inequality supply the expert and gate restriction needed for the global loss.

\subsection{Proof of Theorem~\ref{thm:densemcfunc}}\label{app:densemcfuncproof}
We first construct an informative covariate set on which the true cluster intervals are separated.

\begin{lemma}[Clustered good set and separated cluster intervals]\label{lem:mcgoodset}
Assume Definition~\ref{def:mcgeom} with any $q\in\{1,\ldots,k\}$ and Assumptions~\ref{ass:densesep} and \ref{ass:denseac}. There is a measurable set $E_*\subset\cX$ with $P_X(E_*)\ge\rho_*:=1/2$ on which
\[
a_*\DeltaSep\le |h_j^*(x)-h_\ell^*(x)|\le A_*\DeltaSep
\]
for every distinct within-cluster pair and, only when $q\ge2$,
\[
|h_j^*(x)-h_\ell^*(x)|\ge d_*\DeltaMacro
\]
for every inter-cluster pair, with $a_*,A_*,d_*>0$ depending only on $k,C_0,C_{\rm ac}$. When $q\ge2$, if additionally
\begin{equation}
A_*\DeltaSep<d_*\DeltaMacro,
\label{eq:good-set-scale-condition}
\end{equation}
then, for $x\in E_*$,
\begin{equation}
\operatorname{dist}\big(I_m^*(x),I_{m'}^*(x)\big)\ge d_*\DeltaMacro,
\qquad m\ne m'.
\label{eq:good-set-interval-gap}
\end{equation}
\end{lemma}

When $q=1$, the within-cluster inequalities cover every pair and give Assumption~\ref{ass:mccompat} with $\rho_{\rm cl}=1/2$, $a_{\rm cl}=a_*$, and $A_{\rm cl}=A_*$; the sole fitted block is $[k]$, with no localization condition. For $q\ge2$, Eq.~\eqref{eq:good-set-scale-condition} guarantees the true interval gap. The stronger scale condition $A_*\DeltaSep\le d_*\DeltaMacro/8$ gives both true-geometry requirements of Assumption~\ref{ass:mccompat}, with $a_{\rm cl}=a_*$, $A_{\rm cl}=A_*$, $d_{\rm cl}=d_*$ and $\rho_{\rm cl}=\rho_*$. For $q\ge2$, these conclusions do not supply the fitted membership and localization conditions.

\begin{proof}

Put $N_0:=\binom{k}{2}$ and recall $\Delta_{j\ell}:=\|h_j^*-h_\ell^*\|_{L^1(P_X)}$. Assumption~\ref{ass:denseac} gives, for $0<a\le1$,
\[
P_X\{|h_j^*-h_\ell^*|<a\Delta_{j\ell}\}
\le 2C_{\rm ac}a.
\]
Choose
\[
a_*=(8C_{\rm ac}N_0)^{-1}\wedge1.
\]
A union bound over all pairs shows that the event
\[
E_{\rm low}
=\bigcap_{j<\ell}
\{|h_j^*-h_\ell^*|\ge a_*\Delta_{j\ell}\}
\]
has probability at least $3/4$. On this event, a within-cluster pair is at least $a_*\DeltaSep$ apart, while, when $q\ge2$, an inter-cluster pair is at least $a_*\DeltaMacro$ apart.

For a within-cluster pair, Definition~\ref{def:mcgeom} gives
\[
\E|h_j^*-h_\ell^*|\le C_0\DeltaSep.
\]
Markov's inequality yields
\[
P_X\{|h_j^*-h_\ell^*|>A\DeltaSep\}
\le C_0/A.
\]
Choose $A_*=4N_0C_0$. A second union bound shows that the event on which every within-cluster pair is at most $A_*\DeltaSep$ has probability at least $3/4$. Intersecting the two events gives probability at least $1/2$. Thus one may take $\rho_*=1/2$ and $d_*=a_*$.

For $q=1$, the pairwise bounds complete the proof. For $q\ge2$, assume Eq.~\eqref{eq:good-set-scale-condition}. If two cluster intervals overlapped, an endpoint of one would belong to the other. That endpoint and an endpoint of the containing interval are true means from different clusters, with distance at most $A_*\DeltaSep<d_*\DeltaMacro$. This contradicts the inter-cluster lower bound on $E_*$. The intervals are therefore disjoint.

Once two compact intervals are disjoint, their distance is attained at endpoints, and those endpoints are true expert values from the two clusters. The pairwise inter-cluster lower bound therefore yields Eq.~\eqref{eq:good-set-interval-gap}.
\end{proof}
\begin{proof}[Proof of Theorem~\ref{thm:densemcfunc}]
Fix an admissible model pair and its partitions. Write
\[
R:=2\smax-2,\quad
N_{\mathcal C}:=\prod_m s_m!,\quad \rho_0:=\rho_{\rm cl}/N_{\mathcal C},
\quad E:=E_{\rm cl}.
\]
All restricted norms below use $P_X$ without normalization. The proof combines pointwise inversion and matching, selects one permutation on a set of sufficient probability, and then extends its component errors to the full covariate domain.

\medskip\noindent\textbf{Step 1: Pointwise inversion and matching.}
All weights in $M_G(x)$ and $M_{G_*}(x)$ are at least $\underline\pi_k=e^{-2B}/k$, and all locations lie in $[-M,M]$. Put
\[
a:=\min\{1,a_{\rm cl}\},\qquad \delta:=a\DeltaSep,\qquad A:=A_{\rm cl}/a\ge1.
\]
Then $0<\delta\le2M$. On $E$, distinct true means in the same block have distances between $\delta$ and $A\delta$. When $q\ge2$, put $D:=d_{\rm cl}\DeltaMacro$. Assumption~\ref{ass:mccompat} gives $A\delta\le D/8$, true interval gaps at least $D$, and, by Eq.~\eqref{eq:full-cluster-gap}, full interval gaps at least $3D/4$. When $q=1$, neither a gap nor a fitted-location condition is needed. The fitted and true block counts are $s_m$ in either case. Corollary~\ref{cor:scalarMCstable}, including its one-block case, therefore gives
\begin{equation}\label{eq:mc-pointwise-transport}
e(x):=\|p_G(\cdot\mid x)-p_{G_*}(\cdot\mid x)\|_{L^2(dy)}
\ge c_{\rm G}\delta^R w(x)
=c_{\rm G}a^R\DeltaSep^R w(x),\qquad x\in E,
\end{equation}
where $w(x):=W_1(M_G(x),M_{G_*}(x))$ and $c_{\rm G}>0$ is uniform in $x,G,G_*$ and $\DeltaSep$.

For $\sigma\in\Sigma_{\mathcal C}$ define
\begin{equation}\label{eq:mc-pointwise-cost}
\begin{aligned}
d_\sigma(x):={}&\sum_{j=1}^k\Bigl[
\pi_j^*(x)|h_{\sigma(j)}(x)-h_j^*(x)|
+\DeltaSep|\pi_{\sigma(j)}(x)-\pi_j^*(x)|\Bigr]\\
&+\omega_q\sum_{m=1}^q|\Pi_m(x;G)-\Pi_m^*(x)|,
\qquad d_{\min}(x):=\min_{\sigma\in\Sigma_{\mathcal C}}d_\sigma(x).
\end{aligned}
\end{equation}
Lemma~\ref{lem:hierassign} controls the same matching cost with component scale $\delta=a\DeltaSep$ and cluster scale $\lambda_q$, where $\lambda_1=0$ and $\lambda_q=D$ for $q\ge2$. Denote that cost for a fixed permutation by $\widetilde d_\sigma(x)$. Then
\[
d_\sigma(x)\le K_0\widetilde d_\sigma(x),\qquad
K_0:=\begin{cases}
\max\{1,a^{-1}\},&q=1,\\
\max\{1,a^{-1},d_{\rm cl}^{-1}\},&q\ge2.
\end{cases}
\]
Taking minima over the same finite set of permutations and using the lemma yields
\begin{equation}\label{eq:mc-transport-to-matching}
d_{\min}(x)\le C_{\rm A}w(x),\qquad x\in E,
\end{equation}
with a uniform constant $C_{\rm A}>0$. Combining the two comparisons gives
\begin{equation}\label{eq:mc-pointwise-density}
e(x)\ge c_{\rm pt}\DeltaSep^R d_{\min}(x),\qquad x\in E,
\qquad c_{\rm pt}:=c_{\rm G}a^R/C_{\rm A}>0.
\end{equation}

\medskip\noindent\textbf{Step 2: A fixed permutation and integration.}
Choose an ordering of $\Sigma_{\mathcal C}$ and let $\sigma_x$ be its first minimizer of $d_\sigma(x)$ on $E$. Every cost is measurable, so the sets
\[
A_\sigma:=\{x\in E:\sigma_x=\sigma\},\qquad \sigma\in\Sigma_{\mathcal C},
\]
form a measurable disjoint partition of $E$. There are $N_{\mathcal C}$ such sets, and $P_X(E)\ge\rho_{\rm cl}$. Thus some $\sigma^\dagger\in\Sigma_{\mathcal C}$ satisfies
\begin{equation}\label{eq:mc-fixed-permutation}
P_X(A_{\sigma^\dagger})\ge\rho_{\rm cl}/N_{\mathcal C}=\rho_0.
\end{equation}
Set $A_0:=A_{\sigma^\dagger}$ and $L:=\|d_{\sigma^\dagger}\|_{L^2(A_0,P_X)}$. On $A_0$, $d_{\min}=d_{\sigma^\dagger}$. By Tonelli's theorem and Eq.~\eqref{eq:mc-pointwise-density},
\begin{equation}\label{eq:mc-integrated}
\begin{aligned}
\|p_G-p_{G_*}\|_{L^2(P_X\otimes dy)}^2
&=\int_{\cX}e(x)^2\,dP_X(x)
\ge c_{\rm pt}^2\DeltaSep^{2R}\int_{A_0}d_{\sigma^\dagger}(x)^2\,dP_X(x),\\
\|p_G-p_{G_*}\|_{L^2(P_X\otimes dy)}
&\ge c_{\rm pt}\DeltaSep^R L.
\end{aligned}
\end{equation}
The use of the original measure $P_X$ is essential here: the restricted integral introduces no factor $P_X(A_0)^{-1}$.

\medskip\noindent\textbf{Step 3: Restriction of component and cluster errors.}
On all of $\cX$, define
\[
q_j:=\pi_j^*(h_{\sigma^\dagger(j)}-h_j^*),\qquad
r_j:=\pi_{\sigma^\dagger(j)}-\pi_j^*,\qquad
z_m:=\Pi_m(\cdot;G)-\Pi_m^*=\sum_{j\in\mathcal C_m^*}r_j.
\]
The last equality uses preservation of the blocks by $\sigma^\dagger$. Each summand in
\[
d_{\sigma^\dagger}=\sum_{j=1}^k(|q_j|+\DeltaSep|r_j|)
+\omega_q\sum_{m=1}^q|z_m|
\]
is nonnegative. Since $P_X(A_0)\ge\rho_0$, Assumption~\ref{ass:denseR} gives
\begin{equation}\label{eq:mc-expert-restriction}
L\ge\|q_j\|_{L^2(A_0,P_X)}
\ge c_{\rm res}(\rho_0)\|q_j\|_{L^2(P_X)}.
\end{equation}
For any $v\in L^2(P_X)$, Cauchy--Schwarz gives
\[
\|v\|_{L^1(A_0,P_X)}\le P_X(A_0)^{1/2}\|v\|_{L^2(A_0,P_X)}
\le\|v\|_{L^2(A_0,P_X)}.
\]
Consequently,
\begin{equation}\label{eq:mc-mass-restriction}
L\ge\DeltaSep\|r_j\|_{L^2(A_0,P_X)}
\ge\DeltaSep\|r_j\|_{L^1(A_0,P_X)}
\ge c_{\rm res}(\rho_0)\DeltaSep\|r_j\|_{L^1(P_X)}.
\end{equation}
When $q\ge2$, Assumption~\ref{ass:clusterR} similarly yields
\[
L\ge\DeltaMacro\|z_m\|_{L^2(A_0,P_X)}
\ge c_{\rm cl}(\rho_0)\DeltaMacro\|z_m\|_{L^1(P_X)}.
\]
When $q=1$, $z_1=0$ and this additional step is unnecessary. All inequalities are valid for identically zero errors.

\medskip\noindent\textbf{Step 4: The global recovery loss.}
Define
\[
\nu_q:=\begin{cases}0,&q=1,\\q,&q\ge2,\end{cases}
\qquad c_0:=\begin{cases}
c_{\rm res}(\rho_0),&q=1,\\
\min\{c_{\rm res}(\rho_0),c_{\rm cl}(\rho_0)\},&q\ge2.
\end{cases}
\]
Adding the $2k$ component inequalities and, when applicable, the $q$ cluster inequalities gives
\begin{equation}\label{eq:mc-sum-restrictions}
(2k+\nu_q)L\ge c_0\left\{
\sum_{j=1}^k\bigl[\|q_j\|_{L^2(P_X)}+\DeltaSep\|r_j\|_{L^1(P_X)}\bigr]
+\omega_q\sum_{m=1}^q\|z_m\|_{L^1(P_X)}\right\}.
\end{equation}
The expression in braces is the cost of the admissible permutation $\sigma^\dagger$ in Eq.~\eqref{eq:clustered-loss}, hence is at least $\Dmc(G,G_*)$. Substitution into Eq.~\eqref{eq:mc-integrated} proves
\[
\|p_G-p_{G_*}\|_{L^2(P_X\otimes dy)}
\ge\frac{c_{\rm pt}c_0}{2k+\nu_q}\DeltaSep^{2\smax-2}\Dmc(G,G_*).
\]
The displayed constant has the dependencies stated in the theorem. In particular, no macroscopic gap or cluster restriction constant enters when $q=1$.
\end{proof}

\subsection{Proof of Corollary~\ref{thm:densemain}}\label{appendix:proof_thm:densemain}
\begin{proof}[Proof of Corollary~\ref{thm:densemain}]
For $q=1$, the true and fitted partitions both consist of $[k]$. Lemma~\ref{lem:mcgoodset} supplies $E$ with $P_X(E)\ge1/2$ and constants $a_0,A_0$ depending only on $k,C_0,C_{\rm ac}$ such that
\[
a_0\DeltaSep\le|h_j^*(x)-h_\ell^*(x)|\le A_0\DeltaSep,
\qquad x\in E,\quad j\ne\ell.
\]
Thus Assumption~\ref{ass:mccompat} holds with $\rho_{\rm cl}=1/2$, $a_{\rm cl}=a_0$ and $A_{\rm cl}=A_0$ for every fitted model in the common class. Here $N_{\mathcal C}=k!$, $\smax=k$, $\omega_1=0$ and $\Dmc=\Dfunc$. Applying Theorem~\ref{thm:densemcfunc} at level $1/(2k!)$ gives the claim and its stated constant dependence.
\end{proof}

% \begin{remark}[Roles of the clustered recovery assumptions]\label{rem:dense-assumption-audit}
% Compact logits and bounded experts provide positive scalar weights and bounded supports. The pointwise geometry gives the Gaussian inverse estimate and the matching comparison. Fixed fitted membership makes the set of admissible permutations independent of $x$ on the good set; component restriction and, for $q\ge2$, cluster-mass restriction then recover the global loss. Functional clustering and anti-concentration provide the good-set constructions above, but are not used separately once Assumption~\ref{ass:mccompat} is imposed. Real analyticity is likewise not separately used after the restriction inequalities are assumed.
% \end{remark}

% \begin{remark}[Dense recovery and routing regions]\label{rem:dense-no-cells}
% The dense argument works on a set of fixed positive probability under $P_X$, without conditioning on a routing region. The Top-$K$ result in Section~\ref{sec:topk} instead applies the scalar estimates under conditional laws on joint routing regions. Its separation and restriction conditions must therefore be checked under those laws; small absolute region probabilities are handled by the rare-region remainder.
% \end{remark}

\subsection{Proof of Theorem~\ref{thm:denseU}}\label{app:denseUproof}
We first strengthen the scalar inverse estimate to control a labelled matching directly. We then verify the two restriction inequalities needed to extend one pointwise permutation to the whole covariate domain.

\begin{lemma}[Scalar labelled recovery without a diameter condition]\label{lem:scalar-labelled}
Under the hypotheses of Lemma~\ref{lem:scalar}, let $\DeltaSep>0$ and $a>0$ satisfy $\delta\ge a\DeltaSep$, and define
\[
A_{\DeltaSep}(\mu,\mu_*):=
\min_{\sigma\in\Sigma_m}\sum_{j=1}^m
\left[b_j|s_{\sigma(j)}-t_j|+\DeltaSep|a_{\sigma(j)}-b_j|\right].
\]
Then
\begin{equation}\label{eq:scalar-labelled-inverse}
\|\phiSig*(\mu-\mu_*)\|_{L^2(dy)}
\ge c_{\rm lab}\delta^{2m-2}A_{\DeltaSep}(\mu,\mu_*),
\end{equation}
where $c_{\rm lab}>0$ depends only on $m,M,\sigma_0,p_0,a$. No upper bound on the support diameter proportional to $\DeltaSep$ is required.
\end{lemma}

\begin{proof}
Write $e=\|\phiSig*(\mu-\mu_*)\|_2$, $W=W_1(\mu,\mu_*)$, and $I_j=(t_j-\delta/4,t_j+\delta/4)$. We distinguish whether every true atom has a nearby fitted label.

\medskip\noindent\textbf{Case 1: Every $I_j$ contains a fitted label.}
The intervals are disjoint and there are exactly $m$ fitted labels, so each interval contains exactly one label and no labels lie outside their union. Relabel so that $s_j\in I_j$. For $i\ne j$,
\[
|s_i-t_j|\ge |t_i-t_j|-|s_i-t_i|>3\delta/4,
\qquad |s_j-t_j|<\delta/4.
\]
Thus $s_j$ is the nearest fitted point to $t_j$. Every coupling must transport mass $b_j$ to $t_j$, giving
\[
W\ge\sum_jb_j|s_j-t_j|.
\]
For an optimal labelled coupling $\gamma$, let $O=\sum_{i\ne j}\gamma_{ij}$. The off-diagonal distance bound gives $O\le4W/(3\delta)$. Its marginals also give
\[
\sum_j|a_j-b_j|\le2O\le\frac{8W}{3\delta}.
\]
Consequently $A_{\DeltaSep}\le(1+8/(3a))W$. Applying Lemma~\ref{lem:scalar} proves Eq.~\eqref{eq:scalar-labelled-inverse} in this case, with no smallness requirement on $W$.

\medskip\noindent\textbf{Case 2: Some $I_j$ contains no fitted label.}
Fix such a $j$ and set
\[
P_j(z):=\prod_{\ell\ne j}(z-t_\ell)\prod_{i=1}^m(z-s_i),
\qquad F_j(z):=e^{-z^2/(4\sigma_0)}P_j(z).
\]
The degree is $2m-1$, and the sum of absolute polynomial coefficients is at most $(1+M)^{2m-1}$. Lemma~\ref{lem:rkhspoly} therefore bounds $\|F_j\|_{\mathcal H_{\sigma_0}}$ by a constant depending only on $m,M,\sigma_0$. All fitted locations and all true locations except $t_j$ are zeros, including any repeated fitted locations. The kernel-mean identity and Cauchy--Schwarz imply
\[
e\|F_j\|_{\mathcal H_{\sigma_0}}
\ge\left|\int F_j\,d(\mu-\mu_*)\right|
=b_je^{-t_j^2/(4\sigma_0)}|P_j(t_j)|.
\]
Let $D$ be the diameter of the combined supports. Every factor in $P_j(t_j)$ has magnitude at least $\delta/4$. At least one has magnitude at least $D/2$: one endpoint of the combined support is at distance at least $D/2$ from $t_j$, and this endpoint occurs among the roots. Hence
\[
|P_j(t_j)|\ge\frac D2\left(\frac\delta4\right)^{2m-2},
\qquad e\ge cD\delta^{2m-2}.
\]
For any permutation, the true-weighted displacement is at most $D$ and the weight-vector discrepancy is at most two. Since $\DeltaSep\le\delta/a\le D/a$,
\[
A_{\DeltaSep}\le D+2\DeltaSep\le(1+2/a)D.
\]
This proves Eq.~\eqref{eq:scalar-labelled-inverse}. To make the separation independence explicit, put $r=2m-2$ and define
\[
R_m:=\kappa_{\sigma_0}^{-1/2}(1+M)^{2m-1}
\max_{0\le q\le2m-1}\sqrt{(2\sigma_0)^q q!}.
\]
The coefficient bound and Eq.~\eqref{eq:gaussian-polynomial-rkhs-norm} give $\|F_j\|_{\mathcal H_{\sigma_0}}\le R_m$. If $c_G=c_G(m,M,\sigma_0,p_0)>0$ is the constant in Lemma~\ref{lem:scalar}, the two cases show that a valid choice is
\begin{equation}\label{eq:scalar-labelled-constant}
c_{\rm lab}:=\min\left\{
\frac{c_G}{1+8/(3a)},\;
\frac{p_0e^{-M^2/(4\sigma_0)}}{2\cdot4^{r}R_m(1+2/a)}
\right\}>0.
\end{equation}
Neither $R_m$ nor either entry of this minimum depends on $\delta$, $\DeltaSep$, or the combined support diameter $D$. In the second case, the factor $D$ in the lower bound for the test-function integral cancels the bound $A_{\DeltaSep}\le(1+2/a)D$; no factor $D/\delta$ is absorbed into the constant.
\end{proof}

\begin{lemma}[Expert restriction inherited from class-wide transport restriction]\label{lem:OT-expert-restriction}
If Assumption~\ref{ass:OTR} holds on the full product class, then Eq.~\eqref{eq:expert-restriction} holds at every $\rho\in(0,1]$ with constant $c_{\rm OT}(\rho)$.
\end{lemma}
\begin{proof}
Fix $j$ and $\eta\in\Omega_\eta$. Form a tuple $H$ retaining all true gates and all true experts except replacing expert $j$ by $h(\cdot,\eta)$. The product structure makes $H$ admissible. At each $x$,
\[
M_H(x)-M_{G_*}(x)
=\pi_j^*(x)\{\delta_{h(x,\eta)}-\delta_{h_j^*(x)}\}.
\]
Coupling all unchanged mass identically gives an upper bound $\pi_j^*|h(\cdot,\eta)-h_j^*|$ for $w_H$. The difference of the means, tested against the $1$-Lipschitz function $z\mapsto z$ or its negative, gives the same lower bound. Thus
\[
w_H(x)=\pi_j^*(x)|h(x,\eta)-h_j^*(x)|.
\]
Apply Assumption~\ref{ass:OTR} to this tuple. This proves the assertion, including zero expert differences. For a restricted fitted class, the argument requires that these replacement tuples be admissible, or that transport restriction hold on an ambient product class containing them.
\end{proof}

\begin{lemma}[Restriction for bounded affine softmax gates]\label{lem:affine-gate-restriction}
Let $C\subseteq\cX$ be a convex body of positive volume and $P_C$ normalized Lebesgue measure on $C$. Fix two label sets $S,T$ of size $K\le k$, and use the corresponding normalized affine softmax weights on $C$, with all logits bounded in absolute value by $B$. For every $\rho\in(0,1]$, $P_C(A)\ge\rho$, $i\in T$, and $j\in S$,
\[
\|\pi_{i,G}^{(K)}-\pi_{j,G_*}^{(K)}\|_{L^1(A,P_C)}
\ge c_\pi(K,B,d_x,\rho)
\|\pi_{i,G}^{(K)}-\pi_{j,G_*}^{(K)}\|_{L^1(P_C)},
\]
where $c_\pi>0$ is independent of the volume and shape of $C$, of expert separation, and of collisions of logit slopes. The same conclusion holds for differences of dense aggregate gate probabilities of fixed label blocks, with a constant depending on $k,B,d_x,\rho$.
\end{lemma}

\begin{proof}
The numerator of a component difference after cross multiplication is
\[
F_{ij}=\sum_{b\in S}e^{\ell_i+\ell_b^*}
-\sum_{a\in T}e^{\ell_j^*+\ell_a},
\qquad
\pi_{i,G}^{(K)}-\pi_{j,G_*}^{(K)}=\frac{F_{ij}}{Z_TZ_S^*}.
\]
It has at most $N=2K$ exponential terms. Absorb intercepts into their coefficients. Each remaining linear exponent $f$ satisfies $\sup_{x,y\in C}f(x-y)\le4B$. The multivariate Tur\'an--Nazarov inequality \citep[Theorem~3.2]{BrudnyiYomdin2016} therefore gives, for every measurable $E\subseteq C$ of positive relative measure,
\begin{equation}\label{eq:multivariate-exp-remez}
\|F\|_{L^\infty(C)}
\le e^{4B}\left(\frac{c d_x}{P_C(E)}\right)^{N-1}
\|F\|_{L^\infty(E)}.
\end{equation}
Here $c\ge1$ is numerical, and subset suprema are essential suprema. One obtains the measurable-set version from the Borel version by discarding null sets. Repeated exponents may be combined and only reduce the number of terms, so no separation of exponents is needed.

Put $u=\|F\|_{L^1(A,P_C)}$. If $u>0$, the set
\[
E=A\cap\{|F|\le2u/\rho\}
\]
has $P_C(E)\ge\rho/2$. Applying Eq.~\eqref{eq:multivariate-exp-remez} and bounding the full $L^1$ norm by the supremum gives
\[
\|F\|_{L^1(A,P_C)}
\ge c_F(N,B,d_x,\rho)\|F\|_{L^1(P_C)},
\qquad
c_F:=\frac\rho2e^{-4B}\left(\frac\rho{2cd_x}\right)^{N-1}>0.
\]
If $u=0$, the same Remez inequality applied to a full-measure subset of $A$ gives $F=0$ on $C$, so the conclusion still holds. Since
\[
K^2e^{-2B}\le Z_TZ_S^*\le K^2e^{2B},
\]
the desired constant is $c_\pi=e^{-4B}c_F(2K,B,d_x,\rho)$.

For a dense aggregate error, its numerator is
\[
\left(\sum_{i\in\widehat{\mathcal C}_m}e^{\ell_i}\right)Z_*
-\left(\sum_{j\in\mathcal C_m^*}e^{\ell_j^*}\right)Z.
\]
There are at most $2k^2$ terms, each with the same exponent-variation bound. Apply the preceding argument with $N=2k^2$ and denominator $ZZ_*$ to obtain the aggregate conclusion. None of these steps concerns the expert family.
\end{proof}

\begin{proof}[Proof of Theorem~\ref{thm:denseU}]
Fix $(G,G_*)$ in the full product class, and write
\[
e_G(x):=\|p_G(\cdot\mid x)-p_{G_*}(\cdot\mid x)\|_{L^2(dy)},
\qquad p_0=\underline\pi_k.
\]
All expert values, weights, and pointwise matching costs below are measurable; $e_G$ is measurable by integration. Bounded experts and weights ensure all functional norms are finite.

\medskip\noindent\textbf{Step 1: A separated good set.}
Set $N_k=\binom{k}{2}$ and $a_U=\min\{1,(4N_kC_{\rm ac})^{-1}\}$. Define
\begin{equation}\label{eq:unstructured-good-set}
E_U:=\bigcap_{j<\ell}\{x:|h_j^*(x)-h_\ell^*(x)|\ge a_U\Delta_{j\ell}\}.
\end{equation}
Anti-concentration gives $P_X\{|h_j^*-h_\ell^*|<a_U\Delta_{j\ell}\}\le2C_{\rm ac}a_U$. A union bound yields
\begin{equation}\label{eq:unstructured-good-mass}
P_X(E_U)\ge1-2N_kC_{\rm ac}a_U\ge\frac12.
\end{equation}
Thus, on $E_U$,
\begin{equation}\label{eq:unstructured-pointwise-spacing}
\delta_x:=\min_{j\ne\ell}|h_j^*(x)-h_\ell^*(x)|\ge a_U\DeltaSep.
\end{equation}

\medskip\noindent\textbf{Step 2: Direct pointwise labelled inversion.}
For $\sigma\in\Sigma_k$, put
\[
d_\sigma(x):=\sum_j\left[
\pi_j^*(x)|h_{\sigma(j)}(x)-h_j^*(x)|
+\DeltaSep|\pi_{\sigma(j)}(x)-\pi_j^*(x)|\right],
\qquad d_{\min}(x):=\min_\sigma d_\sigma(x).
\]
At each $x\in E_U$, the conditional mixing measures have $k$ labelled terms, weights at least $p_0$, and means in $[-M,M]$. Apply Lemma~\ref{lem:scalar-labelled} with $m=k$, $a=a_U$, and spacing $\delta_x$ to obtain
\begin{equation}\label{eq:unstructured-pointwise-bound}
e_G(x)\ge c_{\rm lab}\delta_x^{2k-2}d_{\min}(x)
\ge c_{\rm lab}a_U^{2k-2}\DeltaSep^{2k-2}d_{\min}(x).
\end{equation}
The scalar constant is uniform in $x,G,G_*,\DeltaSep$ under the stated model bounds.

\medskip\noindent\textbf{Step 3: One permutation on a large subset.}
Order the $k!$ permutations and assign each $x\in E_U$ to its first minimizing permutation. These measurable occurrence sets partition $E_U$. Hence some fixed $\sigma^\dagger$ minimizes on a measurable $A\subseteq E_U$ with
\[
P_X(A)\ge\rho_0:=\frac1{2k!}.
\]
Squaring and integrating Eq.~\eqref{eq:unstructured-pointwise-bound} over $A$ gives
\begin{equation}\label{eq:unstructured-restricted-bound}
\|p_G-p_{G_*}\|_{L^2(P_X\otimes dy)}
\ge c_{\rm lab}a_U^{2k-2}\DeltaSep^{2k-2}\Lambda,
\qquad \Lambda:=\|d_{\sigma^\dagger}\|_{L^2(A,P_X)}.
\end{equation}
The restricted norm uses the original probability measure, without renormalizing on $A$.

\medskip\noindent\textbf{Step 4: Restriction and the global loss.}
For each $j$, write $q_j=\pi_j^*(h_{\sigma^\dagger(j)}-h_j^*)$ and $r_j=\pi_{\sigma^\dagger(j)}-\pi_j^*$. Lemma~\ref{lem:OT-expert-restriction} and nonnegativity of the summands in $d_{\sigma^\dagger}$ imply
\[
\Lambda\ge\|q_j\|_{L^2(A,P_X)}
\ge c_{\rm OT}(\rho_0)\|q_j\|_{L^2(P_X)}.
\]
Lemma~\ref{lem:affine-gate-restriction}, applied on the cube with $K=k$, and Cauchy--Schwarz give
\[
\Lambda\ge\DeltaSep\|r_j\|_{L^2(A,P_X)}
\ge\DeltaSep\|r_j\|_{L^1(A,P_X)}
\ge c_\pi(k,B,d_x,\rho_0)\DeltaSep\|r_j\|_{L^1(P_X)}.
\]
Set $c_*:=\min\{c_{\rm OT}(\rho_0),c_\pi(k,B,d_x,\rho_0)\}$. Summing these $2k$ inequalities yields
\[
2k\Lambda\ge c_*\sum_j
\left[\|q_j\|_{L^2(P_X)}+\DeltaSep\|r_j\|_{L^1(P_X)}\right]
\ge c_*\Dfunc(G,G_*).
\]
Substitution into Eq.~\eqref{eq:unstructured-restricted-bound} proves the theorem with
\[
C_{\rm U,func}:=\frac{c_{\rm lab}a_U^{2k-2}c_*}{2k}>0.
\]
For a direct audit of the dependencies, take $c_{\rm lab}$ from Eq.~\eqref{eq:scalar-labelled-constant} with $m=k$, $p_0=e^{-2B}/k$, and $a=a_U$. The gate proof supplies the explicit choice
\[
c_\pi(k,B,d_x,\rho_0)
=\frac{\rho_0}{2}e^{-8B}
\left(\frac{\rho_0}{2cd_x}\right)^{2k-1},
\]
where $c\ge1$ is the numerical Remez constant. Both $a_U$ and $\rho_0=1/(2k!)$ are fixed independently of $\DeltaSep$. Thus no additional power of separation is hidden in the scalar bound, the gate restriction, or the permutation selection. Along a sequence of truths, a common positive $C_{\rm U,func}$ follows when the displayed model bounds are uniform and $\inf c_{\rm OT}(\rho_0)>0$. Mere positivity of a different restriction constant for each truth would not imply this uniform conclusion; it is the class-uniform requirement in Assumption~\ref{ass:OTR} that supplies it. If expert restriction is assumed directly at $\rho_0$, replace $c_{\rm OT}(\rho_0)$ by that constant in Step~4, with the same uniformity requirement; the remainder of the proof is unchanged.
\end{proof}

\section{Proofs for Results in Section~\ref{sec:topk}}\label{app:topproof}

On each joint routing region, the true and fitted active sets are fixed and contain $K$ labels. The scalar inversion and matching estimates therefore apply to these active components under the conditional covariate law. This appendix proves Theorem~\ref{thm:rare} by aggregating regional inverse bounds and controlling omitted regions by their total probability. It then proves Corollary~\ref{cor:top1} and gives a sufficient condition for regional separation.

\subsection{Proof of Theorem~\ref{thm:rare}}\label{app:rareproof}
We first record the routing partition identities and prove the inverse inequality on a retained region. We then aggregate these bounds and control the omitted regions.

Under the fixed priority rule, specifying a Top-$K$ set amounts to finitely many comparisons of routing scores, with a strict or non-strict inequality according to the priority of the two labels. Each comparison defines an open or closed halfspace. Consequently the sets $\mathcal R_{S,T}$ in Eq.~\eqref{eq:topk-regions} are Borel measurable, pairwise disjoint, and cover $\cX$. There are at most $\binom{k}{K}^2$ such sets. The sparse weights and densities are measurable because they are defined by measurable selection indicators and positive softmax denominators.

For $C\in\mathcal R_+$ and every nonnegative measurable $g$,
\begin{equation}\label{eq:sparse-conditioning-identity}
\int_C g(x)\,dP_X(x)=w_C\int_C g(x)\,dP_C(x).
\end{equation}
Thus, writing
\[
e_K(x):=\|p_G^{(K)}(\cdot\mid x)-p_{G_*}^{(K)}(\cdot\mid x)\|_{L^2(dy)},
\]
Tonelli's theorem and the finite partition give
\begin{equation}\label{eq:sparse-density-decomposition}
\|p_G^{(K)}-p_{G_*}^{(K)}\|_{L^2(P_X\otimes dy)}^2
=\sum_{C\in\mathcal R_+}w_C\|e_K\|_{L^2(P_C)}^2.
\end{equation}
Every conditional mixture has $L^2(dy)$ norm at most $\|\phiSig\|_{L^2(dy)}$, so the displayed norms are finite. Regions of zero probability contribute nothing. Within a positive-probability region, restriction to $A$ means integration against $P_C$ over $A$, without further normalization by $P_C(A)$.

\label{app:topmainproof}
The following lemma gives the conditional inverse bound used in the aggregation step.
\begin{lemma}[Uniform inverse inequality on a retained routing region]\label{lem:heavycell-inverse}
Under the hypotheses of Theorem~\ref{thm:rare}, every $C\in\mathcal R_\tau$ satisfies
\begin{equation}\label{eq:sparse-cell-inverse}
\|p_G^{(K)}-p_{G_*}^{(K)}\|_{L^2(P_C\otimes dy)}
\ge C_{K,\tau}\DeltaSep^{2K-2}D_C,
\end{equation}
with the common constant in Eq.~\eqref{eq:tauconstant-expanded}.
\end{lemma}

\begin{proof}
Fix a retained region $C=\mathcal R_{S,T}$. The proof follows the dense one-cluster argument under $P_C$: pointwise inversion controls a matching cost on $E_C$, one bijection is selected on a sufficiently large subset, and conditional restriction extends its errors to all of $C$. All constants are uniform over the retained regions.

\medskip\noindent\textbf{Step 1: Pointwise inversion for the matching cost.}
Assumption~\ref{ass:topR1} provides $E_C\subseteq C$ with $P_C(E_C)\ge\rho_\tau$. Remove the null set on which the simultaneous pairwise bounds fail, retaining the same notation. For $x\in E_C$, define
\[
\delta_C(x):=\min_{\substack{j,\ell\in S\\j\ne\ell}}|h_j^*(x)-h_\ell^*(x)|,
\qquad
w_K(x):=W_1(M_G^{(K)}(x),M_{G_*}^{(K)}(x)).
\]
Then $\delta_C(x)\ge a_\tau\DeltaSep>0$. The conditional mixing measures have exactly $K$ labelled terms, with all active weights at least $p_0=\underline\pi_K$ by Eq.~\eqref{eq:topk-weight-floor}. Their locations are in $[-M,M]$, and the true locations are distinct on $E_C$. Fitted locations may coincide. Hence Lemma~\ref{lem:scalar}, applied with $m=K$ and the actual true spacing $\delta_C(x)$, gives
\begin{equation}\label{eq:sparse-scalar-inverse}
e_K(x)\ge c_{\rm sc}\delta_C(x)^{2K-2}w_K(x)
\ge c_{\rm sc}a_\tau^{2K-2}\DeltaSep^{2K-2}w_K(x).
\end{equation}
For $\sigma\in\operatorname{Bij}(S,T)$, define
\begin{equation}\label{eq:sparse-pointwise-cost}
\begin{aligned}
d_\sigma(x):=\sum_{j\in S}\Bigl[
&\pi_{j,G_*}^{(K)}(x)|h_{\sigma(j)}(x)-h_j^*(x)|\\
&+\DeltaSep|\pi_{\sigma(j),G}^{(K)}(x)-\pi_{j,G_*}^{(K)}(x)|\Bigr],
\qquad d_{\min}(x):=\min_\sigma d_\sigma(x).
\end{aligned}
\end{equation}
The upper bound in Eq.~\eqref{eq:topk-cell-geometry} gives a true active-support diameter at most $A_\tau\DeltaSep$. Together with the lower spacing bound, this verifies all the geometric hypotheses of Lemma~\ref{lem:assign}, after enumerating $S$ and $T$ by $[K]$. Therefore
\[
d_{\min}(x)\le C_A(a_\tau,A_\tau)w_K(x),\qquad x\in E_C.
\]
Combining this inequality with Eq.~\eqref{eq:sparse-scalar-inverse} yields
\begin{equation}\label{eq:sparse-pointwise-inverse}
e_K(x)\ge c_{\rm pt}\DeltaSep^{2K-2}d_{\min}(x),
\qquad c_{\rm pt}:=\frac{c_{\rm sc}a_\tau^{2K-2}}{C_A(a_\tau,A_\tau)}>0.
\end{equation}
The factor $\DeltaSep^{2K-2}$ comes from scalar inversion. The upper diameter bound is used separately to compare transport with the labelled matching cost.

\medskip\noindent\textbf{Step 2: A fixed bijection and conditional integration.}
Order the $K!$ bijections from $S$ to $T$ as $\sigma^{(1)},\ldots,\sigma^{(K!)}$. Each $d_\sigma$ is measurable. The sets
\[
B_r:=E_C\cap\{d_{\sigma^{(r)}}=d_{\min}\}
\cap\bigcap_{s<r}\{d_{\sigma^{(s)}}>d_{\min}\},
\qquad 1\le r\le K!,
\]
are measurable and form a disjoint partition of $E_C$. Thus some $r_C$ satisfies
\[
P_C(B_{r_C})\ge\frac{P_C(E_C)}{K!}\ge\frac{\rho_\tau}{K!}.
\]
Put $B_C:=B_{r_C}$ and $\sigma_C:=\sigma^{(r_C)}$. On $B_C$, $d_{\min}=d_{\sigma_C}$. Squaring Eq.~\eqref{eq:sparse-pointwise-inverse}, integrating under $P_C$, and discarding the nonnegative contribution from $C\setminus B_C$ gives
\[
\|e_K\|_{L^2(P_C)}^2
\ge c_{\rm pt}^2\DeltaSep^{4K-4}
\int_{B_C}d_{\sigma_C}(x)^2\,dP_C(x).
\]
Taking square roots, and defining $\Lambda_C:=\|d_{\sigma_C}\|_{L^2(B_C,P_C)}$, we obtain
\begin{equation}\label{eq:sparse-integrated-cell}
\|e_K\|_{L^2(P_C)}\ge c_{\rm pt}\DeltaSep^{2K-2}\Lambda_C.
\end{equation}
This step takes place entirely under $P_C$. The absolute region probability $w_C$ enters only when the regional bounds are aggregated.

\medskip\noindent\textbf{Step 3: Conditional restriction of expert and gate errors.}
For $j\in S$, define on $C$
\[
q_j:=\pi_{j,G_*}^{(K)}(h_{\sigma_C(j)}-h_j^*),
\qquad r_j:=\pi_{\sigma_C(j),G}^{(K)}-\pi_{j,G_*}^{(K)}.
\]
Then $d_{\sigma_C}=\sum_{j\in S}(|q_j|+\DeltaSep|r_j|)$. The summands are nonnegative, and $P_C(B_C)\ge\rho_\tau/K!$, so Assumption~\ref{ass:topR2} gives
\begin{equation}\label{eq:sparse-component-restriction}
\Lambda_C\ge\|q_j\|_{L^2(B_C,P_C)}
\ge c_{h,\tau}\|q_j\|_{L^2(P_C)}.
\end{equation}
For the gate errors, Cauchy--Schwarz and $P_C(B_C)\le1$ imply
\[
\|r_j\|_{L^1(B_C,P_C)}
\le P_C(B_C)^{1/2}\|r_j\|_{L^2(B_C,P_C)}
\le\|r_j\|_{L^2(B_C,P_C)}.
\]
Consequently,
\begin{equation}\label{eq:sparse-mass-restriction}
\begin{aligned}
\Lambda_C&\ge\DeltaSep\|r_j\|_{L^2(B_C,P_C)}
\ge\DeltaSep\|r_j\|_{L^1(B_C,P_C)}\\
&\ge c_{\pi,\tau}\DeltaSep\|r_j\|_{L^1(P_C)}.
\end{aligned}
\end{equation}
These inequalities remain valid when an error function vanishes identically.

\medskip\noindent\textbf{Step 4: The regionwise recovery loss.}
Let $c_{\rm res,\tau}:=\min\{c_{h,\tau},c_{\pi,\tau}\}$. Summing the $2K$ inequalities in Eqs.~\eqref{eq:sparse-component-restriction} and \eqref{eq:sparse-mass-restriction} gives
\begin{equation}\label{eq:sparse-sum-restrictions}
\begin{aligned}
2K\Lambda_C
&\ge c_{\rm res,\tau}\sum_{j\in S}
\left[\|q_j\|_{L^2(P_C)}+\DeltaSep\|r_j\|_{L^1(P_C)}\right]\\
&=c_{\rm res,\tau}L_C(\sigma_C)
\ge c_{\rm res,\tau}D_C.
\end{aligned}
\end{equation}
Inserting this bound into Eq.~\eqref{eq:sparse-integrated-cell} proves Eq.~\eqref{eq:sparse-cell-inverse} with
\[
C_{K,\tau}
=\frac{c_{\rm sc}a_\tau^{2K-2}c_{\rm res,\tau}}{2K\,C_A(a_\tau,A_\tau)}.
\]
The constants depend on the region only through the uniform parameters in the two conditional assumptions. This proves the claimed uniformity over $\mathcal R_\tau$.
\end{proof}

\begin{proof}[Proof of Theorem~\ref{thm:rare}]
Use the same $C_{K,\tau}$ as in Lemma~\ref{lem:heavycell-inverse}.

\medskip\noindent\textbf{Step 1: Aggregation over retained regions.}
The exact decomposition \eqref{eq:sparse-density-decomposition}, followed by the conditional inverse inequality, gives
\begin{equation}\label{eq:sparse-retained-aggregation}
\begin{aligned}
\|p_G^{(K)}-p_{G_*}^{(K)}\|_{L^2(P_X\otimes dy)}^2
&\ge\sum_{C\in\mathcal R_\tau}w_C\|e_K\|_{L^2(P_C)}^2\\
&\ge C_{K,\tau}^2\DeltaSep^{4K-4}
\sum_{C\in\mathcal R_\tau}w_CD_C^2\\
&=C_{K,\tau}^2\DeltaSep^{4K-4}D_{K,\tau}(G,G_*)^2.
\end{aligned}
\end{equation}
Taking square roots proves Eq.~\eqref{eq:topk-inverse-retained}. If $\mathcal R_\tau$ is empty, both sums and the retained loss are zero, and this conclusion remains valid.

\medskip\noindent\textbf{Step 2: A uniform bound on omitted regions.}
For every fitted and true label, $|h_i-h_j^*|\le2M$. Also $0<\DeltaSep\le2M$ by Eq.~\eqref{eq:functional-separation} and boundedness. Fix any $C=\mathcal R_{S,T}\in\mathcal R_+$ and any bijection $\sigma:S\to T$. Since each active weight belongs to $[0,1]$ and $P_C$ is a probability measure,
\[
\begin{aligned}
\sum_{j\in S}\|\pi_{j,G_*}^{(K)}(h_{\sigma(j)}-h_j^*)\|_{L^2(P_C)}
&\le 2M\sum_{j\in S}\|\pi_{j,G_*}^{(K)}\|_{L^2(P_C)}
\le2KM,\\
\DeltaSep\sum_{j\in S}\|\pi_{\sigma(j),G}^{(K)}-\pi_{j,G_*}^{(K)}\|_{L^1(P_C)}
&\le K\DeltaSep\le2KM.
\end{aligned}
\]
Hence
\begin{equation}\label{eq:sparse-uniform-cell-loss}
D_C\le L_C(\sigma)\le4KM.
\end{equation}
The bound holds on every positive-probability region, including those on which the inverse assumptions are not imposed.

\medskip\noindent\textbf{Step 3: The rare-region contribution.}
By the definitions of the full and retained losses,
\begin{equation}\label{eq:sparse-rare-loss-decomposition}
\begin{aligned}
\Dtop(G,G_*)^2
&=D_{K,\tau}(G,G_*)^2
  +\sum_{C\in\mathcal R_+\setminus\mathcal R_\tau}w_CD_C^2\\
&\le D_{K,\tau}(G,G_*)^2+16K^2M^2w_{\rm rare}(\tau).
\end{aligned}
\end{equation}
Taking square roots and using $\sqrt{u+v}\le\sqrt u+\sqrt v$ for $u,v\ge0$, we obtain
\[
\Dtop(G,G_*)\le D_{K,\tau}(G,G_*)+4KM\sqrt{w_{\rm rare}(\tau)}.
\]
Substitute the upper bound for $D_{K,\tau}$ from Eq.~\eqref{eq:sparse-retained-aggregation} to obtain Eq.~\eqref{eq:topk-inverse-robust}.

\medskip\noindent\textbf{Step 4: A bound on the omitted probability.}
There are at most $\binom{k}{K}$ possible true active sets and the same number of fitted active sets. Therefore $|\mathcal R_+|\le\binom{k}{K}^2$. Each omitted region has probability less than $\tau$, so
\[
w_{\rm rare}(\tau)\le|\mathcal R_+\setminus\mathcal R_\tau|\tau
\le\binom{k}{K}^2\tau.
\]
Since the regions are disjoint, their total probability is also at most one. Combining the two bounds proves Eq.~\eqref{eq:topk-rare-mass}.
\end{proof}

% \begin{remark}[The square-root remainder]
% The loss is a quadratic average of the $D_C$. Thus a collection of regions of total probability $w$ and regionwise loss bounded by $4KM$ contributes at most $4KM\sqrt w$. This is the source of the remainder in Eq.~\eqref{eq:topk-inverse-robust}; it does not require any inverse estimate on the omitted regions.
% \end{remark}

\subsection{Proof of Corollary~\ref{cor:top1}}\label{app:top1proof}
\begin{proof}[Proof of Corollary~\ref{cor:top1}]
When $K=1$, the only active softmax weight is one. Fix $x$ and put
\[
u:=h_{i_G(x)}(x),\qquad v:=h_{i_*(x)}^*(x),\qquad z:=u-v.
\]
The Gaussian convolution identity in Lemma~\ref{lem:rkhsidentity} gives
\begin{equation}\label{eq:sparse-top1-Gaussian-distance}
\begin{aligned}
\|\phiSig(\cdot-u)-\phiSig(\cdot-v)\|_{L^2(dy)}^2
&=K_{\sigma_0}(u,u)+K_{\sigma_0}(v,v)-2K_{\sigma_0}(u,v)\\
&=\frac{1}{\sqrt{\pi\sigma_0}}
\left(1-e^{-z^2/(4\sigma_0)}\right).
\end{aligned}
\end{equation}
Using the variance convention in Eq.~\eqref{eq:gaussian-density} and the expert bound, $|z|\le2M$, so $0\le t:=z^2/(4\sigma_0)\le M^2/\sigma_0$. Using
\[
1-e^{-t}=\int_0^t e^{-s}\,ds,
\]
we obtain the uniform bounds
\[
e^{-M^2/\sigma_0}t\le1-e^{-t}\le t.
\]
Consequently,
\begin{equation}\label{eq:sparse-top1-pointwise}
c_-^2|u-v|^2
\le\|\phiSig(\cdot-u)-\phiSig(\cdot-v)\|_{L^2(dy)}^2
\le c_+^2|u-v|^2,
\end{equation}
where one explicit choice is
\[
c_+^2:=\frac{1}{4\sigma_0\sqrt{\pi\sigma_0}},
\qquad c_-^2:=e^{-M^2/\sigma_0}c_+^2>0.
\]
These inequalities include $u=v$. Substitute the selected expert values, integrate with respect to $P_X$, and use Tonelli's theorem to obtain
\[
c_-^2D_{\rm sel}(G,G_*)^2
\le\|p_G^{(1)}-p_{G_*}^{(1)}\|_{L^2(P_X\otimes dy)}^2
\le c_+^2D_{\rm sel}(G,G_*)^2.
\]
Taking square roots proves Eq.~\eqref{eq:top1-equivalence}. The comparison depends only on the two selected means and their common variance. It therefore requires no separation between latent experts.

On a positive-probability region $C=\mathcal R_{\{j\},\{i\}}$, the sole matching has zero gate error and
\[
D_C=\|h_i-h_j^*\|_{L^2(P_C)}.
\]
Multiplying its square by $w_C$ and summing over the partition proves $\Dtop^2=D_{\rm sel}^2$ for $K=1$.
\end{proof}

% \begin{remark}[Limits of Top-1 router identifiability]
% Under the routing convention in Section~\ref{sec:topk}, the selected label depends only on $\beta_{1i}^{\top}x$, while the sole active weight is one. Thus the intercepts $\beta_{0i}$ do not affect $p_G^{(1)}$. Corollary~\ref{cor:top1} identifies the selected expert function in the stated norm, not the raw router parameters. Additional identifiability conditions would be needed for a parameter-recovery conclusion.
% \end{remark}

\subsection{Retained-region geometry from conditional density and diameter bounds}
\begin{proposition}[Conditional density and diameter bounds imply the retained-region geometry]\label{prop:conditionalR1}
Under Assumption~\ref{ass:densesep}, fix $2\le K\le k$, and let $C=\mathcal R_{S,T}$ be a positive-probability region. Suppose, for fixed constants $C_{\rm ac}^{\rm cell}>0$ and $C_{\rm cl}^{\rm cell}<\infty$ and every distinct $j,\ell\in S$, the difference $Z_{j\ell}:=h_j^*(X)-h_\ell^*(X)$ under $X\sim P_C$ has a Lebesgue density $q_{j\ell\mid C}$ satisfying
\begin{equation}\label{eq:sparse-conditional-primitive}
\|q_{j\ell\mid C}\|_\infty\le C_{\rm ac}^{\rm cell}/\DeltaSep,
\qquad
\mathbb E_{P_C}|Z_{j\ell}|\le C_{\rm cl}^{\rm cell}\DeltaSep.
\end{equation}
Then the geometry of Assumption~\ref{ass:topR1} holds on $C$ with conditional probability at least $1/2$. If these constants are uniform over $\mathcal R_\tau$, that assumption holds uniformly with $\rho_\tau=1/2$.
\end{proposition}

\begin{proof}
Put $N_K:=\binom K2$ and choose
\[
a_0:=\min\{1,(8N_KC_{\rm ac}^{\rm cell})^{-1}\},
\qquad A_0:=4N_KC_{\rm cl}^{\rm cell}+1.
\]
For each active pair, integrating the density over an interval of length $2a_0\DeltaSep$ gives
\[
P_C(|Z_{j\ell}|<a_0\DeltaSep)\le2a_0 C_{\rm ac}^{\rm cell}.
\]
The union bound over the $N_K$ pairs bounds the probability of any such near collision by $2N_Ka_0C_{\rm ac}^{\rm cell}\le1/4$. Markov's inequality gives
\[
P_C(|Z_{j\ell}|>A_0\DeltaSep)\le C_{\rm cl}^{\rm cell}/A_0,
\]
so a second union bound makes the probability of any excessive diameter at most $N_KC_{\rm cl}^{\rm cell}/A_0\le1/4$. Therefore the measurable set
\[
E_C:=\bigcap_{\substack{j,\ell\in S\\j<\ell}}
\{x\in C:a_0\DeltaSep\le|h_j^*(x)-h_\ell^*(x)|\le A_0\DeltaSep\}
\]
has $P_C(E_C)\ge1/2$. Take $a_\tau=a_0$ and $A_\tau=A_0$. The formulae show the asserted dependence and uniformity of the constants.
\end{proof}

% \begin{remark}[Conditional bounds inherited from global geometry]\label{rem:top-cell-warning}
% Under Assumption~\ref{ass:denseac}, the conditional difference law on $C$ has a density bounded by $C_{\rm ac}/(w_C\Delta_{j\ell})\le C_{\rm ac}/(w_C\DeltaSep)$. Indeed, for every Borel set $H\subset\mathbb R$, its conditional probability is at most $w_C^{-1}\int_H q_{j\ell}(t)\,dt$. If the global single-cluster condition also holds, then
% \[
% \mathbb E_{P_C}|h_j^*-h_\ell^*|
% \le\frac{\|h_j^*-h_\ell^*\|_{L^1(P_X)}}{w_C}
% \le\frac{C_0\DeltaSep}{w_C}.
% \]
% Thus on $\mathcal R_\tau$ the proposition applies with $C_{\rm ac}^{\rm cell}=C_{\rm ac}/\tau$ and $C_{\rm cl}^{\rm cell}=C_0/\tau$. This is one sufficient route to Assumption~\ref{ass:topR1}; it is not an additional hypothesis of Theorem~\ref{thm:rare}. The expert part of Assumption~\ref{ass:topR2} still needs a separate verification; its gate part follows from Lemma~\ref{lem:affine-gate-restriction} at the required relative probability level. Without a positive lower bound on $w_C$, these conditional estimates need not be uniform.
% \end{remark}

\section{Proofs for Results in Section~\ref{sec:mle}}\label{app:mle}

This appendix proves the likelihood density rates and transfers them to the recovery losses. The dense argument uses a finite cover by pointwise upper functions. The sparse argument retains routing exactly, counts its realizable patterns on the sample, and covers the smooth expert and weight functions. Both arguments use likelihood affinity and finite-class probability bounds \citep{VandeGeer2000,BoucheronLugosiMassart2013}, avoiding regular likelihood expansions that may fail at mixture singularities \citep{GassiatVanHandel2013}.

The density bounds require compact parameter spaces, uniformly bounded expert values, covariate continuity, and the $L^\infty$ parameter regularity in Assumption~\ref{ass:expertregularity}. Analyticity enters only through covariate continuity. Separation, anti-concentration, and the inverse restriction assumptions are used only in the subsequent transfer to recovery. Estimators are assumed measurable; uniform empirical-process events are interpreted in outer probability when necessary.

\subsection{Proof of Lemma~\ref{lem:expert-entropy}}\label{app:entropyproof}

We construct uniform covers for the expert and logit classes. The covering argument accommodates compact parameter sets that need not be convex.

\begin{proof}[Proof of Lemma~\ref{lem:expert-entropy}]
By Assumption~\ref{ass:expertregularity}, the map $H(\eta):=h(\cdot,\eta)$ is continuously differentiable into $L^\infty(\cX)$ on an open neighborhood $U$ of $\Omega_\eta$. Assumption~\ref{ass:densecompact}\ref{cond:compactspaces} gives $r>0$ such that the compact neighborhood
\[
K_r:=\{\theta\in\R^{d_\eta}:\dist(\theta,\Omega_\eta)\le r\}
\]
is contained in $U$. Continuity of the derivative gives
\[
L_0:=\sup_{\theta\in K_r}\|DH(\theta)\|_{\mathrm{op}}<\infty.
\]
If $\eta,\eta'\in\Omega_\eta$ and $\|\eta-\eta'\|_2\le r$, their entire line segment lies in $K_r$. The fundamental theorem of calculus for Banach-space-valued maps yields
\[
\|H(\eta)-H(\eta')\|_{L^\infty}
\le L_0\|\eta-\eta'\|_2.
\]
If instead $\|\eta-\eta'\|_2>r$, boundedness gives
\[
\|H(\eta)-H(\eta')\|_{L^\infty}
\le2M\le(2M/r)\|\eta-\eta'\|_2.
\]
The two cases give a global Lipschitz constant $L_h:=\max\{L_0,2M/r\}$ on $\Omega_\eta$. Covariate continuity and full support of $P_X$ identify the essential supremum with the pointwise supremum, so
\begin{equation}\label{eq:expert-parameter-lipschitz}
\sup_{x\in\cX}|h(x,\eta)-h(x,\eta')|
\le L_h\|\eta-\eta'\|_2,
\qquad \eta,\eta'\in\Omega_\eta.
\end{equation}
In particular, $\eta\mapsto h(x,\eta)$ is continuous for every $x$.

For the entropy bound, enclose $\Omega_\eta$ in a fixed Euclidean ball. A maximal $u$-separated subset of $\Omega_\eta$ is a $u$-net with centers in $\Omega_\eta$. Comparing volumes of disjoint balls of radius $u/2$ shows that its cardinality is at most $(C/u)^{d_\eta}$ for $0<u\le1$, after enlarging a fixed $C$. Set
\[
u:=\frac{\varepsilon}{\max\{1,L_h\}}.
\]
By Eq.~\eqref{eq:expert-parameter-lipschitz}, the expert functions at these parameter-net points form a uniform $\varepsilon$-net. Increasing $A_h\ge e$ gives Eq.~\eqref{eq:expert-entropy} with $d_h=d_\eta$. 

For the logit class, Cauchy--Schwarz gives
\[
\sup_{x\in\cX}|(\beta_1-\beta_1')^\top x+\beta_0-\beta_0'|
\le\sqrt{1+d_xb^2}\,\| (\beta_1,\beta_0)-(\beta_1',\beta_0')\|_2.
\]
The same Euclidean net argument on the compact set $\Theta_\beta\subset\R^{d_x+1}$ proves Eq.~\eqref{eq:logit-entropy}. Neither argument uses expert identifiability, a lower bound on the parameter derivative, or expert separation.
\end{proof}

\subsection{Proof of Theorem~\ref{thm:densemle}}\label{app:densemleproof}

For the dense model, compactness and continuity give a measurable global MLE. A finite collection of upper functions then dominates the model densities while adding only a small amount of mass. The Hellinger affinity of these upper functions controls the likelihood event, yielding the finite-sample rate.

%\paragraph{Existence of the dense maximum likelihood estimator.}
\begin{lemma}[Hellinger-to-$L^2$ comparison for bounded conditional densities]\label{lem:hell-l2}
For any two conditional densities $f,g$ in the dense or Top-$K$ Gaussian MoE model of Sections~\ref{sec:intro-setup} and \ref{sec:topk},
\begin{equation}
\|f-g\|_{L^2(P_X\otimes dy)}^2
\le 8Q_{\max}\,\Hdist^2(f,g),
\qquad
Q_{\max}=(2\pi\sigma_0)^{-1/2}.
\label{eq:hellinger-density-conversion}
\end{equation}
\end{lemma}

\begin{proof}

Every conditional Gaussian mixture is bounded pointwise by the supremum of one Gaussian density:
\[
0\le f(y\mid x),g(y\mid x)\le Q_{\max}.
\]

Factor the difference as
\[
f-g=(\sqrt f-\sqrt g)(\sqrt f+\sqrt g).
\]
Therefore
\[
(f-g)^2
=(\sqrt f-\sqrt g)^2(\sqrt f+\sqrt g)^2
\le4Q_{\max}(\sqrt f-\sqrt g)^2.
\]

Integrating with respect to $P_X\otimes dy$ and recalling the convention
\[
\Hdist^2(f,g)=\frac12\int(\sqrt f-\sqrt g)^2dP_Xdy
\]
proves the stated factor $8Q_{\max}$.
\end{proof}

\begin{lemma}[Existence and measurable selection for the dense global MLE]\label{lem:dense-mle-existence}
Under Assumptions~\ref{ass:densecompact} and \ref{ass:expertregularity}, the dense model admits a Borel measurable estimator $\widehat G_n$ satisfying
\[
\ell_n(\widehat G_n)=\max_G\ell_n(G)
\]
for every sample.
\end{lemma}

\begin{proof}
By compactness of the parameter space and Lemma~\ref{lem:expert-entropy}, expert values vary continuously with their parameters. Softmax weights and Gaussian translates are continuous, and their mixture is strictly positive. Hence $G\mapsto\ell_n(G)$ is continuous for every sample and has a nonempty compact set of maximizers.

For fixed $G$, the criterion is Borel measurable in the sample. It is therefore a Carath\'eodory function on the sample space and the compact metric parameter space. The measurable maximum theorem gives a Borel measurable selection of the argmax correspondence.
\end{proof}

%\paragraph{Finite upper likelihood cover.}

\begin{lemma}[Dense upper likelihood cover]\label{lem:denseupper}
Under Assumptions~\ref{ass:densecompact} and \ref{ass:expertregularity}, for every $0<\eta\le1/16$ the dense model admits a finite pointwise upper cover $\{U_b:b\in\mathcal B_\eta\}$ of nonnegative measurable functions satisfying
\begin{equation}
1\le \int U_b(y\mid x)\,dy\le1+\eta
\quad b\in\mathcal B_\eta,\ x\in\cX,
\label{eq:upper-cover-mass-condition}
\end{equation}
and
\begin{equation}
\log|\mathcal B_\eta|
\le C d_{\rm den}\log(C/\eta).
\label{eq:upper-cover-cardinality}
\end{equation}
Here every $p_G$ is dominated pointwise by at least one $U_b$, and $C>0$ depends only on the fixed model class.
\end{lemma}

\begin{proof}
We approximate expert functions and logits uniformly, then multiply each template density by a small envelope that dominates every density assigned to that template. A Gaussian exponential-moment bound controls the excess mass required by Lemma~\ref{lem:sqrtupper}.

\medskip\noindent\textbf{Step 1: Expert and logit nets.}
Let $0<\delta\le1$, to be fixed as a constant multiple of $\eta$. Lemma~\ref{lem:expert-entropy} supplies a finite $\delta$-net $\mathcal N_h(\delta)$ for the expert family in $\|\cdot\|_\infty$ with
\begin{equation}
\log|\mathcal N_h(\delta)|
\le d_h\log(A_h/\delta).
\label{eq:expert-net-cardinality}
\end{equation}
Similarly, compactness of the affine-logit class gives a finite $\delta$-net $\mathcal N_\ell(\delta)$ with
\begin{equation}
\log|\mathcal N_\ell(\delta)|
\le(d_x+1)\log(A_\ell/\delta).
\label{eq:logit-net-cardinality}
\end{equation}
Choose one expert-net element and one logit-net element for each label. The Cartesian product of these choices forms a template collection with cardinality satisfying
\begin{equation}
\begin{aligned}
\log N_\delta
&\le k d_h\log(A_h/\delta)
+k(d_x+1)\log(A_\ell/\delta)\\
&\le d_{\rm den}\log(A/\delta),
\end{aligned}
\label{eq:dense-template-cardinality}
\end{equation}
where $A=\max(A_h,A_\ell,e)$ and $d_{\rm den}=k(d_h+d_x+1)$. Define
\[
\mathcal B_\eta:=\bigl(\mathcal N_h(\delta)\times\mathcal N_\ell(\delta)\bigr)^k,
\]
where Step~5 specifies $\delta=\delta(\eta)$. An index $b\in\mathcal B_\eta$ identifies a template $T$; we use $U_b$ and $U_T$ for the same upper function. In this construction, the scalar $\eta$ denotes excess-mass tolerance, while expert parameters retain their indexed notation from the model.

For every dense parameter $G$, choose a template $T=T(G)$ such that
\begin{equation}
\max_i\|h_i-\widetilde h_i\|_\infty\le\delta,
\qquad
\max_i\|\ell_i-\widetilde\ell_i\|_\infty\le\delta.
\label{eq:dense-template-approximation}
\end{equation}

\medskip\noindent\textbf{Step 2: Softmax weight approximation.}
For $z,w\in\mathbb R^k$, define
\[
\operatorname{LSE}(z)=\log\sum_{j=1}^ke^{z_j}.
\]
If $\|z-w\|_\infty\le\delta$, then
\[
e^{-\delta}\sum_je^{w_j}
\le\sum_je^{z_j}
\le e^\delta\sum_je^{w_j},
\]
so
\begin{equation}
|\operatorname{LSE}(z)-\operatorname{LSE}(w)|\le\delta.
\label{eq:log-sum-exp-lipschitz}
\end{equation}
Since
\[
\log\pi_i=\ell_i-\operatorname{LSE}(\ell_1,\ldots,\ell_k),
\]
equations \eqref{eq:dense-template-approximation}--\eqref{eq:log-sum-exp-lipschitz} imply
\begin{equation}
|\log\pi_i(x)-\log\widetilde\pi_i(x)|\le2\delta
\label{eq:dense-log-weight-approximation}
\end{equation}
for every $i,x$. Equivalently,
\begin{equation}
e^{-2\delta}\widetilde\pi_i(x)
\le\pi_i(x)
\le e^{2\delta}\widetilde\pi_i(x).
\label{eq:dense-weight-ratio}
\end{equation}

\medskip\noindent\textbf{Step 3: Gaussian mean shifts.}
For $u,\widetilde h\in[-M-1,M+1]$,
\begin{equation}
\begin{aligned}
\log\frac{\phi_{\sigma_0}(y-u)}{\phi_{\sigma_0}(y-\widetilde h)}
&=-\frac{(y-u)^2-(y-\widetilde h)^2}{2\sigma_0}\\
&=\frac{u-\widetilde h}{\sigma_0}
\left(y-\frac{u+\widetilde h}{2}\right).
\end{aligned}
\label{eq:gaussian-log-density-ratio}
\end{equation}
Under Eq.~\eqref{eq:dense-template-approximation}, $|u-\widetilde h|\le\delta$, and for $\delta\le1$ the two means lie in $[-M-1,M+1]$. Hence
\begin{equation}
\left|
\log\frac{\phi_{\sigma_0}(y-u)}{\phi_{\sigma_0}(y-\widetilde h)}
\right|
\le C_G\delta(1+|y|),
\label{eq:gaussian-log-ratio-bound}
\end{equation}
where one may take, for example,
\[
C_G=\sigma_0^{-1}(M+2).
\]
Only uniformity of this constant over the model class is needed.

\medskip\noindent\textbf{Step 4: A pointwise upper envelope.}
Combine Eqs.~\eqref{eq:dense-log-weight-approximation} and \eqref{eq:gaussian-log-ratio-bound}. After increasing a fixed constant $C_1$,
\begin{equation}
\pi_i(x)\phi_{\sigma_0}(y-h_i(x))
\le
\exp\{C_1\delta(1+|y|)\}
\widetilde\pi_i(x)\phi_{\sigma_0}(y-\widetilde h_i(x)).
\label{eq:component-upper-envelope}
\end{equation}
Summing the componentwise upper bounds gives
\begin{equation}
p_G(y\mid x)
\le U_T(y\mid x)
:=e^{C_1\delta(1+|y|)}\widetilde p_T(y\mid x).
\label{eq:dense-upper-envelope}
\end{equation}
Thus one upper function is associated with each smooth template.

\medskip\noindent\textbf{Step 5: Control of excess mass.}
Fix $x$ and let $Y$ have density $\widetilde p_T(\cdot\mid x)$. Conditional on a component index, $Y=V+\sqrt{\sigma_0}Z$, where $Z\sim N(0,1)$ and $|V|\le M+1$. Therefore
\begin{equation}
\begin{aligned}
\int U_T(y\mid x)dy
&=\E\exp\{C_1\delta(1+|Y|)\}\\
&\le\exp\{C_1\delta(M+2)\}
\E\exp\{C_1\delta\sqrt{\sigma_0}|Z|\}.
\end{aligned}
\label{eq:upper-envelope-mass}
\end{equation}
For $t\ge0$,
\begin{equation}
\E e^{t|Z|}
\le \E e^{tZ}+\E e^{-tZ}
=2e^{t^2/2}.
\label{eq:gaussian-absolute-mgf-bound}
\end{equation}
To control excess mass as $\delta\downarrow0$, we use the sharper identity
\begin{equation}
\E e^{t|Z|}=2e^{t^2/2}\Phi(t),
\label{eq:gaussian-absolute-mgf}
\end{equation}
which is continuously differentiable for $t$ in any compact interval and equals one at $t=0$. Hence there are fixed $\delta_0,c_0>0$ such that for $0\le\delta\le\delta_0$,
\begin{equation}
\sup_{x,T}\int U_T(y\mid x)dy
\le1+c_0\delta.
\label{eq:upper-envelope-excess-mass}
\end{equation}
Every template density integrates to one, and the envelope multiplier is at least one. Therefore $\int U_T(y\mid x)dy\ge1$, including for templates that are not assigned any original model density.

Choose
\begin{equation}
\delta=\min\{\delta_0,\eta/(2c_0)\}.
\label{eq:upper-cover-resolution}
\end{equation}
In either case, $c_0\delta\le\eta/2$, so Eq.~\eqref{eq:upper-envelope-excess-mass} gives
\begin{equation}
\int U_T(y\mid x)dy\le1+\eta.
\label{eq:upper-envelope-mass-bound}
\end{equation}

\medskip\noindent\textbf{Step 6: The upper-cover entropy.}
Since $\delta$ is a fixed constant multiple of $\eta$ in the nontrivial small-$\eta$ range, Eq.~\eqref{eq:dense-template-cardinality} yields
\begin{equation}
\log|\mathcal B_\eta|
\le C d_{\rm den}\log(C/\eta).
\label{eq:upper-cover-entropy-bound}
\end{equation}
This completes the construction.
\end{proof}

\begin{lemma}[Square-root likelihood tail from upper functions]\label{lem:sqrtupper}
Let $(X_a,Y_a)_{a=1}^n$ be iid with conditional density $p_{G_*}$ in a fitted conditional-density class, as in the dense sampling model of Section~\ref{sec:intro-setup}. Let $\widehat p_n$ be a measurable global likelihood maximizer in that class. Suppose that for every $0<\eta\le1/16$ there is a finite pointwise upper cover $\{U_b:b\in\mathcal B_\eta\}$ such that each model density is dominated by one $U_b$ and
\[
\sup_x\int U_b(y\mid x)dy\le1+\eta.
\]
Then for every $0<r\le1$,
\begin{equation}
\Pp_*
\{\Hdist(\widehat p_n,p_{G_*})>r\}
\le
|\mathcal B_{r^4/16}|\exp(-3nr^2/4).
\label{eq:square-root-likelihood-tail}
\end{equation}
\end{lemma}

The lemma applies to any correctly specified conditional-density class with the stated upper covers; Gaussian structure is used only to construct those covers.

\begin{proof}
Set
\begin{equation}
\eta=r^4/16.
\label{eq:likelihood-cover-tolerance}
\end{equation}
The proof bounds the affinity of an upper function whose cover cell contains a density at Hellinger distance greater than $r$. Global likelihood optimality then reduces the estimation event to a finite union of likelihood-ratio events.

\medskip\noindent\textbf{Step 1: Cover elements containing separated densities.}
Associate with each upper function $U_b$ the set of model densities it dominates according to the cover construction. Call the index $b$ \emph{separated} if its cell contains a density $f_b$ satisfying
\begin{equation}
\Hdist(f_b,p_{G_*})>r.
\label{eq:separated-cover-representative}
\end{equation}
For every separated $b$, choose one such $f_b$ only for the following deterministic calculation. The set of separated indices depends on the model class, $p_{G_*}$, and $r$, but not on the sample.

\medskip\noindent\textbf{Step 2: Distance to a separated representative.}
Because $U_b\ge f_b$ pointwise,
\begin{equation}
(\sqrt{U_b}-\sqrt{f_b})^2
\le U_b-f_b.
\label{eq:square-root-domination}
\end{equation}
Integrating first in $y$ and then in $x$,
\begin{equation}
\begin{aligned}
\|\sqrt{U_b}-\sqrt{f_b}\|_{L^2(P_X\otimes dy)}^2
&\le\int\left\{\int(U_b-f_b)dy\right\}dP_X(x)\\
&\le\eta,
\end{aligned}
\label{eq:upper-cover-root-distance}
\end{equation}
because $\int f_b(y\mid x)dy=1$ and the upper function has conditional mass at most $1+\eta$. Hence
\begin{equation}
\|\sqrt{U_b}-\sqrt{f_b}\|_2\le\sqrt\eta=r^2/4.
\label{eq:upper-cover-root-radius}
\end{equation}

\medskip\noindent\textbf{Step 3: The affinity deficit.}
By definition,
\begin{equation}
\int\sqrt{f_bp_{G_*}}\,dP_Xdy
=1-\Hdist^2(f_b,p_{G_*})
<1-r^2.
\label{eq:representative-affinity-deficit}
\end{equation}
Also $\|\sqrt{p_{G_*}}\|_2=1$. Therefore Cauchy--Schwarz and Eq.~\eqref{eq:upper-cover-root-radius} give
\begin{equation}
\begin{aligned}
\int\sqrt{U_bp_{G_*}}
&=\int\sqrt{f_bp_{G_*}}
+\int(\sqrt{U_b}-\sqrt{f_b})\sqrt{p_{G_*}}\\
&\le1-r^2+\|\sqrt{U_b}-\sqrt{f_b}\|_2\\
&\le1-r^2+r^2/4
=1-3r^2/4.
\end{aligned}
\label{eq:upper-cover-affinity-deficit}
\end{equation}
The excess-mass condition controls the distance from $U_b$ to the separated representative. Thus the affinity estimate applies directly to $U_b$, without normalizing it to a density.

\medskip\noindent\textbf{Step 4: Containment of the likelihood event.}
Correct specification means that $p_{G_*}$ belongs to the likelihood class. Thus global optimality gives
\begin{equation}
\prod_{a=1}^n\widehat p_n(Y_a\mid X_a)
\ge
\prod_{a=1}^np_{G_*}(Y_a\mid X_a).
\label{eq:mle-likelihood-product}
\end{equation}
Suppose $\Hdist(\widehat p_n,p_{G_*})>r$. The cover cell containing $\widehat p_n$ is then separated by definition, and its upper function $U_b$ satisfies
\[
U_b(y\mid x)\ge\widehat p_n(y\mid x)
\]
for all $(x,y)$. Consequently
\begin{equation}
\prod_{a=1}^n\frac{U_b(Y_a\mid X_a)}{p_{G_*}(Y_a\mid X_a)}
\ge 1.
\label{eq:upper-cover-likelihood-event}
\end{equation}
Therefore the estimation-error event is contained in the finite union of the events in Eq.~\eqref{eq:upper-cover-likelihood-event} over separated indices $b$.

\medskip\noindent\textbf{Step 5: The square-root likelihood moment.}
For fixed $b$, define
\begin{equation}
Z_b
=\prod_{a=1}^n
\left\{\frac{U_b(Y_a\mid X_a)}{p_{G_*}(Y_a\mid X_a)}\right\}^{1/2}.
\label{eq:square-root-likelihood-ratio}
\end{equation}
The observations $(X_a,Y_a)$ are iid under the truth. Hence
\begin{equation}
\begin{aligned}
\E_*Z_b
&=\prod_{a=1}^nE_*
\left[\left\{\frac{U_b(Y_a\mid X_a)}{p_{G_*}(Y_a\mid X_a)}\right\}^{1/2}\right]\\
&=\left\{
\int\sqrt{U_b(y\mid x)p_{G_*}(y\mid x)}\,dy\,dP_X(x)
\right\}^n\\
&\le(1-3r^2/4)^n
\le e^{-3nr^2/4},
\end{aligned}
\label{eq:square-root-likelihood-moment}
\end{equation}
where the final inequality uses $1-z\le e^{-z}$. The event \eqref{eq:upper-cover-likelihood-event} is exactly $\{Z_b\ge 1\}$. Markov's inequality therefore gives
\begin{equation}
\Pp_*(Z_b\ge 1)\le e^{-3nr^2/4}.
\label{eq:upper-cover-likelihood-tail}
\end{equation}

\medskip\noindent\textbf{Step 6: A union bound over the cover.}
There are at most $|\mathcal B_\eta|$ separated indices. Combining the preceding event containment with Eq.~\eqref{eq:upper-cover-likelihood-tail},
\[
\Pp_*\{\Hdist(\widehat p_n,p_{G_*})>r\}
\le|\mathcal B_\eta|e^{-3nr^2/4}.
\]
Substitute $\eta=r^4/16$ from Eq.~\eqref{eq:likelihood-cover-tolerance} to prove the lemma.
\end{proof}

%\paragraph{Proof of the dense density rate.}

\begin{proof}[Proof of Theorem~\ref{thm:densemle}]
Let $\widehat p_n=p_{\widehat G_n}$. Lemma~\ref{lem:dense-mle-existence} supplies the measurable global maximizer used in the probability statement.

\medskip\noindent\textbf{Step 1: Entropy and the likelihood tail.}
Lemmas~\ref{lem:denseupper} and \ref{lem:sqrtupper} imply that for every $0<r\le1$,
\begin{equation}
\begin{aligned}
\log \Pp_*
\{\Hdist(\widehat p_n,p_{G_*})>r\}
&\le
\log|\mathcal B_{r^4/16}|-\frac34nr^2\\
&\le
C_0d_{\rm den}\log\left(\frac{C_1}{r^4}\right)
-\frac34nr^2.
\end{aligned}
\label{eq:dense-entropy-tail-bound}
\end{equation}
All constants depend only on the fixed model class.

\medskip\noindent\textbf{Step 2: Choice of the Hellinger radius.}
Define
\begin{equation}
R_n(t)
=C_2\frac{d_{\rm den}\log(C_3n)+t}{n}.
\label{eq:dense-tail-radius}
\end{equation}
If $R_n(t)\ge1$, the event $\Hdist^2(\widehat p_n,p_{G_*})>R_n(t)$ is empty because squared Hellinger distance is at most one. It remains to consider
\begin{equation}
R_n(t)<1
\label{eq:dense-nontrivial-radius}
\end{equation}
and set
\begin{equation}
r=\sqrt{R_n(t)}.
\label{eq:dense-hellinger-radius}
\end{equation}
By increasing $C_2$ if necessary and using $t\ge1$, we may assume
\begin{equation}
r^2\ge C_2/n,
\qquad\text{hence}\qquad
r\ge n^{-1/2}.
\label{eq:dense-radius-lower-bound}
\end{equation}
Therefore
\begin{equation}
\log(r^{-4})\le2\log n.
\label{eq:dense-radius-log-bound}
\end{equation}
Since $C_3n\ge e$, equations \eqref{eq:dense-radius-lower-bound}--\eqref{eq:dense-radius-log-bound} give
\begin{equation}
\log(C_1/r^4)
\le C_4\log(C_3n).
\label{eq:dense-entropy-at-radius}
\end{equation}

\medskip\noindent\textbf{Step 3: Control of the tail exponent.}
Substitute Eqs.~\eqref{eq:dense-tail-radius}--\eqref{eq:dense-entropy-at-radius} into Eq.~\eqref{eq:dense-entropy-tail-bound}:
\begin{equation}
\begin{aligned}
\log \Pp_*
\{\Hdist(\widehat p_n,p_{G_*})>r\}
&\le
C_0C_4d_{\rm den}\log(C_3n)\\
&\quad-\frac34C_2\{d_{\rm den}\log(C_3n)+t\}.
\end{aligned}
\label{eq:dense-tail-exponent}
\end{equation}
Choose $C_2$ large enough that
\begin{equation}
\frac34C_2\ge C_0C_4+1
\qquad\text{and}\qquad
\frac34C_2\ge1.
\label{eq:dense-tail-constant-choice}
\end{equation}
Then the right side of Eq.~\eqref{eq:dense-tail-exponent} is at most $-t$. Thus
\begin{equation}
\Pp_*
\left\{
\Hdist^2(\widehat p_n,p_{G_*})
>C_2\frac{d_{\rm den}\log(C_3n)+t}{n}
\right\}
\le e^{-t}.
\label{eq:dense-final-hellinger-tail}
\end{equation}
This is the finite-sample Hellinger claim.

\medskip\noindent\textbf{Step 4: Conversion to the density norm.}
Lemma~\ref{lem:hell-l2} gives deterministically
\begin{equation}
\|\widehat p_n-p_{G_*}\|_{L^2(P_X\otimes dy)}
\le\sqrt{8Q_{\max}}\,
\Hdist(\widehat p_n,p_{G_*}).
\label{eq:dense-mle-density-conversion}
\end{equation}
Hence the same root-$n$ order, with the same logarithmic complexity factor, holds for the density norm used by the inverse theorems.
\end{proof}

% \begin{remark}[Density estimation without Fisher-information regularity]
% The finite-cover argument uses Hellinger affinity and does not require a score expansion or an invertible information matrix. The density bound consequently remains valid near expert collisions, where functional inversion can deteriorate.
% \end{remark}

\subsection{Proof of Corollary~\ref{cor:densemle-recovery}}\label{app:mletransfer}
The deterministic inverse inequalities apply directly to the likelihood estimators under the stated uniform hypotheses. The weight floors then convert the labelled functional losses into individual-expert errors.

\begin{proof}[Proof of Corollary~\ref{cor:densemle-recovery}]
By Theorem~\ref{thm:densemle} and Lemma~\ref{lem:hell-l2},
\[
\|p_{\widehat G_n}-p_{G_*}\|_{L^2(P_X\otimes dy)}=\mathcal{O}_P(\varepsilon_{n,\rm den}).
\]
The inverse hypotheses hold uniformly over the fitted class and hence apply to the estimator on every sample outcome. Theorem~\ref{thm:densemcfunc} gives
\[
\Dmc(\widehat G_n,G_*)
\le C_{\rm cluster}^{-1}\DeltaSep^{-(2\smax-2)}
\|p_{\widehat G_n}-p_{G_*}\|_{L^2(P_X\otimes dy)}.
\]
This proves the clustered loss rate, including its $q=1$ specialization. To obtain the expert rate in part~(i), choose the first minimizer $\widehat\sigma$ of the clustered loss in a fixed ordering of the admissible permutations. For every true label $j$, the weight floor and $\|f\|_{L^1(P_X)}\le\|f\|_{L^2(P_X)}$ give
\[
\begin{aligned}
\|\widehat h_{\widehat\sigma(j)}-h_j^*\|_{L^1(P_X)}
&\le\underline\pi_k^{-1}
\|\pi_j^*(\widehat h_{\widehat\sigma(j)}-h_j^*)\|_{L^2(P_X)}\\
&\le\underline\pi_k^{-1}\Dmc(\widehat G_n,G_*).
\end{aligned}
\]
Here $\widehat h_i=h(\cdot,\widehat\eta_i)$. Taking the maximum over $j$ and minimizing over the admissible permutations proves Eq.~\eqref{eq:dense-clustered-expert-rate}. In fact, the same bound already holds under the loss-minimizing permutation $\widehat\sigma$.

For part~(ii), Theorem~\ref{thm:denseU} gives directly
\[
\Dfunc(\widehat G_n,G_*)
\le C_{\rm U,func}^{-1}\DeltaSep^{-(2k-2)}
\|p_{\widehat G_n}-p_{G_*}\|_{L^2(P_X\otimes dy)}
=\mathcal O_P\!\left(C_{\rm U,func}^{-1}\DeltaSep^{-(2k-2)}\varepsilon_{n,\rm den}\right).
\]
This proves Eq.~\eqref{eq:dense-unstructured-functional-rate}. Choose a minimizing permutation for $\Dfunc$ in a fixed ordering. The same weight-floor argument as in part~(i) gives
\[
\mathcal E_{\Sigma_k}(\widehat G_n,G_*)
\le\underline\pi_k^{-1}\Dfunc(\widehat G_n,G_*),
\]
proving Eq.~\eqref{eq:dense-unstructured-expert-rate}. These inequalities hold for every fitted tuple, so no event localizing transport error relative to separation is needed. For a restricted likelihood class, use transport restriction on its ambient product class or the direct expert-restriction version of the inverse proof. Uniform constants give the same conclusions along sequences of truths.

\end{proof}

\subsection{Proof of Lemma~\ref{lem:routingtrace}}\label{app:routingtraceproof}

\begin{proof}[Proof of Lemma~\ref{lem:routingtrace}]
Fix covariates $x_1,\ldots,x_m$. Each realizable sequence of active sets is determined by pairwise score comparisons evaluated at these points. We bound their number by counting faces in router-parameter space, including faces corresponding to ties.

\medskip\noindent\textbf{Step 1: Routing decisions as pairwise comparisons.}
At covariate $x_a$, the routing scores are
\[
r_i(x_a)=\beta_{1i}^\top x_a,
\qquad i\in[k].
\]
The complete weak ordering of these $k$ real numbers is determined by the signs
\begin{equation}
\operatorname{sgn}\{(\beta_{1i}-\beta_{1j})^\top x_a\},
\qquad 1\le i<j\le k.
\label{eq:routing-comparison-signs}
\end{equation}
Once the weak ordering is known, the fixed deterministic priority rule resolves all equalities and therefore determines the unique Top-$K$ subset. Thus it is enough to count realizable simultaneous sign patterns of all pairwise comparisons at all $m$ covariates.

\medskip\noindent\textbf{Step 2: Hyperplanes in parameter space.}
Concatenate the $k$ slope vectors into
\begin{equation}
\vartheta=(\beta_{11},\ldots,\beta_{1k})\in\mathbb R^{v},
\qquad v=kd_x.
\label{eq:concatenated-router-parameters}
\end{equation}
For a triple $(a,i,j)$ with $i<j$, the comparison in Eq.~\eqref{eq:routing-comparison-signs} is the sign of the homogeneous linear functional
\begin{equation}
L_{aij}(\vartheta)
=(\beta_{1i}-\beta_{1j})^\top x_a.
\label{eq:routing-linear-comparison}
\end{equation}
Its zero set is a hyperplane in $\mathbb R^v$, possibly the entire space if $x_a=0$; such degenerate comparisons can only reduce the number of patterns and may be discarded. There are at most
\begin{equation}
N=mq_k,
\qquad q_k=\binom{k}{2},
\label{eq:routing-hyperplane-count}
\end{equation}
nontrivial hyperplanes.

\medskip\noindent\textbf{Step 3: Counting faces, including ties.}
Routing ties correspond to parameter values lying on one or more comparison hyperplanes. Counting only open regions would therefore miss some active-set patterns. Let $F(N,v)$ denote the maximum total number of relatively open faces of all dimensions in an arrangement of $N$ hyperplanes in $\mathbb R^v$.

Add the $N$th hyperplane $H$. Every old face not intersected in its relative interior remains one face. Every old face cut by $H$ is split into two faces, while its intersection with $H$ becomes a face of the induced arrangement on $H\cong\mathbb R^{v-1}$. A coarse recurrence sufficient for our purposes is
\begin{equation}
F(N,v)
\le F(N-1,v)+2F(N-1,v-1),
\label{eq:arrangement-face-recursion}
\end{equation}
with $F(0,v)=1$ and $F(N,0)=1$. Induction on $N+v$ gives
\begin{equation}
F(N,v)
\le\sum_{r=0}^{v}2^r\binom Nr.
\label{eq:arrangement-face-count}
\end{equation}
Indeed, Pascal's identity transforms the right side according to the same recurrence. Finally,
\begin{equation}
\sum_{r=0}^{v}2^r\binom Nr
\le\sum_{r=0}^{v}(2N)^r
\le(1+2N)^v,
\label{eq:arrangement-polynomial-bound}
\end{equation}
which provides the polynomial bound needed for the entropy calculation.

\medskip\noindent\textbf{Step 4: The active-set pattern bound.}
On the relative interior of one face, every linear form $L_{aij}$ has a fixed sign in $\{-1,0,+1\}$. Therefore every pairwise comparison at every covariate is fixed, including the equality information. The deterministic priority rule then fixes
\[
(S_G(x_1),\ldots,S_G(x_m)).
\]
Consequently the number of active-set patterns is at most the number of faces:
\begin{equation}
\Gamma_K(m)
\le F(mq_k,kd_x)
\le(1+2mq_k)^{kd_x}.
\label{eq:routing-pattern-bound-proof}
\end{equation}
If $K=k$, every active set is identically $[k]$ regardless of the router and $\Gamma_K(m)=1$.
\end{proof}

% \begin{remark}[Complexity of realizable routing patterns]
% There are $\binom{k}{K}^m$ arbitrary sequences of active sets on $m$ covariates, but a common affine router realizes only polynomially many. This restriction, rather than independent choices at each covariate, determines the routing contribution to the likelihood bound.
% \end{remark}

\subsection{Proof of Theorem~\ref{thm:topmle}}\label{app:topmleproof}

Write $p_\circ:=p_{G_*}^{(K)}$ for the true conditional density in this subsection. Probabilities and expectations use the corresponding sampling law $\Pp_*^{(K)}$. Using the routing-pattern bound of Lemma~\ref{lem:routingtrace}, we approximate the expert values and active weights with finite smooth templates. Conditional likelihood control yields an empirical Hellinger bound, and a relative empirical inequality transfers it to the population loss.

Fix $0<\delta\le1$. Let $\mathcal T_\delta$ be the product of a $\delta$-net for each expert function and, when $K\ge2$, a $\delta$-net for each affine logit. Then
\begin{equation}
\log|\mathcal T_\delta|
\le d_S\log(A/\delta).
\label{eq:sparse-template-cardinality}
\end{equation}
For an admissible routing map $R:x\mapsto S_R(x)$ and a smooth template $T$, define the enlarged conditional density
\begin{equation}
g_{R,T}(y\mid x)
=
\sum_{i\in S_R(x)}
\frac{e^{\widetilde\ell_{i,T}(x)}}{\sum_{q\in S_R(x)}e^{\widetilde\ell_{q,T}(x)}}
\phiSig(y-\widetilde h_{i,T}(x)).
\label{eq:enlarged-routed-density}
\end{equation}
For $K=1$, the displayed weight is interpreted as one. The covering construction allows $R$ and the smooth logit template to vary separately, so $g_{R,T}$ need not belong to the original model. This enlarged class is used to approximate likelihoods; the MLE remains defined over the original fitted class.

\begin{lemma}[Smooth approximation at fixed routing]\label{lem:topsmooth}
Under Assumptions~\ref{ass:densecompact} and \ref{ass:expertregularity}, for every $0<\delta\le1$ and every Top-$K$ density $p_G^{(K)}$, there is a smooth template $T\in\mathcal T_\delta$ such that
\begin{equation}
\left|\log p_G^{(K)}(y\mid x)-\log g_{R_G,T}(y\mid x)\right|
\le C\delta(1+|y|)
\label{eq:fixed-routing-log-approximation}
\end{equation}
for all $x,y$, and
\begin{equation}
\Hdist^2(p_G^{(K)},g_{R_G,T})\le C\delta.
\label{eq:fixed-routing-hellinger-approximation}
\end{equation}
Here $C>0$ depends only on the fixed model bounds.
\end{lemma}

\begin{proof}
Choose $T$ from the product net so that
\begin{equation}
\max_i\|h_i-\widetilde h_i\|_\infty\le\delta
\label{eq:sparse-expert-net-approximation}
\end{equation}
and, when $K\ge2$,
\begin{equation}
\max_i\|\ell_i-\widetilde\ell_i\|_\infty\le\delta.
\label{eq:sparse-logit-net-approximation}
\end{equation}
For $K=1$ no logit approximation is needed because the only active weight is one. The following argument covers $K\ge2$; omitting the weight comparisons gives the $K=1$ case.

\medskip\noindent\textbf{Step 1: Exact routing.}
For every covariate $x$, both $p_G^{(K)}(\cdot\mid x)$ and its approximation use the same active set
\begin{equation}
S=S_G(x).
\label{eq:common-active-set}
\end{equation}
The common active set permits uniform approximation of the component values and weights without perturbing a routing decision.

\medskip\noindent\textbf{Step 2: Active softmax weight approximation.}
For $z,w\in\mathbb R^K$, the argument in Eq.~\eqref{eq:log-sum-exp-lipschitz} gives
\[
|\operatorname{LSE}(z)-\operatorname{LSE}(w)|\le\|z-w\|_\infty.
\]
On the fixed active set $S$, Eq.~\eqref{eq:sparse-logit-net-approximation} therefore implies
\begin{equation}
|\log\pi_{i,G}^{(K)}(x)-\log\widetilde\pi_i(x)|
\le2\delta,
\qquad i\in S.
\label{eq:sparse-log-weight-approximation}
\end{equation}

\medskip\noindent\textbf{Step 3: Control of the log-density ratio.}
Equation \eqref{eq:gaussian-log-density-ratio} remains valid for the active expert means. Hence, uniformly in $x,y$ and active $i$,
\begin{equation}
\left|
\log\frac{\phi_{\sigma_0}(y-h_i(x))}
{\phi_{\sigma_0}(y-\widetilde h_i(x))}
\right|
\le C_G\delta(1+|y|).
\label{eq:sparse-gaussian-log-ratio}
\end{equation}
Combining Eqs.~\eqref{eq:sparse-log-weight-approximation}--\eqref{eq:sparse-gaussian-log-ratio}, for every active component
\begin{equation}
e^{-C\delta(1+|y|)}
\le
\frac{
\pi_{i,G}^{(K)}(x)\phi_{\sigma_0}(y-h_i(x))
}{
\widetilde\pi_i(x)\phi_{\sigma_0}(y-\widetilde h_i(x))
}
\le
e^{C\delta(1+|y|)}.
\label{eq:sparse-component-ratio}
\end{equation}
If positive numbers $a_i,b_i$ satisfy $e^{-c}\le a_i/b_i\le e^c$ for every $i$, then the same inequality holds for $(\sum_i a_i)/(\sum_i b_i)$. Applying this observation to Eq.~\eqref{eq:sparse-component-ratio} gives
\[
e^{-C\delta(1+|y|)}
\le\frac{p_G^{(K)}(y\mid x)}{g_{R_G,T}(y\mid x)}
\le e^{C\delta(1+|y|)},
\]
which is exactly Eq.~\eqref{eq:fixed-routing-log-approximation}.

\medskip\noindent\textbf{Step 4: An $\ell_1$ bound for softmax weights.}
Let $z,w\in\mathbb R^K$ with $\|z-w\|_\infty\le\delta$, and define
\[
z(t)=w+t(z-w),\qquad 0\le t\le1.
\]
Let $\pi(t)=\operatorname{softmax}\{z(t)\}$ and $h=z-w$. Differentiation gives
\begin{equation}
\frac{d}{dt}\pi_i(t)
=\pi_i(t)\left(h_i-\sum_q\pi_q(t)h_q\right).
\label{eq:softmax-path-derivative}
\end{equation}
Therefore
\begin{equation}
\begin{aligned}
\sum_i\left|\frac{d}{dt}\pi_i(t)\right|
&\le\sum_i\pi_i(t)|h_i|
+\left|\sum_q\pi_q(t)h_q\right|\sum_i\pi_i(t)\\
&\le2\|h\|_\infty
\le2\delta.
\end{aligned}
\label{eq:softmax-derivative-sum}
\end{equation}
Integrating in $t$,
\begin{equation}
\sum_{i\in S}|\pi_{i,G}^{(K)}(x)-\widetilde\pi_i(x)|
\le2\delta.
\label{eq:softmax-lone-approximation}
\end{equation}

\medskip\noindent\textbf{Step 5: An $L_1$ bound for Gaussian translates.}
In this step, $\|\cdot\|_1$ denotes $L^1(dy)$. The derivative of a Gaussian translate with respect to its mean is
\[
\frac{\partial}{\partial u}\phi_{\sigma_0}(y-u)
=\frac{y-u}{\sigma_0}\phi_{\sigma_0}(y-u).
\]
Its $L^1(dy)$ norm is
\begin{equation}
\int\left|\frac{\partial}{\partial u}\phi_{\sigma_0}(y-u)\right|dy
=\frac{\E|N(0,\sigma_0)|}{\sigma_0}
=\sqrt{\frac{2}{\pi\sigma_0}}.
\label{eq:gaussian-mean-derivative-norm}
\end{equation}
By the fundamental theorem of calculus in the mean parameter,
\begin{equation}
\|\phi_{\sigma_0}(\cdot-u)-\phi_{\sigma_0}(\cdot-v)\|_1
\le\sqrt{\frac{2}{\pi\sigma_0}}|u-v|.
\label{eq:gaussian-translate-lone-bound}
\end{equation}

\medskip\noindent\textbf{Step 6: The conditional $L_1$ density bound.}
On the common active set $S$,
\begin{equation}
\begin{aligned}
\|p_G^{(K)}(\cdot\mid x)-g_{R_G,T}(\cdot\mid x)\|_1
&\le
\sum_{i\in S}|\pi_{i,G}^{(K)}-\widetilde\pi_i|\|\phi_{\sigma_0}(\cdot-h_i)\|_1\\
&\quad+
\sum_{i\in S}\widetilde\pi_i
\|\phi_{\sigma_0}(\cdot-h_i)-\phi_{\sigma_0}(\cdot-\widetilde h_i)\|_1\\
&\le2\delta+
\sqrt{\frac{2}{\pi\sigma_0}}\delta
\le C\delta.
\end{aligned}
\label{eq:fixed-routing-lone-approximation}
\end{equation}
The bound is uniform in $x$.

\medskip\noindent\textbf{Step 7: Conversion to Hellinger distance.}
For any two nonnegative numbers $a,b$,
\[
(\sqrt a-\sqrt b)^2\le|a-b|.
\]
Therefore, under the paper's Hellinger convention,
\[
\Hdist^2(p_G^{(K)},g_{R_G,T})
\le\frac12\int|p_G^{(K)}-g_{R_G,T}|dP_Xdy
\le C\delta.
\]
This proves Eq.~\eqref{eq:fixed-routing-hellinger-approximation}.
\end{proof}

% \begin{remark}[Role of the enlarged approximation class]
% The class $\{g_{R,T}\}$ decouples the exact router from the smooth logit template. Its elements need not belong to the original MoE model. They are used only in the likelihood comparison; the estimator is defined over the original fitted class.
% \end{remark}

For a conditional density $g$, define
\begin{equation}
L_n(g)=\sum_{a=1}^n\log\frac{g(Y_a\mid X_a)}{p_\circ(Y_a\mid X_a)},
\label{eq:sparse-log-likelihood-ratio}
\end{equation}
\begin{equation}
h_x^2(g,p_\circ)
=1-\int\sqrt{g(y\mid x)p_\circ(y\mid x)}dy,
\label{eq:conditional-hellinger-discrepancy}
\end{equation}
and
\begin{equation}
\Hdist_n^2(g,p_\circ)=\frac1n\sum_{a=1}^nh_{X_a}^2(g,p_\circ).
\label{eq:empirical-conditional-hellinger}
\end{equation}

\begin{lemma}[Conditional square-root likelihood inequality]\label{lem:conditionalLR}
Under the Top-$K$ sampling model of Section~\ref{sec:intro-setup}, for fixed $g$ and fixed covariates $X_{1:n}$,
\begin{equation}
\E_*^{(K)}\left[e^{L_n(g)/2}\mid X_{1:n}\right]
\le e^{-n\Hdist_n^2(g,p_\circ)}.
\label{eq:conditional-likelihood-moment}
\end{equation}
Consequently, for $a,v\ge0$,
\begin{equation}
\Pp_*^{(K)}\left\{L_n(g)\ge-a,\ n\Hdist_n^2(g,p_\circ)>v\mid X_{1:n}\right\}
\le e^{a/2-v}.
\label{eq:conditional-likelihood-tail}
\end{equation}
\end{lemma}

\begin{proof}
\medskip\noindent\textbf{Step 1: The conditional likelihood moment.}
Conditional independence of the responses gives
\[
\begin{aligned}
\E_*^{(K)}[e^{L_n(g)/2}\mid X_{1:n}]
&=\prod_{a=1}^n\int\sqrt{g(y\mid X_a)p_\circ(y\mid X_a)}dy\\
&=\prod_{a=1}^n\{1-h_{X_a}^2(g,p_\circ)\}\\
&\le\exp\left\{-\sum_{a=1}^nh_{X_a}^2(g,p_\circ)\right\},
\end{aligned}
\]
which is Eq.~\eqref{eq:conditional-likelihood-moment}.

\medskip\noindent\textbf{Step 2: The conditional tail bound.}
On $L_n(g)\ge-a$, $e^{L_n(g)/2}\ge e^{-a/2}$. Markov's inequality and Eq.~\eqref{eq:conditional-likelihood-moment} prove Eq.~\eqref{eq:conditional-likelihood-tail}.
\end{proof}

For fixed covariates $X_{1:n}$, let $\mathscr R_{X}$ be the set of realizable active-set patterns
\begin{equation}
\mathscr R_X
=\left\{(S_R(X_1),\ldots,S_R(X_n)):R\text{ admissible}\right\}.
\label{eq:observed-routing-patterns}
\end{equation}

For the enlarged approximation class $\{g_{R,T}\}$, set
\begin{equation}
N_X:=|\mathcal T_\delta|\,|\mathscr R_X|
\le|\mathcal T_\delta|\Gamma_K(n).
\label{eq:conditional-class-cardinality}
\end{equation}
The proof below bounds the number of distinct conditional-density vectors on $X_{1:n}$ by $N_X$.

\begin{lemma}[Uniform likelihood control over the finite conditional pattern class]\label{lem:finite-trace-lr}
Under the Top-$K$ sampling model of Section~\ref{sec:intro-setup}, fix the covariates $X_{1:n}$ and the finite template class constructed under Assumptions~\ref{ass:densecompact} and \ref{ass:expertregularity}. For every $a,t\ge0$, conditionally on $X_{1:n}$,
\begin{equation}
\begin{aligned}
\Pp_*^{(K)}
\Big\{\exists g\in\{g_{R,T}\}:\;&L_n(g)\ge-a,\\
&n\Hdist_n^2(g,p_\circ)
>\log N_X+a/2+t
\ \Bigm|\ X_{1:n}\Big\}
\le e^{-t}.
\end{aligned}
\label{eq:conditional-pattern-likelihood-tail}
\end{equation}
\end{lemma}

Consequently the same statement remains true after replacing $\log N_X$ by the deterministic upper bound
\begin{equation}
\log|\mathcal T_\delta|+\log\Gamma_K(n).
\label{eq:deterministic-pattern-entropy}
\end{equation}

\begin{proof}
\medskip\noindent\textbf{Step 1: Reduction to observed routing patterns.}
Fix a smooth template $T$. Suppose routers $R$ and $R'$ satisfy
\begin{equation}
S_R(X_a)=S_{R'}(X_a),
\qquad a=1,\ldots,n.
\label{eq:coincident-routing-patterns}
\end{equation}
By construction of the enlarged density $g_{R,T}$, once $T$ is fixed the router enters only through the active subset. Therefore
\begin{equation}
g_{R,T}(\cdot\mid X_a)
=g_{R',T}(\cdot\mid X_a)
\qquad\text{for every }a.
\label{eq:coincident-conditional-densities}
\end{equation}
Thus a conditional-density vector is completely determined by a pair consisting of one smooth template and one active-set pattern, proving Eq.~\eqref{eq:conditional-class-cardinality}.

\medskip\noindent\textbf{Step 2: The pattern-template density vectors.}
For an active-set pattern
\[
s=(S_1,\ldots,S_n)\in\mathscr R_X
\]
and a template $T$, define directly
\begin{equation}
g_{s,T,a}(y)
=
\sum_{i\in S_a}
\frac{e^{\widetilde\ell_{i,T}(X_a)}}
{\sum_{q\in S_a}e^{\widetilde\ell_{q,T}(X_a)}}
\phi_{\sigma_0}(y-\widetilde h_{i,T}(X_a)).
\label{eq:pattern-template-density}
\end{equation}
Every router realizing $s$ gives this conditional-density vector on the observed covariates. Defining the vector directly avoids choosing a representative router.

Define the conditional log-likelihood ratio and empirical Hellinger loss for this vector by
\begin{equation}
L_n(s,T)
=\sum_{a=1}^n\log\frac{g_{s,T,a}(Y_a)}{p_\circ(Y_a\mid X_a)},
\label{eq:pattern-template-likelihood}
\end{equation}
\begin{equation}
H_{n,s,T}^2
=\frac1n\sum_{a=1}^n
\left\{1-\int\sqrt{g_{s,T,a}(y)p_\circ(y\mid X_a)}dy\right\}.
\label{eq:pattern-template-hellinger}
\end{equation}

\medskip\noindent\textbf{Step 3: The fixed-candidate likelihood bound.}
For a fixed pair $(s,T)$, Lemma~\ref{lem:conditionalLR} gives, for every threshold $v$,
\begin{equation}
\Pp_*^{(K)}
\{L_n(s,T)\ge-a,\ nH_{n,s,T}^2>v\mid X_{1:n}\}
\le e^{a/2-v}.
\label{eq:pattern-template-likelihood-tail}
\end{equation}
Take
\begin{equation}
v=\log N_X+a/2+t.
\label{eq:conditional-union-threshold}
\end{equation}
Then
\begin{equation}
e^{a/2-v}=N_X^{-1}e^{-t}.
\label{eq:conditional-union-tail}
\end{equation}

\medskip\noindent\textbf{Step 4: A union bound over density vectors.}
There are at most $N_X$ pattern--template pairs, and hence at most $N_X$ distinct conditional-density vectors. Summing Eq.~\eqref{eq:conditional-union-tail} over these candidates gives
\[
\Pp_*^{(K)}
\Big\{\exists(s,T):L_n(s,T)\ge-a,\ nH_{n,s,T}^2>v\mid X_{1:n}\Big\}
\le e^{-t},
\]
which is Eq.~\eqref{eq:conditional-pattern-likelihood-tail}.

\medskip\noindent\textbf{Step 5: Substitution of the rounded estimator.}
Outside the exceptional event, the bound holds for every conditional-density vector with log-likelihood ratio at least $-a$. It therefore applies to the rounded estimator, including its dependence on the responses.
\end{proof}

% \begin{remark}[Likelihood control without bounded log-density ratios]
% No concentration inequality is applied directly to $\log(g/p_\circ)$, which is unbounded in $y$. The square-root transform uses the exact conditional expectation
% \[
% \E_*^{(K)}\!\left[
% \left.\sqrt{\frac{g(Y\mid X)}{p_\circ(Y\mid X)}}\,\right|X=x\right]
% =\int_{\R}\sqrt{g(y\mid x)p_\circ(y\mid x)}\,dy,
% \]
% so only Hellinger affinity appears. This is why the argument remains valid on the unbounded Gaussian response space.
% \end{remark}

\begin{lemma}[Relative empirical-to-population bound]\label{lem:relative}
Let $X_1,\ldots,X_n$ be iid with law $P$, write $P_nv:=n^{-1}\sum_{a=1}^n v(X_a)$, and let $\mathcal V$ be a class of measurable functions taking values in $[0,1]$. Suppose that for every $x_1,\ldots,x_{2n}$ the number of distinct vectors
\[
(v(x_1),\ldots,v(x_{2n})),\qquad v\in\mathcal V,
\]
is at most $S_{2n}$. Then, for every $t\ge0$, with probability at least $1-e^{-t}$,
\begin{equation}
Pv\le2P_nv+C\frac{\log(2S_{2n})+t}{n}
\quad\text{for all }v\in\mathcal V.
\label{eq:relative-empirical-bound}
\end{equation}
Here $C>0$ is a numerical constant.
\end{lemma}

In the likelihood application, $P=P_X$ and the iid requirement is part of the sampling model in Section~\ref{sec:intro-setup}.

\begin{proof}
\medskip\noindent\textbf{Step 1: Reduction to a ghost sample.}
Let $X_1',\ldots,X_n'$ be an independent copy of the covariate sample and put $P_n'$ for its empirical measure. Fix $u>0$ with $nu\ge32$. If for some $v$,
\begin{equation}
\mu:=Pv>2P_nv+u,
\label{eq:relative-deviation-event}
\end{equation}
then $\mu>u$ and $P_nv<\mu/2$. Since $0\le v\le1$,
\[
\operatorname{Var}(P_n'v)\le\mu/n.
\]
Chebyshev's inequality yields
\begin{equation}
\Pp\{P_n'v<3\mu/4\}
\le\frac{16}{n\mu}\le\frac12.
\label{eq:ghost-sample-lower-tail}
\end{equation}
For each original sample on which Eq.~\eqref{eq:relative-deviation-event} holds, fix a witnessing function, interpreting the event in outer probability if needed. The independent ghost sample satisfies the displayed lower-tail complement with probability at least $1/2$. The original event is therefore bounded by twice the probability that some $v$, with $\mu=Pv>u$, satisfies
\begin{equation}
P_nv\le\mu/2,
\qquad
P_n'v\ge3\mu/4.
\label{eq:paired-sample-deviation}
\end{equation}
Write
\[
a=\sum_{i=1}^nv(X_i),
\qquad
b=\sum_{i=1}^nv(X_i').
\]
Then $a\le2b/3$, so
\begin{equation}
b-a\ge(a+b)/5,
\label{eq:paired-sample-relative-gap}
\end{equation}
and
\begin{equation}
a+b\ge b\ge3n\mu/4\ge3nu/4.
\label{eq:paired-sample-mass-bound}
\end{equation}

\medskip\noindent\textbf{Step 2: Independent swaps of paired observations.}
Independently swap the two observations in each pair $(X_i,X_i')$ with probability $1/2$. This preserves the joint sample distribution. Conditional on the pooled observations, the difference $b-a$ for a fixed value pattern has the Rademacher representation
\[
\sum_{i=1}^n\epsilon_i\{v(X_i')-v(X_i)\}.
\]
Let $S=a+b$. Since $0\le v\le1$,
\begin{equation}
\sum_i\{v(X_i')-v(X_i)\}^2
\le\sum_i\{v(X_i')+v(X_i)\}=S.
\label{eq:swap-variance-bound}
\end{equation}
The Rademacher exponential bound therefore gives
\begin{equation}
\Pp_\epsilon\{b-a\ge S/5\}
\le e^{-S/50}.
\label{eq:swap-rademacher-tail}
\end{equation}

\medskip\noindent\textbf{Step 3: A union bound over value patterns.}
On Eq.~\eqref{eq:paired-sample-mass-bound}, the tail is at most $e^{-3nu/200}$. A union bound over the at most $S_{2n}$ value patterns, followed by the ghost-sample comparison, gives
\begin{equation}
\Pp\{\exists v:Pv>2P_nv+u\}
\le2S_{2n}e^{-3nu/200}.
\label{eq:relative-pattern-tail}
\end{equation}

\medskip\noindent\textbf{Step 4: Choice of the deviation level.}
Choose
\[
u=C\{\log(2S_{2n})+t\}/n
\]
with $C$ sufficiently large. This also guarantees $nu\ge32$. Equation \eqref{eq:relative-empirical-bound} follows.
\end{proof}

\begin{proof}[Proof of Theorem~\ref{thm:topmle}]
Set $\delta_n:=n^{-2}$ and approximate $\widehat p_n^{(K)}$ by a smooth template $\widehat g_n$ with the same routing map. We control its likelihood loss, derive empirical and population Hellinger bounds, and finally return to the original estimator.

\medskip\noindent\textbf{Step 1: Likelihood control for the rounded estimator.}
By global optimality and correct specification,
\begin{equation}
L_n(\widehat p_n^{(K)})\ge0.
\label{eq:sparse-mle-likelihood-bound}
\end{equation}
Lemma~\ref{lem:topsmooth} gives
\begin{equation}
L_n(\widehat g_n)
\ge
L_n(\widehat p_n^{(K)})
-C\delta_n\sum_{a=1}^n(1+|Y_a|).
\label{eq:rounded-likelihood-approximation}
\end{equation}
Under the truth, write $Y_a=V_a+\sqrt{\sigma_0}Z_a$, where $|V_a|\le M$ and $Z_a$ are independent standard Gaussians, independent of the covariates and latent component choices. Choose a fixed $s>0$ and $L_Z:=\log\E e^{s|Z|}<\infty$. Exponential Markov inequality gives, for every $t\ge1$,
\[
\Pp_*^{(K)}\left\{\sum_{a=1}^n|Z_a|>(nL_Z+t)/s\right\}\le e^{-t}.
\]
Consequently, with probability at least $1-e^{-t}$,
\begin{equation}\label{eq:mle-response-moment}
\sum_{a=1}^n(1+|Y_a|)\le C(n+t).
\end{equation}
On this event, Eq.~\eqref{eq:rounded-likelihood-approximation} implies
\begin{equation}\label{eq:mle-rounded-likelihood}
L_n(\widehat g_n)\ge-C\delta_n(n+t)=:-a_n(t).
\end{equation}
The tolerance $a_n(t)$ is deterministic. We can therefore apply the conditional likelihood inequality with this tolerance and then intersect its event with the response-moment event.

\medskip\noindent\textbf{Step 2: Empirical Hellinger control.}
Apply Lemma~\ref{lem:finite-trace-lr} with $a=a_n(t)$. Since $\delta_n=n^{-2}$, $a_n(t)\le C(1+t)$. On the response-moment event and outside the conditional likelihood exceptional event, the rounded estimator is among the candidates to which this uniform bound applies. The simultaneous likelihood event has conditional failure probability at most $e^{-t}$. On its intersection with the response-moment event, the smooth-template entropy bound \eqref{eq:sparse-template-cardinality} and Lemma~\ref{lem:routingtrace} give
\begin{equation}
\Hdist_n^2(\widehat g_n,p_\circ)
\le
C\frac{\mathfrak E_{n,K}+t}{n}.
\label{eq:rounded-empirical-hellinger-bound}
\end{equation}

\medskip\noindent\textbf{Step 3: Passage to the population loss.}
Consider
\[
\mathcal V_n
=
\{x\mapsto h_x^2(g_{R,T},p_\circ):R,\ T\in\mathcal T_{\delta_n}\}.
\]
Each member takes values in $[0,1]$. On any $2n$ covariates its value vector is determined by one smooth template and one active-set pattern, so
\begin{equation}
S_{2n}\le|\mathcal T_{\delta_n}|\Gamma_K(2n).
\label{eq:ghost-pattern-cardinality}
\end{equation}
Lemma~\ref{lem:relative} gives, simultaneously over the enlarged grid class,
\begin{equation}
\Hdist^2(g,p_\circ)
\le
2\Hdist_n^2(g,p_\circ)
+C\frac{\mathfrak E_{n,K}+t}{n}
\label{eq:rounded-population-comparison}
\end{equation}
with failure probability at most $e^{-t}$. In particular, it applies to the data-dependent $\widehat g_n$.

\medskip\noindent\textbf{Step 4: Return to the MLE.}
Combining Eqs.~\eqref{eq:rounded-empirical-hellinger-bound}--\eqref{eq:rounded-population-comparison},
\begin{equation}
\Hdist^2(\widehat g_n,p_\circ)
\le
C\frac{\mathfrak E_{n,K}+t}{n}.
\label{eq:rounded-population-hellinger-bound}
\end{equation}
Lemma~\ref{lem:topsmooth} also gives
\[
\Hdist^2(\widehat p_n^{(K)},\widehat g_n)\le Cn^{-2}.
\]
The Hellinger triangle inequality, squared, gives
\[
\Hdist^2(\widehat p_n^{(K)},p_\circ)
\le2\Hdist^2(\widehat p_n^{(K)},\widehat g_n)
+2\Hdist^2(\widehat g_n,p_\circ)
\le C\frac{\mathfrak E_{n,K}+t}{n}.
\]
The three exceptional events are the response-moment event, the conditional likelihood event (integrated over the covariates), and the empirical-to-population event. Each has probability at most $e^{-t}$. Their union therefore has probability at most $3e^{-t}$ for every $t\ge1$, proving Eq.~\eqref{eq:top-density-rate}.
\end{proof}

\subsection{Proof of Corollary~\ref{cor:topmle}}\label{app:topmletransfer}
\begin{proof}[Proof of Corollary~\ref{cor:topmle}]
Theorem~\ref{thm:topmle} and Lemma~\ref{lem:hell-l2} imply that, except on an event of probability at most $3e^{-t}$,
\[
\|p_{\widehat G_n^{(K)}}^{(K)}-p_{G_*}^{(K)}\|_{L^2(P_X\otimes dy)}
\le C\sqrt{\frac{\mathfrak E_{n,K}+t}{n}}.
\]
For $K\ge2$, the inverse assumptions hold uniformly for the fitted class and the chosen deterministic threshold. Substitution into Eq.~\eqref{eq:topk-inverse-robust} gives Eq.~\eqref{eq:topmle-recovery-tail}. The rare-mass bound \eqref{eq:topk-rare-mass} holds for every fitted model, even when that model is data-dependent, so
\[
4KM\sqrt{w_{\rm rare}(\lambda;\widehat G_n^{(K)},G_*)}
\le4KM\binom{k}{K}\sqrt\lambda.
\]
This proves Eq.~\eqref{eq:topmle-recovery-rate}. Likewise Eq.~\eqref{eq:topk-inverse-retained} gives the retained-loss rate in Table~\ref{tab:intro-rates}. Both conclusions follow by deterministic comparison on the same density-estimation event. For one deterministic threshold $\lambda_n$ at each sample size, no additional union bound over regions or thresholds is required.

For the expert conclusions, fix a model pair and use the loss-minimizing bijection $\sigma_C$ on each region. The active-weight floor and $\|u\|_{L^1(P_C)}\le\|u\|_{L^2(P_C)}$ imply
\[
\sum_{j\in S}\|h_{\sigma_C(j)}-h_j^*\|_{L^1(C,P_X)}
\le\underline\pi_K^{-1}w_CD_C.
\]
Summing over retained regions and applying Cauchy--Schwarz gives
\[
\mathcal E_{K,\lambda}(G,G_*)
\le\underline\pi_K^{-1}\sum_{C\in\mathcal R_\lambda}w_CD_C
\le\underline\pi_K^{-1}D_{K,\lambda}(G,G_*),
\]
since $\sum_{C\in\mathcal R_\lambda}w_C\le1$. The same argument over the full partition gives $\mathcal E_K\le\underline\pi_K^{-1}\Dtop$. Applying the retained inverse bound on the density-estimation event proves Eq.~\eqref{eq:topk-retained-expert-rate}; combining the full-partition comparison with Eq.~\eqref{eq:topmle-recovery-rate} proves Eq.~\eqref{eq:topk-full-expert-rate}.

\end{proof}

\subsection{Proof of Proposition~\ref{prop:sieve-rate}}\label{app:mlethreshold}
\begin{proof}[Proof of Proposition~\ref{prop:sieve-rate}]
Corollary~\ref{cor:topmle}, the assumed polynomial lower bound for $C_{K,\lambda}$, and the finite count of joint routing regions give
\[
\Dtop(\widehat G_n^{(K)},G_*)
=\mathcal{O}_P\bigl(q_{n,K}\lambda^{-\alpha}+\lambda^{1/2}\bigr).
\]
Since $q_{n,K}\to0$, the choice \eqref{eq:lambda-opt} belongs to $(0,1]$ and lies in the range of the assumed inverse estimates for all sufficiently large $n$. Substitution gives
\[
q_{n,K}\lambda_n^{-\alpha}\asymp q_{n,K}^{1/(2\alpha+1)},
\qquad
\lambda_n^{1/2}\asymp q_{n,K}^{1/(2\alpha+1)},
\]
which proves Eq.~\eqref{eq:sieve-rate}. The deterministic comparison $\mathcal E_K\le\underline\pi_K^{-1}\Dtop$ from Corollary~\ref{cor:topmle} then gives Eq.~\eqref{eq:sieve-expert-rate} under the same regional matchings.
\end{proof}

The polynomial exponent of the inverse constant follows from the conditional geometry and restriction bounds. Suppose $a_\lambda\asymp\lambda$ and $c_{\rm res,\lambda}\ge c>0$, with the other model constants uniform. The matching estimate \eqref{eq:assign-constant-bound} then gives
\[
C_A(a_\lambda,A_\lambda)\lesssim
1+\frac{A_\lambda+2}{a_\lambda}.
\]
If $A_\lambda\lesssim\lambda^{-1}$, then $C_A\lesssim\lambda^{-2}$. The explicit constant \eqref{eq:tauconstant-expanded} therefore satisfies
\[
C_{K,\lambda}\gtrsim\lambda^{2K-2}\lambda^2=\lambda^{2K}.
\]
If $A_\lambda\lesssim1$, then $C_A\lesssim\lambda^{-1}$ instead, giving
\[
C_{K,\lambda}\gtrsim\lambda^{2K-1}.
\]
Applying Proposition~\ref{prop:sieve-rate} with $\alpha=2K$ or $2K-1$ yields the stated exponents. These calculations require the uniform restriction bound; its deterioration would contribute further powers of $\lambda$.

Consider affine experts $h(x,\eta)=a^\top x+c$ on the fixed cube, satisfying global single-cluster geometry and Assumption~\ref{ass:denseac}. The $L^1(P_X)$ and supremum norms are equivalent on the finite-dimensional space of affine functions. Indeed, on the Euclidean unit sphere of affine coefficients, the first norm has a positive minimum and the second a finite maximum. Thus a constant $C_{\rm aff}$ depending only on the cube and dimension satisfies
\[
\|h_j^*-h_\ell^*\|_\infty
\le C_{\rm aff}\|h_j^*-h_\ell^*\|_{L^1(P_X)}
\le C_{\rm aff}C_0\DeltaSep.
\]
This upper bound survives conditioning on any routing region. For a retained region $C$ of probability at least $\lambda$, anti-concentration gives
\[
P_C\{|h_j^*-h_\ell^*|<a\DeltaSep\}
\le\frac{2C_{\rm ac}a}{\lambda}.
\]
With $N_K=\binom K2$, choose
\[
a_\lambda:=c_a\lambda,\qquad
c_a:=\min\{1,C_{\rm aff}C_0,(4N_KC_{\rm ac})^{-1}\},
\qquad A_\lambda:=C_{\rm aff}C_0.
\]
A union bound over active pairs leaves conditional probability at least $1/2$ on which all lower bounds hold, and the upper bound holds everywhere. Hence Assumption~\ref{ass:topR1} holds with these constants and $\rho_\lambda=1/2$. If Assumption~\ref{ass:topR2} additionally has restriction constants bounded below uniformly in $\lambda$, the preceding calculation gives Eq.~\eqref{eq:linear-sieve-rate}. No such restriction bound is inferred merely from affinity of the experts.

% \subsection{Proof of Eq.~\eqref{eq:top1-mle-rate}}\label{app:top1mleproof}
% \begin{proof}
% Theorem~\ref{thm:topmle} and Lemma~\ref{lem:hell-l2} give a density $L^2$ error of order $\varepsilon_{n,1}$. The lower bound in Eq.~\eqref{eq:top1-equivalence} then yields
% \[
% D_{\rm sel}(\widehat G_n^{(1)},G_*)
% \le c_-^{-1}\|p_{\widehat G_n^{(1)}}^{(1)}-p_{G_*}^{(1)}\|_{L^2(P_X\otimes dy)}
% =\mathcal{O}_P(\varepsilon_{n,1}),
% \]
% which proves Eq.~\eqref{eq:top1-mle-rate} without separation or a region threshold.
% \end{proof}

\section{Proofs for Results in Section~\ref{sec:linear}}\label{app:linear}
This appendix verifies the claims for homogeneous linear experts in Section~\ref{sec:linear}. Projection moments compare coefficient and functional separation, while conditioning on one coordinate gives the density bound. Remez inequalities control expert errors on convex routing regions; binary gate errors reduce to affine discrepancies. The final calculation converts coefficient clustering into retained-region separation with explicit threshold dependence.

\subsection{Proof of Proposition~\ref{prop:linearsep}}\label{app:linearsepproof}
\begin{proof}[Proof of Proposition~\ref{prop:linearsep}]
Let $Z:=a^\top X$ and write $\|Z\|_p:=(\E|Z|^p)^{1/p}$ for its moment norm under $P_X$. The assertion is immediate for $a=0$. For $a\ne0$, independence and symmetry of the cube coordinates yield
\[
\begin{aligned}
\E Z^2&=\frac{b^2}{3}\|a\|_2^2,\\
\E Z^4&=\frac{b^4}{5}\sum_r a_r^4
+\frac{2b^4}{3}\sum_{r<s}a_r^2a_s^2
\le\frac{b^4}{3}\|a\|_2^4.
\end{aligned}
\]
Cauchy--Schwarz yields the upper bound in Eq.~\eqref{eq:linear-norm-equivalence}. For the lower bound, interpolation between the $L^1$ and $L^4$ norms gives
\[
\|Z\|_2\le\|Z\|_1^{1/3}\|Z\|_4^{2/3},
\qquad
\|Z\|_1\ge\frac{\|Z\|_2^3}{\|Z\|_4^2}
\ge\frac{(b\|a\|_2/\sqrt3)^3}{b^2\|a\|_2^2/\sqrt3}
=\frac b3\|a\|_2.
\]
Apply this two-sided comparison to $a=\eta_j^*-\eta_\ell^*$ and minimize over pairs to obtain Eq.~\eqref{eq:linear-separation-equivalence}.
\end{proof}

\subsection{Proof of Proposition~\ref{prop:linearac}}\label{app:linearacproof}
\begin{proof}[Proof of Proposition~\ref{prop:linearac}]
Choose $r$ with $|a_r|=\|a\|_\infty>0$. Conditional on all other coordinates,
\[
Z=a_rX_r+\sum_{s\ne r}a_sX_s
\]
is uniform on an interval of length $2b|a_r|$. Averaging these conditional densities gives an unconditional density with
\[
\|f_Z\|_\infty\le\frac{1}{2b\|a\|_\infty}
\le\frac{\sqrt{d_x}}{2b\|a\|_2}.
\]
Proposition~\ref{prop:linearsep} gives $b\|a\|_2\ge\sqrt3\,\E|Z|$, proving Eq.~\eqref{eq:linear-ac}. Taking $a=\eta_j^*-\eta_\ell^*$ proves Eq.~\eqref{eq:linear-ac-pair} at the pair-specific scale $\Delta_{j\ell}$.
\end{proof}

\subsection{Proof of Proposition~\ref{prop:linearrestriction}}\label{app:linearrestrictionproof}
We derive Eq.~\eqref{eq:aff-remez} from the supremum form of the degree-one Remez inequality. On a convex body $C$, that inequality gives
\[
\|L\|_{L^\infty(C)}\le R(d_x,\theta)\|L\|_{L^\infty(E)}
\quad\text{whenever }P_C(E)\ge\theta>0,
\]
where the supremum on a measurable subset is interpreted essentially and $R$ is independent of $C$. Fix $A$ with $P_C(A)\ge\rho$ and put $s=\|L\|_{L^p(A,P_C)}$. If $s>0$, Markov's inequality shows that
\[
E:=A\cap\{|L|\le(2/\rho)^{1/p}s\}
\quad\text{satisfies}\quad P_C(E)\ge\rho/2.
\]
Hence
\[
\|L\|_{L^p(P_C)}\le\|L\|_{L^\infty(C)}
\le R(d_x,\rho/2)(2/\rho)^{1/p}s.
\]
This proves Eq.~\eqref{eq:aff-remez} with
$c_{\rm aff}(d_x,p,\rho)=(\rho/2)^{1/p}/R(d_x,\rho/2)$.
If $s=0$, Remez applied directly to $A$ makes $L$ zero throughout $C$, so the same inequality holds.

\begin{proof}[Proof of Proposition~\ref{prop:linearrestriction}]
The fixed tie rule expresses each active-set event as an intersection of strict or non-strict linear halfspaces. Identically equal scores impose either a redundant or an impossible constraint. All other ties lie on proper hyperplanes and have Lebesgue measure zero. Hence each positive-probability joint region agrees almost everywhere with a full-dimensional convex polytope, on which $P_C$ is normalized Lebesgue measure.

For active labels $i,j$, let $L_{ij}(x)=(\eta_i-\eta_j^*)^\top x$. The bounds $\underline\pi_K\le\pi_{j,G_*}^{(K)}\le1$ and Eq.~\eqref{eq:aff-remez} give
\[
\begin{aligned}
\|\pi_{j,G_*}^{(K)}L_{ij}\|_{L^2(A,P_C)}
&\ge\underline\pi_K\|L_{ij}\|_{L^2(A,P_C)}\\
&\ge\underline\pi_Kc_{\rm aff}(d_x,2,\rho)\|L_{ij}\|_{L^2(P_C)}\\
&\ge\underline\pi_Kc_{\rm aff}(d_x,2,\rho)
\|\pi_{j,G_*}^{(K)}L_{ij}\|_{L^2(P_C)}.
\end{aligned}
\]
This includes the zero-error case.

For $K=2$, write $T=\{i,i'\}$ and $S=\{j,j'\}$ and define
\[
g(x)=\ell_i(x)-\ell_{i'}(x),\qquad
g_*(x)=\ell_j^*(x)-\ell_{j'}^*(x),\qquad
s(z)=\frac{1}{1+e^{-z}}.
\]
The two weights being compared are $s(g)$ and $s(g_*)$, with $g,g_*\in[-2B,2B]$. Since
\[
m_B:=\min_{|z|\le2B}s'(z)>0,\qquad s'(z)\le\frac14,
\]
the mean-value theorem gives
\[
m_B|g-g_*|\le|s(g)-s(g_*)|\le\frac14|g-g_*|.
\]
The function $g-g_*$ is affine, so Eq.~\eqref{eq:aff-remez} with $p=1$ yields
\[
\begin{aligned}
\|s(g)-s(g_*)\|_{L^1(A,P_C)}
&\ge m_Bc_{\rm aff}(d_x,1,\rho)\|g-g_*\|_{L^1(P_C)}\\
&\ge4m_Bc_{\rm aff}(d_x,1,\rho)
\|s(g)-s(g_*)\|_{L^1(P_C)}.
\end{aligned}
\]
Thus $c_\pi=4m_Bc_{\rm aff}(d_x,1,\rho)$ is valid. For the dense conclusions, apply the same arguments on $C=\cX$, using the dense weight floor and, for the binary case, the two full logits.
\end{proof}

For general gates, write $D=Z_TZ_S^*$ in Eq.~\eqref{eq:linear-gate-numerator}. The denominator bounds and Eq.~\eqref{eq:exp-remez} give
\[
\left\|\frac F D\right\|_{L^1(A,P_C)}
\ge\frac{c_{\exp}(\rho)}{K^2e^{2B}}\|F\|_{L^1(P_C)}
\ge e^{-4B}c_{\exp}(\rho)
\left\|\frac F D\right\|_{L^1(P_C)}.
\]
For dense cluster masses, put $Z=\sum_i e^{\ell_i}$ and $Z_*=\sum_j e^{\ell_j^*}$. The numerator for $\Pi_m-\Pi_m^*$ is
\[
\left(\sum_{i\in\widehat{\mathcal C}_m}e^{\ell_i}\right)Z_*
-\left(\sum_{j\in\mathcal C_m^*}e^{\ell_j^*}\right)Z.
\]
It has at most $2k^2$ exponential terms, and the denominator $ZZ_*$ has the same bounds with $K=k$. Restriction for these aggregate numerators therefore suffices for Assumption~\ref{ass:clusterR}.

Lemma~\ref{lem:affine-gate-restriction} verifies these numerator restriction inequalities in every covariate dimension. Its proof applies the multivariate Tur\'an--Nazarov estimate to a subset of $A$ on which the numerator is bounded by a multiple of its restricted $L^1$ norm, then uses the denominator bounds above. The number of exponential terms and the uniform logit bound control the constants, independently of the region's volume and shape and of collisions of exponents. The classical one-dimensional estimate \citep{Nazarov1993} is a special case of this argument.

\subsection{Proof of Proposition~\ref{prop:heavylinear}}\label{app:heavylinearproof}
We first express the coefficient-cluster assumptions in the functional scales used by the recovery theorems. For a within-block pair satisfying Eq.~\eqref{eq:linear-coefficient-blocks}, Proposition~\ref{prop:linearsep} gives
\[
\Delta_{j\ell}\le\frac b{\sqrt3}C_\eta\delta_\eta
\le\sqrt3C_\eta\DeltaSep.
\]
A coefficient gap $D_\eta$ between blocks gives $\Delta_{j\ell}\ge bD_\eta/3=\DeltaMacro$. Under Eq.~\eqref{eq:linear-cluster-scale},
\[
\DeltaSep\le\frac b{\sqrt3}\delta_\eta
\le\frac{bD_\eta}{24\sqrt3C_\eta}
=\frac{\DeltaMacro}{8C_0},\qquad C_0=\sqrt3C_\eta.
\]
For the global single-cluster condition \eqref{eq:eta-cluster}, the same calculation proves the first bound in Eq.~\eqref{eq:function-cluster}. The second follows from
\[
\sup_{x\in\cX}|(\eta_j^*-\eta_\ell^*)^\top x|
\le b\sqrt{d_x}C_\eta\delta_\eta
\le3\sqrt{d_x}C_\eta\DeltaSep.
\]

\begin{proof}[Proof of Proposition~\ref{prop:heavylinear}]
Let $\DeltaSep>0$ and fix $C=\mathcal R_{S,T}$ with $w_C\ge\tau$. For each distinct $j,\ell\in S$, Proposition~\ref{prop:linearac} yields
\[
\begin{aligned}
P_C\{|h_j^*-h_\ell^*|<a\DeltaSep\}
&\le\frac{2a\DeltaSep}{w_C}\frac{\sqrt{d_x}}{2\sqrt3\,\Delta_{j\ell}}\\
&\le\frac a\tau\sqrt{\frac{d_x}{3}}.
\end{aligned}
\]
Taking $a=a_\tau$ from Eq.~\eqref{eq:heavy-constants} and summing over the $\binom K2$ active pairs bounds the probability of any failed lower bound by $1/4$. On the complementary set $E_C$, all active true differences are at least $a_\tau\DeltaSep$, and $P_C(E_C)\ge3/4$. The pointwise upper bound in Eq.~\eqref{eq:function-cluster} holds everywhere, so Assumption~\ref{ass:topR1} follows with the stated constants.

As $\tau\downarrow0$, $a_\tau\asymp\tau$ and $A_\tau$ is fixed. Equation~\eqref{eq:assign-constant-bound} gives $C_A(a_\tau,A_\tau)\lesssim\tau^{-1}$. If $c_{\rm res,\tau}=\min\{c_{h,\tau},c_{\pi,\tau}\}$ is bounded below, then Eq.~\eqref{eq:tauconstant-expanded} gives
\[
C_{K,\tau}
=\frac{c_{\rm sc}a_\tau^{2K-2}c_{\rm res,\tau}}{2K\,C_A(a_\tau,A_\tau)}
\gtrsim\tau^{2K-1}.
\]
All constants are uniform over the retained regions and fitted models under the stated bounds. For each fixed $K$, Proposition~\ref{prop:linearrestriction} supplies expert restriction and Lemma~\ref{lem:affine-gate-restriction} supplies gate restriction at the fixed relative level $\rho=3/(4K!)$. This verifies the uniform restriction constants used in Section~\ref{sec:lineargeom}; for $K=2$, the elementary binary proof already suffices.
\end{proof}

\section{Partial Transport Interpretation of the Dense Recovery Losses}\label{app:partialot}

This appendix explains the displacement and mass-discrepancy terms in the dense recovery losses through partial transport, which has also been used to characterize heterogeneous parameter estimation rates in finite mixture models \citep{Le2026HeterogeneousPartialTransport}. We first distinguish prescribed transported mass from a penalty for unmatched mass. We then constrain transport to a fixed expert matching and compare maximal common-mass displacement with true-weighted displacement. The dense weight floor makes their functional versions equivalent. The transport penalty $\lambda$ used here is distinct from the routing-probability threshold in Section~\ref{sec:topmle}.

\subsection{Prescribed-mass and penalized partial transport}\label{app:partial-definition}
Partial transport allows only part of each measure to be matched. We begin with the standard prescribed-mass formulation \citep{Figalli2010PartialTransport,Le2022MultimarginalPartialTransport} under absolute-distance cost. Introducing an unmatched-mass penalty afterward makes explicit which transport problem corresponds to each term of the recovery loss.

\begin{definition}[Partial optimal transport with prescribed mass]\label{def:partial-transport}
Let $\mu,\nu$ be finite nonnegative Borel measures on a compact interval $I\subset\R$. Write $|\mu|:=\mu(I)$ and $|\nu|:=\nu(I)$. For a finite nonnegative measure $\gamma$ on $I\times I$, let $\gamma_1,\gamma_2$ denote its first and second marginals, and write $|\gamma|:=\gamma(I\times I)$. The set of partial transport plans is
\begin{equation}\label{eq:partial-plans}
\Gamma_{\le}(\mu,\nu)
:=\{\gamma\ge0:\gamma_1\le\mu,\ \gamma_2\le\nu\},
\end{equation}
where domination means inequality on every Borel set. For a prescribed transported mass $\alpha\in[0,\min\{|\mu|,|\nu|\}]$, define
\begin{equation}\label{eq:partial-fixed-mass}
\mathsf{POT}_{\alpha}(\mu,\nu)
:=\min_{\substack{\gamma\in\Gamma_{\le}(\mu,\nu)\\|\gamma|=\alpha}}
\int_{I\times I}|u-v|\,d\gamma(u,v).
\end{equation}
\end{definition}
The marginals $\gamma_1$ and $\gamma_2$ specify the transported portions; their residuals remain unmatched. The feasible set is nonempty and weakly compact, and continuity of the cost ensures attainment. For probability measures, transporting all mass gives $\mathsf{POT}_1(\mu,\nu)=W_1(\mu,\nu)$, whereas $\mathsf{POT}_0(\mu,\nu)=0$. Prescribing $\alpha$ fixes the amount that must be transported.

\begin{definition}[Penalized partial optimal transport]\label{def:penalized-partial-transport}
For $\lambda>0$, define
\begin{equation}\label{eq:partial-penalized-measures}
\mathsf{PT}_{\lambda}(\mu,\nu)
:=\min_{\gamma\in\Gamma_{\le}(\mu,\nu)}
\left\{\int|u-v|\,d\gamma
+\lambda\bigl(|\mu|+|\nu|-2|\gamma|\bigr)\right\}.
\end{equation}
\end{definition}
This problem chooses the transported amount and charges $\lambda$ per unit of unmatched mass on each side. Grouping plans by their total mass gives the exact relation
\begin{equation}\label{eq:partial-mass-penalty-relation}
\mathsf{PT}_{\lambda}(\mu,\nu)
=\min_{0\le\alpha\le\min\{|\mu|,|\nu|\}}
\left\{\mathsf{POT}_{\alpha}(\mu,\nu)
+\lambda(|\mu|+|\nu|-2\alpha)\right\}.
\end{equation}
The penalized problem therefore optimizes over the transported amount as well as the plan. For probability measures, leaving mass $1-\alpha$ unmatched on each side costs $2\lambda(1-\alpha)$. This makes transport over distance greater than $2\lambda$ more expensive than leaving the corresponding mass unmatched.

\subsection{Transport constrained by a fixed expert matching}\label{app:partial-matching}
We now apply these definitions to $\mu=M_{G_*}(x)$ and $\nu=M_G(x)$ at a fixed covariate $x$, with $I=[-M,M]$. Abbreviate
\[
a_i=\pi_i(x),\qquad b_j=\pi_j^*(x),\qquad
s_i=h_i(x),\qquad t_j=h_j^*(x).
\]
In labelled matrix form, Eq.~\eqref{eq:partial-penalized-measures} becomes
\begin{equation}\label{eq:partial-unrestricted}
\begin{aligned}
\mathsf{PT}_\lambda(x)
&:=\mathsf{PT}_\lambda(M_{G_*}(x),M_G(x))\\
&=\min_{\substack{\gamma_{ij}\ge0\\
\sum_i\gamma_{ij}\le b_j,\ \sum_j\gamma_{ij}\le a_i}}
\left\{\sum_{i,j}\gamma_{ij}|s_i-t_j|
+\lambda\left(2-2\sum_{i,j}\gamma_{ij}\right)\right\}.
\end{aligned}
\end{equation}
The entry $\gamma_{ij}$ transports mass from true label $j$ to fitted label $i$. Retaining labels leaves the unrestricted optimum unchanged: mass at a shared numerical location can be distributed among its labels without changing the cost.

Unrestricted conditional transport permits mass splitting and correspondences that vary with the covariate, as in the target $\DOT$. To compare individual experts, we instead fix an admissible permutation $\sigma$ and restrict transport to the pairs $j\mapsto\sigma(j)$. Section~\ref{sec:denseclusters} uses the same permutation throughout the covariate domain; here we examine its cost at a fixed $x$. Define the distance and available common mass of each pair by
\[
r_j:=|s_{\sigma(j)}-t_j|,\qquad
m_j:=\min\{a_{\sigma(j)},b_j\},\qquad
\alpha_\sigma:=\sum_jm_j.
\]
The corresponding prescribed-mass partial optimal transport problem is
\begin{equation}\label{eq:partial-matched-fixed-mass}
\mathsf{POT}_{\alpha,\sigma}(x)
:=\min_{\substack{0\le u_j\le m_j\\\sum_j u_j=\alpha}}
\sum_j u_jr_j,
\qquad 0\le\alpha\le\alpha_\sigma.
\end{equation}
This is prescribed-mass partial transport subject to the fixed matching. Its largest feasible amount is $\alpha_\sigma$. At that amount, all pairwise capacities are attained, so $u_j=m_j$ for every $j$ and
\begin{equation}\label{eq:partial-maximal-mass}
\mathsf{POT}_{\alpha_\sigma,\sigma}(x)=\sum_jm_jr_j,
\qquad
2(1-\alpha_\sigma)=\sum_j|a_{\sigma(j)}-b_j|.
\end{equation}
The second identity follows from $2\min\{a,b\}=a+b-|a-b|$ and the fact that both weight vectors sum to one.

Adding the unmatched-mass penalty to this prescribed-mass optimum gives
\begin{equation}\label{eq:partial-common-cost}
\begin{aligned}
\cP_{\lambda,\sigma}(x)
&:=\mathsf{POT}_{\alpha_\sigma,\sigma}(x)
+2\lambda(1-\alpha_\sigma)\\
&=\sum_jm_jr_j+\lambda\sum_j|a_{\sigma(j)}-b_j|.
\end{aligned}
\end{equation}
The cost transports the largest common mass permitted by the matching and penalizes the residual mass on each side.

For comparison, allowing the amount to vary under the same matching gives the penalized optimum
\begin{equation}\label{eq:partial-restricted}
\begin{aligned}
\mathsf{PT}_{\lambda,\sigma}(x)
&=\min_{0\le\alpha\le\alpha_\sigma}
\{\mathsf{POT}_{\alpha,\sigma}(x)+2\lambda(1-\alpha)\}\\
&=\min_{0\le u_j\le m_j}\sum_j
\left[u_jr_j+\lambda\{a_{\sigma(j)}+b_j-2u_j\}\right].
\end{aligned}
\end{equation}
Minimizing each linear summand yields
\begin{equation}\label{eq:partial-truncated}
\mathsf{PT}_{\lambda,\sigma}(x)
=\sum_jm_j\min\{r_j,2\lambda\}
+\lambda\sum_j|a_{\sigma(j)}-b_j|.
\end{equation}
Indeed, $u_j=m_j$ is optimal when $r_j<2\lambda$, $u_j=0$ is optimal when $r_j>2\lambda$, and any amount is optimal at equality. Consequently $\mathsf{PT}_\lambda\le\mathsf{PT}_{\lambda,\sigma}\le\cP_{\lambda,\sigma}$, with the second inequality an equality whenever every matched distance is at most $2\lambda$. The prescribed-mass cost \eqref{eq:partial-common-cost} retains full expert displacements even when discarding additional mass would be cheaper.

The pointwise counterpart of the paper's loss instead weights displacement by the true mass:
\begin{equation}\label{eq:point-partial-cost}
\cC_{\lambda,\sigma}(x)
:=\sum_j b_jr_j+\lambda\sum_j|a_{\sigma(j)}-b_j|.
\end{equation}
This cost first moves all true mass to the matched fitted locations, producing $\sum_jb_j\delta_{s_{\sigma(j)}}$, and then measures the remaining discrepancy between labelled weights. Half of that weight discrepancy is the total variation distance between the matched probability vectors. Since the comparison retains labels, coincident numerical locations do not cancel componentwise mass errors. Weighting displacement by the true gate reflects the expert's contribution to the conditional law; the positive weight floor also ensures control of unweighted displacement.

Both terms are needed because expert locations and gate probabilities can change separately. Taking $\lambda=\DeltaSep$ measures component-mass error at the separation scale: moving mass $u$ between two fixed atoms at distance $d$ costs $ud$ in Wasserstein distance and changes their weight vector by $2u$ in $\ell_1$. When $d\asymp\DeltaSep$, the transport cost and the separation-weighted mass discrepancy are comparable. Lemma~\ref{lem:assign} establishes the corresponding bound for the combined matching cost under one-cluster geometry.

\begin{proposition}[Comparison with common-mass transport]\label{prop:partialot}
Under the dense weight floor Eq.~\eqref{eq:weight-floor}, for every $x$, $\sigma$ and $\lambda>0$,
\begin{equation}\label{eq:partial-local-equivalence}
\mathsf{PT}_{\lambda,\sigma}(x)
\le\cP_{\lambda,\sigma}(x)
\le\cC_{\lambda,\sigma}(x)
\le\underline\pi_k^{-1}\cP_{\lambda,\sigma}(x).
\end{equation}
If every matched distance is at most $2\lambda$, the first inequality is an equality.
\end{proposition}

\begin{proof}
The first assertion follows from Eq.~\eqref{eq:partial-truncated}. For the remaining inequalities, the weight floor and $b_j\le1$ imply
$\underline\pi_k b_j\le m_j\le b_j$. Multiply by $r_j\ge0$, sum, and add the identical nonnegative mass-penalty term.
\end{proof}

The dense pointwise cost is comparable to $\cP_{\lambda,\sigma}$, which transports the maximal common mass under the matching. The penalized optimum $\mathsf{PT}_{\lambda,\sigma}$ equals $\cP_{\lambda,\sigma}$ when every matched distance is at most $2\lambda$, but can be smaller otherwise. In particular, with $\lambda=\DeltaSep$, discarding a substantially displaced expert can cost an amount tending to zero with $\DeltaSep$. The full displacement term in Eq.~\eqref{eq:point-partial-cost} retains information about its location.

\subsection{Equivalence of common-mass and true-weighted functional losses}
For an admissible permutation $\sigma\in\Sigma_{\mathcal C}$, the three contributions to the loss are
\begin{equation}\label{eq:partial-loss-components}
\begin{aligned}
E_\sigma&:=\sum_{j=1}^k
\|\pi_j^*(h_{\sigma(j)}-h_j^*)\|_{L^2(P_X)},
&&\text{expert displacement},\\
V_\sigma&:=\DeltaSep\sum_{j=1}^k
\|\pi_{\sigma(j)}-\pi_j^*\|_{L^1(P_X)},
&&\text{component mass discrepancy},\\
V_{\mathcal C}&:=\omega_q\sum_{m=1}^q
\|\Pi_m(\cdot;G)-\Pi_m^*\|_{L^1(P_X)},
&&\text{cluster mass discrepancy}.
\end{aligned}
\end{equation}
Thus $\Dmc=\min_{\sigma\in\Sigma_{\mathcal C}}(E_\sigma+V_\sigma+V_{\mathcal C})$. For one cluster, $\omega_1=0$ and $\Sigma_{\mathcal C}=\Sigma_k$, giving $\Dfunc$. For several clusters, $\omega_q=\DeltaMacro$ and the permutation preserves the prescribed true and fitted blocks.

The aggregate term $V_{\mathcal C}$ distinguishes redistribution within a block from net transfer between blocks. The former leaves the block mass unchanged, whereas the latter crosses a gap of order $\DeltaMacro$ on the informative set. This term is independent of the within-block permutation and retains control of mass discrepancies as the within-block separation decreases. Lemma~\ref{lem:hierassign} bounds the component and cluster terms together by a single Wasserstein cost; they are not a decomposition of the cost of an optimal plan.

The expert terms use $L^2(P_X)$ to match the covariate norm in the density error, while the mass terms use $L^1(P_X)$ to measure average absolute gate errors. Summing the component norms retains separate control of each matched expert and gate. These norms also specify the discrepancies to which the restriction assumptions apply.

For each admissible $\sigma$, let
\[
m_{\sigma,j}(x):=\min\{\pi_{\sigma(j)}(x),\pi_j^*(x)\}.
\]
The functional version of the common-mass cost, with the same norms and cluster correction as Definition~\ref{def:Dmc}, is
\begin{equation}\label{eq:partial-functional-loss}
\begin{aligned}
D_{\rm partial,cluster}(G,G_*):=
\min_{\sigma\in\Sigma_{\mathcal C}}\Bigg\{&
\sum_j\Bigl[
\|m_{\sigma,j}(h_{\sigma(j)}-h_j^*)\|_{L^2(P_X)}
+\DeltaSep\|\pi_{\sigma(j)}-\pi_j^*\|_{L^1(P_X)}\Bigr]\\
&+\omega_q\sum_m\|\Pi_m(\cdot;G)-\Pi_m^*\|_{L^1(P_X)}\Bigg\}.
\end{aligned}
\end{equation}
Denote its one-cluster specialization by $D_{\rm partial}$.

\begin{proposition}[Equivalence of the functional losses]\label{prop:partial-functional}
Under Eq.~\eqref{eq:weight-floor},
\begin{equation}\label{eq:partial-functional-equivalence}
\begin{aligned}
D_{\rm partial,cluster}&\le\Dmc
\le\underline\pi_k^{-1}D_{\rm partial,cluster},\\
D_{\rm partial}&\le\Dfunc
\le\underline\pi_k^{-1}D_{\rm partial}.
\end{aligned}
\end{equation}
The constants are independent of $\DeltaSep$ when $k$ and $B$ are fixed, and no localization of the fitted experts is needed for these comparisons.
\end{proposition}

\begin{proof}
For every $x,j,\sigma$,
$\underline\pi_k\pi_j^*(x)\le m_{\sigma,j}(x)\le\pi_j^*(x)$.
Multiplication by the absolute expert difference and monotonicity of the $L^2$ norm compare the displacement terms. The component and cluster mass terms are identical and nonnegative. Sum for each fixed permutation and then minimize over the same set of permutations. Taking $q=1$ gives the second assertion.
\end{proof}

The equivalence compares two functional losses under the same admissible matching and cluster-mass correction. Taking component norms before minimizing preserves one correspondence between whole expert functions. Integrating unrestricted pointwise transport would allow the matching and mass allocation to vary with the covariate and would define a different recovery target.

This structure also determines the clustered inverse argument in Appendix~\ref{app:densemcfuncproof}. Scalar inversion and the transport-to-matching comparison first control pointwise errors on an informative set. Selecting one permutation on a subset of positive probability then permits Assumptions~\ref{ass:denseR} and \ref{ass:clusterR} to extend its component and cluster-mass errors to the full domain, yielding precisely the norms in Eq.~\eqref{eq:partial-loss-components}.

\bibliographystyle{plain}
\bibliography{references}

@article{Jacobs1991,
  author = {Jacobs, Robert A. and Jordan, Michael I. and Nowlan, Steven J. and Hinton, Geoffrey E.},
  title = {Adaptive Mixtures of Local Experts},
  journal = {Neural Computation},
  volume = {3},
  number = {1},
  pages = {79--87},
  year = {1991},
  doi = {10.1162/neco.1991.3.1.79}
}

@article{JordanJacobs1994,
  author = {Jordan, Michael I. and Jacobs, Robert A.},
  title = {Hierarchical Mixtures of Experts and the {EM} Algorithm},
  journal = {Neural Computation},
  volume = {6},
  number = {2},
  pages = {181--214},
  year = {1994},
  doi = {10.1162/neco.1994.6.2.181}
}

@inproceedings{Shazeer2017,
  author = {Shazeer, Noam and Mirhoseini, Azalia and Maziarz, Krzysztof and Davis, Andy and Le, Quoc and Hinton, Geoffrey and Dean, Jeff},
  title = {Outrageously Large Neural Networks: The Sparsely-Gated Mixture-of-Experts Layer},
  booktitle = {International Conference on Learning Representations},
  year = {2017},

}

@inproceedings{Lepikhin2020,
  author = {Lepikhin, Dmitry and Lee, HyoukJoong and Xu, Yuanzhong and Chen, Dehao and Firat, Orhan and Huang, Yanping and Krikun, Maxim and Shazeer, Noam and Chen, Zhifeng},
  title = {{GShard}: Scaling Giant Models with Conditional Computation and Automatic Sharding},
  booktitle = {International Conference on Learning Representations},
  year = {2021},

}

@article{Fedus2021,
  author = {Fedus, William and Zoph, Barret and Shazeer, Noam},
  title = {Switch Transformers: Scaling to Trillion Parameter Models with Simple and Efficient Sparsity},
  journal = {Journal of Machine Learning Research},
  volume = {23},
  number = {120},
  pages = {1--39},
  year = {2022},

}

@inproceedings{Du2022,
  author = {Du, Nan and Huang, Yanping and Dai, Andrew M. and Tong, Simon and Lepikhin, Dmitry and Xu, Yuanzhong and Krikun, Maxim and Zhou, Yanqi and Yu, Adams Wei and Firat, Orhan and Zoph, Barret and Fedus, Liam and Bosma, Maarten P. and Zhou, Zongwei and Wang, Tao and Wang, Emma and Webster, Kellie and Pellat, Marie and Robinson, Kevin and Meier-Hellstern, Kathleen and Duke, Toju and Dixon, Lucas and Zhang, Kun and Le, Quoc and Wu, Yonghui and Chen, Zhifeng and Cui, Claire},
  title = {{GLaM}: Efficient Scaling of Language Models with Mixture-of-Experts},
  booktitle = {Proceedings of the 39th International Conference on Machine Learning},
  series = {Proceedings of Machine Learning Research},
  volume = {162},
  pages = {5547--5569},
  publisher = {PMLR},
  year = {2022},

}

@inproceedings{Riquelme2021,
  author = {Riquelme, Carlos and Puigcerver, Joan and Mustafa, Basil and Neumann, Maxim and Jenatton, Rodolphe and Pinto, Andr{\'e} Susano and Keysers, Daniel and Houlsby, Neil},
  title = {Scaling Vision with Sparse Mixture of Experts},
  booktitle = {Advances in Neural Information Processing Systems},
  volume = {34},
  pages = {8583--8595},
  year = {2021},

}

@article{Jiang2024,
  author = {Jiang, Albert Q. and Sablayrolles, Alexandre and Roux, Antoine and Mensch, Arthur and Savary, Blanche and Bamford, Chris and Chaplot, Devendra Singh and de las Casas, Diego and Hanna, Emma Bou and Bressand, Florian and others},
  title = {Mixtral of Experts},
  journal = {arXiv preprint arXiv:2401.04088},
  year = {2024},

  eprint = {2401.04088},
  archivePrefix = {arXiv}
}

@inproceedings{NguyenTopK2024,
  author = {Nguyen, Huy and Akbarian, Pedram and Yan, Fanqi and Ho, Nhat},
  title = {Statistical Perspective of Top-{K} Sparse Softmax Gating Mixture of Experts},
  booktitle = {International Conference on Learning Representations},
  year = {2024},

}

@article{NguyenSoftmax2025,
  author = {Nguyen, Huy and Ho, Nhat and Rinaldo, Alessandro},
  title = {Convergence Rates for Softmax Gating Mixture of Experts},
  journal = {IEEE Transactions on Information Theory},
  volume = {72},
  number = {2},
  pages = {1276--1304},
  year = {2026},
  doi = {10.1109/TIT.2025.3647061}
}

@article{Bariletto2026,
  author = {Bariletto, Nicola and Nguyen, Huy and Ho, Nhat and Rinaldo, Alessandro},
  title = {On {Bayesian} Softmax-Gated Mixture-of-Experts Models},
  journal = {arXiv preprint arXiv:2604.20551},
  year = {2026},

  eprint = {2604.20551},
  archivePrefix = {arXiv}
}

@article{Nguyen2013,
  author = {Nguyen, XuanLong},
  title = {Convergence of Latent Mixing Measures in Finite and Infinite Mixture Models},
  journal = {The Annals of Statistics},
  volume = {41},
  number = {1},
  pages = {370--400},
  year = {2013},
  doi = {10.1214/12-AOS1065}
}

@article{HoNguyen2016EJS,
  author = {Ho, Nhat and Nguyen, XuanLong},
  title = {On Strong Identifiability and Convergence Rates of Parameter Estimation in Finite Mixtures},
  journal = {Electronic Journal of Statistics},
  volume = {10},
  number = {1},
  pages = {271--307},
  year = {2016},
  doi = {10.1214/16-EJS1105}
}

@article{HoNguyen2016AOS,
  author = {Ho, Nhat and Nguyen, XuanLong},
  title = {Convergence Rates of Parameter Estimation for Some Weakly Identifiable Finite Mixtures},
  journal = {The Annals of Statistics},
  volume = {44},
  number = {6},
  pages = {2726--2755},
  year = {2016},
  doi = {10.1214/16-AOS1444}
}

@article{HeinrichKahn2018,
  author = {Heinrich, Philippe and Kahn, Jonas},
  title = {Strong Identifiability and Optimal Minimax Rates for Finite Mixture Estimation},
  journal = {The Annals of Statistics},
  volume = {46},
  number = {6A},
  pages = {2844--2870},
  year = {2018},
  doi = {10.1214/17-AOS1641}
}

@article{WuYang2020,
  author = {Wu, Yihong and Yang, Pengkun},
  title = {Optimal Estimation of {Gaussian} Mixtures via Denoised Method of Moments},
  journal = {The Annals of Statistics},
  volume = {48},
  number = {4},
  pages = {1981--2007},
  year = {2020},
  doi = {10.1214/19-AOS1873}
}

@article{NguyenSeparation2026,
  author = {Nguyen, Huy and Le, Dung and Rinaldo, Alessandro and Ho, Nhat},
  title = {On the Geometry of Separation in Finite {Gaussian} Mixtures},
  journal = {arXiv preprint arXiv:2606.16179},
  year = {2026},

  eprint = {2606.16179},
  archivePrefix = {arXiv}
}

@book{VandeGeer2000,
  author = {van de Geer, Sara},
  title = {Empirical Processes in {M}-Estimation},
  series = {Cambridge Series in Statistical and Probabilistic Mathematics},
  volume = {6},
  publisher = {Cambridge University Press},
  year = {2000},

}

@book{Villani2003,
  author = {Villani, C{\'e}dric},
  title = {Topics in Optimal Transportation},
  series = {Graduate Studies in Mathematics},
  volume = {58},
  publisher = {American Mathematical Society},
  year = {2003},

}

@article{BrudnyiGanzburg1973,
  author = {Brudnyi, Yu. A. and Ganzburg, M. I.},
  title = {On an Extremal Problem for Polynomials in $n$ Variables},
  journal = {Mathematics of the USSR-Izvestiya},
  volume = {7},
  number = {2},
  pages = {345--356},
  year = {1973},
  doi = {10.1070/IM1973v007n02ABEH001941}
}

@article{Nazarov1993,
  author = {Nazarov, Fedor L.},
  title = {Local Estimates for Exponential Polynomials and Their Applications to Inequalities of the Uncertainty Principle Type},
  journal = {Algebra i Analiz},
  volume = {5},
  number = {4},
  pages = {3--66},
  year = {1993},

  note = {English translation: St. Petersburg Mathematical Journal 5(4), 663--717, 1994}
}

@book{BerlinetThomasAgnan2004,
  author = {Berlinet, Alain and Thomas-Agnan, Christine},
  title = {Reproducing Kernel {Hilbert} Spaces in Probability and Statistics},
  publisher = {Kluwer Academic Publishers},
  year = {2004},
  doi = {10.1007/978-1-4419-9096-9}
}

@book{DeBoor2001,
  author = {de Boor, Carl},
  title = {A Practical Guide to Splines},
  edition = {Revised},
  series = {Applied Mathematical Sciences},
  volume = {27},
  publisher = {Springer},
  year = {2001},

}

@article{GassiatVanHandel2013,
  author = {Gassiat, Elisabeth and van Handel, Ramon},
  title = {The Local Geometry of Finite Mixtures},
  journal = {Transactions of the American Mathematical Society},
  volume = {366},
  number = {2},
  pages = {1047--1072},
  year = {2014},
  doi = {10.1090/S0002-9947-2013-06041-2}
}

@article{Sriperumbudur2010,
  author = {Sriperumbudur, Bharath K. and Gretton, Arthur and Fukumizu, Kenji and Sch{\"o}lkopf, Bernhard and Lanckriet, Gert R. G.},
  title = {{Hilbert} Space Embeddings and Metrics on Probability Measures},
  journal = {Journal of Machine Learning Research},
  volume = {11},
  number = {50},
  pages = {1517--1561},
  year = {2010},

}

@book{BoucheronLugosiMassart2013,
  author = {Boucheron, St{\'e}phane and Lugosi, G{\'a}bor and Massart, Pascal},
  title = {Concentration Inequalities: A Nonasymptotic Theory of Independence},
  publisher = {Oxford University Press},
  year = {2013},
  doi = {10.1093/acprof:oso/9780199535255.001.0001}
}

@article{HoYangJordan2022,
  author = {Ho, Nhat and Yang, Chiao-Yu and Jordan, Michael I.},
  title = {Convergence Rates for {Gaussian} Mixtures of Experts},
  journal = {Journal of Machine Learning Research},
  volume = {23},
  number = {323},
  pages = {1--81},
  year = {2022},

}

@inproceedings{NguyenDemystifying2023,
  author = {Nguyen, Huy and Nguyen, TrungTin and Ho, Nhat},
  title = {Demystifying Softmax Gating Function in {Gaussian} Mixture of Experts},
  booktitle = {Advances in Neural Information Processing Systems},
  volume = {36},
  year = {2023},
  doi = {10.52202/075280-0206},

}

@inproceedings{Dai2024DeepSeekMoE,
  author = {Damai Dai and Chengqi Deng and Chenggang Zhao and R. X. Xu and Huazuo Gao and Deli Chen and Jiashi Li and Wangding Zeng and Xingkai Yu and Y. Wu and Zhenda Xie and Y. K. Li and Panpan Huang and Fuli Luo and Chong Ruan and Zhifang Sui and Wenfeng Liang},
  title = {{DeepSeekMoE}: Towards Ultimate Expert Specialization in Mixture-of-Experts Language Models},
  booktitle = {Proceedings of the 62nd Annual Meeting of the Association for Computational Linguistics (Volume 1: Long Papers)},
  pages = {1280--1297},
  year = {2024},
  doi = {10.18653/v1/2024.acl-long.70},

}

@article{Figalli2010PartialTransport,
  author = {Figalli, Alessio},
  title = {The Optimal Partial Transport Problem},
  journal = {Archive for Rational Mechanics and Analysis},
  volume = {195},
  number = {2},
  pages = {533--560},
  year = {2010},
  doi = {10.1007/s00205-008-0212-7}
}

@article{NguyenDeepSeekTheory2025,
  author = {Nguyen, Huy and Doan, Thong T. and Pham, Quang and Bui, Nghi D. Q. and Ho, Nhat and Rinaldo, Alessandro},
  title = {On {DeepSeekMoE}: Statistical Benefits of Shared Experts and Normalized Sigmoid Gating},
  journal = {arXiv preprint arXiv:2505.10860},
  year = {2025},

  eprint = {2505.10860},
  archivePrefix = {arXiv}
}

@inproceedings{YanContaminated2025,
  author = {Yan, Fanqi and Nguyen, Huy and Le, Dung and Akbarian, Pedram and Ho, Nhat and Rinaldo, Alessandro},
  title = {On Minimax Estimation of Parameters in Softmax-Contaminated Mixture of Experts},
  booktitle = {Advances in Neural Information Processing Systems},
  volume = {38},
  pages = {7488--7528},
  year = {2025},
  doi = {10.52202/085713-0233},

}

@inproceedings{ChenDeepMoE2022,
  author = {Chen, Zixiang and Deng, Yihe and Wu, Yue and Gu, Quanquan and Li, Yuanzhi},
  title = {Towards Understanding the {Mixture-of-Experts} Layer in Deep Learning},
  booktitle = {Advances in Neural Information Processing Systems},
  volume = {35},
  year = {2022}
}

@inproceedings{ChowdhuryPatchMoE2023,
  author = {Chowdhury, Mohammed Nowaz Rabbani and Zhang, Shuai and Wang, Meng and Liu, Sijia and Chen, Pin-Yu},
  title = {Patch-level Routing in {Mixture-of-Experts} is Provably Sample-efficient for Convolutional Neural Networks},
  booktitle = {Proceedings of the 40th International Conference on Machine Learning},
  series = {Proceedings of Machine Learning Research},
  volume = {202},
  pages = {6074--6114},
  year = {2023}
}

@inproceedings{LiContinualMoE2025,
  author = {Li, Hongbo and Lin, Sen and Duan, Lingjie and Liang, Yingbin and Shroff, Ness B.},
  title = {Theory on {Mixture-of-Experts} in Continual Learning},
  booktitle = {International Conference on Learning Representations},
  year = {2025}
}

@inproceedings{LePromptMoE2024,
  author = {Le, Minh and Nguyen, An and Nguyen, Huy and Nguyen, Trang and Pham, Trang and Ngo, Linh Van and Ho, Nhat},
  title = {Mixture of Experts Meets Prompt-Based Continual Learning},
  booktitle = {Advances in Neural Information Processing Systems},
  volume = {37},
  year = {2024}
}

@article{BrudnyiYomdin2016,
  author = {Brudnyi, Alexander and Yomdin, Yosef},
  title = {Norming Sets and Related {Remez}-type Inequalities},
  journal = {Journal of the Australian Mathematical Society},
  volume = {100},
  number = {2},
  pages = {163--181},
  year = {2016},
  doi = {10.1017/S1446788715000488},

}

@inproceedings{NguyenSigmoid2024,
  author = {Nguyen, Huy and Ho, Nhat and Rinaldo, Alessandro},
  title = {Sigmoid Gating is More Sample Efficient than Softmax Gating in Mixture of Experts},
  booktitle = {Advances in Neural Information Processing Systems},
  volume = {37},
  year = {2024}
}

@inproceedings{NguyenCosine2025,
  author = {Nguyen, Huy and Akbarian Saravi, Pedram and Pham, Trang and Nguyen Vu, Thien Trang and Zhang, Shujian and Ho, Nhat},
  title = {Statistical Advantages of Perturbing Cosine Router in Mixture of Experts},
  booktitle = {International Conference on Learning Representations},
  year = {2025}
}

@article{Le2026HeterogeneousPartialTransport,
  author = {Le, Dung and Nguyen, Huy and Pham, Trang and Rinaldo, Alessandro and Ho, Nhat},
  title = {Characterizing Heterogeneous Rates in Finite Mixture Estimation via Partial Optimal Transport},
  journal = {arXiv preprint arXiv:2609.16622},
  year = {2026},
  eprint = {2609.16622},
  archivePrefix = {arXiv}
}

@inproceedings{Le2022MultimarginalPartialTransport,
  author = {Le, Khang and Nguyen, Huy and Nguyen, Khai and Pham, Tung and Ho, Nhat},
  title = {On Multimarginal Partial Optimal Transport: Equivalent Forms and Computational Complexity},
  booktitle = {Proceedings of the 25th International Conference on Artificial Intelligence and Statistics},
  series = {Proceedings of Machine Learning Research},
  volume = {151},
  pages = {4397--4413},
  publisher = {PMLR},
  year = {2022}
}

@inproceedings{Han2024FuseMoE,
  author = {Han, Xing and Nguyen, Huy and Harris, Carl and Ho, Nhat and Saria, Suchi},
  title = {{FuseMoE}: Mixture-of-Experts Transformers for Fleximodal Fusion},
  booktitle = {Advances in Neural Information Processing Systems},
  volume = {37},
  pages = {67850--67900},
  year = {2024},
  doi = {10.52202/079017-2167}
}

\end{document}